\documentclass[10pt,twocolumn,letterpaper]{article}

\usepackage{cvpr}

\usepackage{booktabs}
\usepackage{multirow} 
\usepackage[most]{tcolorbox}

\usepackage{amssymb}
\usepackage{amsmath}
\usepackage{multirow} 
\usepackage[most]{tcolorbox}
\usepackage{booktabs}
\usepackage{algorithm}
\usepackage{algorithmic}
\usepackage{tabularx}
\usepackage{booktabs}

\definecolor{cvprblue}{rgb}{0.21,0.49,0.74}
\usepackage[pagebackref,breaklinks,colorlinks,allcolors=cvprblue]{hyperref}

\def\confName{CVPR}
\def\confYear{2027}

\title{TPD: Temporal Prior Decoupling for Text-to-Video Diffusion Models}

\author{
   Taewon Kang$^{1}$, Matthias Zwicker$^{1}$ \\
   $^{1}$University of Maryland at College Park, United States \\
   {\large \texttt{taewon@umd.edu, zwicker@umd.edu}} \\
}

\definecolor{rxgreen}{RGB}{0,143,76}

\begin{document}
\maketitle

\begin{abstract}
    Text-to-video diffusion models generate temporally coherent content from natural language, yet when a prompt describes an early scene that persists while a new event emerges on top of it (such as ``a tall sandcastle standing on a beach where a wave rushes in and washes it away''), generation frequently fails to realize the late-segment event in the corresponding frames. We identify this failure as Temporal Prior Suppression (TPS), in which the dominant prior of the early segment captures the cross-attention trajectory across the temporal axis and suppresses the guidance signal needed for late-segment realization. We introduce Temporal Prior Decoupling (TPD), a training-free heuristic that strengthens the suppressed late-segment signal during diffusion sampling. TPD constructs a temporal counterfactual by conditioning on the early segment alone, and defines the discrepancy between the full-prompt and counterfactual predictions as a suppressed signal direction. Rather than removing a direction as in prior subtractive projection methods, TPD adds this direction back to the guidance update whenever the alignment of the update with it falls below a threshold. The threshold follows a hand-designed schedule over diffusion timestep and video frame, so the correction acts mainly on late frames and structure-forming steps and leaves early frames largely untouched. TPD runs entirely within standard diffusion sampling without retraining and is defined purely in classifier-free guidance space, so it applies to any backbone that exposes a CFG update. Experiments show that TPD substantially improves late-concept realization while preserving temporal coherence and visual fidelity, and that the targeted suppression recurs across distinct text-to-video backbones.
\end{abstract}

\section{Introduction}
 
Text-to-video diffusion models synthesize temporally coherent visual content from natural language with high fidelity~\cite{mochi, kong2024hunyuanvideo, yang2025cogvideox, openai, veo2024}. Yet a fundamental limitation persists. When a prompt describes an early scene that should \emph{persist} while a new concept \emph{emerges} over it (\emph{``a tall sandcastle standing on a beach where a wave rushes in and washes it away''}), current models systematically fail to realize the later concept in the frames where it is intended to appear. In practice, the sandcastle-on-a-beach prior dominates the entire denoising trajectory, and even in frames that should depict the wave and the collapse, the model continues to render the intact early scene. This failure is not caused by ambiguous prompts; it arises from the model's tendency to amplify the statistically dominant early-scene prior, which captures cross-attention across the full temporal axis and suppresses the guidance signal required for late-segment realization.
 
We refer to this phenomenon as \textbf{Temporal Prior Suppression (TPS)}. TPS is distinct from negation failures~\cite{anonymous2026negate, alhamoud2025vision, singh2024learn}, where models render explicitly excluded concepts, and from compositional collapse~\cite{anonymous2026dcr, liu2022compositional, chefer2023attend}, where models default to frequent co-occurrence patterns. In TPS the model does not misread the prompt globally, but fails to enforce the requested content at the correct temporal position. While concept neglect also occurs spatially in image generation, placing the right concept in the right \emph{frame} is specific to video, where an early-scene prior propagates its dominance forward along the temporal axis and overwrites the frames in which the later concept should emerge.
 
Existing temporal-control methods do not fit this setting. Segmented-supervision approaches~\cite{cai2025ditctrl, wu2025mind, chen2026prompt, schiber2026tempocontrol} require the description to be pre-decomposed into per-shot prompts, event timestamps, or scene cuts. Training-free transition methods keep the prompt single but assume the change is a smooth attribute \emph{morph}. From-Prompt-to-Progression~\cite{lo2025prompt}, for example, walks each frame along an $\epsilon$-space direction between an initial and a final attribute, sliding a persistent subject from state $A$ to state $B$. Emergence has no such attribute axis, since a new event appears while the early scene persists, rather than the sandcastle morphing into a wave. Our setting takes a single, unsegmented prompt, with no timestamps required, and asks that the emergent event be realized in its temporal region while the early scene is preserved.
 
Constraint-based methods approach related problems from a complementary direction. NEGATE~\cite{anonymous2026negate} enforces an \emph{upper bound} on the projection of the guidance update along a negated concept direction, and DCR~\cite{anonymous2026dcr} removes the component aligned with a counterfactual drift toward the model's frequent completion. Both are \emph{subtractive}. TPS is the inverse problem. The late-segment signal is not undesired, since it is requested and present in the conditioning, yet it is outcompeted by the early-scene prior. A subtractive mechanism cannot recover an omitted signal, so a \emph{restorative} mechanism that strengthens the suppressed signal is needed.
 
We propose \textbf{Temporal Prior Decoupling (TPD)}, a training-free framework proceeding in three stages. First, an LLM decomposes the input prompt into early and late segments and provides heuristic estimates of the transition point and of the relative prior strength of each segment, without user-provided timestamps. Second, TPD constructs a \emph{temporal counterfactual} by conditioning the denoiser on the early segment alone, defining the discrepancy between the full-prompt and counterfactual predictions as the \emph{suppressed signal direction}. Third, whenever the alignment of the guidance update with this direction falls below a threshold $c(\tau, i)$, TPD adds the smallest multiple of the direction that brings the alignment up to the threshold. The threshold is a hand-designed function of diffusion timestep and video frame that is large in late-segment frames and structure-forming steps and close to zero in early frames and fine-detail steps. Where NEGATE imposes $a_t^\top \delta \leq b_t$ and DCR removes a drift-aligned component, TPD imposes $a_\tau^\top \delta \geq c(\tau, i)$, a threshold from below rather than from above. Two elements distinguish TPD from a sign flip of prior work. The suppressed direction is built from a temporal counterfactual rather than a negated or attractor prompt, and the threshold varies over a \emph{two-dimensional} (diffusion-timestep $\times$ video-frame) schedule rather than a single axis. Meeting the threshold is a property of the guidance update rather than of the decoded video, so TPD makes the late concept more likely to appear but does not guarantee it.
 
To enable systematic evaluation, we assemble a dataset of single-prompt transitional descriptions organized by \emph{emergence mechanism} (entity entry, gathering, phenomenon onset, and causal transformation) rather than surface scene type. Standard benchmarks such as WebVid and MSR-VTT contain no structured emergence prompts, while attribute-transition suites such as CAT-Bench~\cite{lo2025prompt} probe persistent-subject morphs rather than new events over a persisting scene.
 
Our contributions are threefold: (1) we identify TPS as a failure mode of text-to-video models, distinguishing it from negation, compositional collapse, and attribute-morph transitions; (2) we introduce TPD, a restorative projection that raises the alignment of the guidance update with the suppressed direction to a heuristic threshold scheduled over diffusion timestep and video frame, together with an attribute-translation (always-add) ablation isolating the value of the adaptive projection; and (3) we construct an evaluation dataset with metrics for late-concept realization and prior suppression. Because TPD is defined entirely in classifier-free guidance space, it applies unchanged to any backbone that exposes a CFG update; we conduct our controlled quantitative study on a single backbone where each component can be cleanly ablated, and observe qualitatively that the same suppression recurs across distinct backbones.

\section{Related Work}
\subsection{Temporal and Multi-Concept Control in Video Diffusion}
One line of work supplies \emph{segmented supervision}. DiTCtrl~\cite{cai2025ditctrl} shares and masks attention across a prompt sequence, Mind-the-Time~\cite{wu2025mind} trains temporally-aware cross-attention to place events at designated timestamps, and Prompt Relay~\cite{chen2026prompt} and TempoControl~\cite{schiber2026tempocontrol} schedule attention so that successive events are handed off over time. All presuppose that the description has already been decomposed into ordered segments. Our input is instead a \emph{single, unsegmented} sentence whose temporal structure is latent, with no timestamps, per-frame captions, or scene cuts, and we recover the early/late decomposition internally, acting on the guidance signal rather than on attention maps. A second line keeps the prompt single but treats temporal change as a smooth attribute \emph{morph}. From-Prompt-to-Progression~\cite{lo2025prompt} walks each frame along an $\epsilon$-space direction between an initial and a final attribute, while Gen-L~\cite{wang2023gen} and FreeNoise~\cite{qiu2023freenoise} interpolate between prompts. Emergence breaks the underlying assumption of a one-dimensional $A{\to}B$ slide of a persistent subject's attribute, because a new entity or event appears while the early scene persists, so morph-style interpolation has no well-defined direction to follow. We make this concrete by including an always-add, attribute-translation variant of our guidance as an ablation. Story-level approaches~\cite{kang2025text2story} stitch separately generated segments and again require a pre-split narrative.

\subsection{Inference-Time Guidance and Semantic Steering}
Classifier-free guidance (CFG)~\cite{ho2022classifier} is refined at sampling time by many training-free methods, including dynamic thresholding~\cite{saharia2022photorealistic}, manifold-constrained updates~\cite{chung2024cfg++}, spectral reweighting~\cite{si2024freeu}, external objective guidance~\cite{bansal2023universal}, energy-based constrained sampling~\cite{zampini2025training, zhang2025constrained}, multi-objective schedules~\cite{xie2025dymo}, and frequency-aware flow matching~\cite{ren2026frequency}. Semantic-steering methods act more selectively. SEGA~\cite{brack2023sega} moves the trajectory along isolated concept vectors, Semantic Guidance Tuning~\cite{kang2023semantic} nudges guidance back toward neglected prompt concepts, and S-CFG~\cite{shen2024rethinking} adapts guidance strength across spatial regions. These methods also intervene on the guidance update, but they operate on static images and modulate steering \emph{magnitude}. Our correction differs on two axes. It is applied only when the alignment of the update with the emergent direction falls below a threshold, and the threshold varies jointly over diffusion timestep \emph{and} video frame, an axis with no analogue in spatially adaptive image guidance~\cite{shen2024rethinking}.

\subsection{Constraint-Based Guidance and the Suppression--Restoration Duality}
Recent work casts inference-time semantic control as a constraint on the guidance update. NEGATE~\cite{anonymous2026negate} bounds the guidance projection along a negated direction from \emph{above}, driving the forbidden concept out, and DCR~\cite{anonymous2026dcr} removes the component aligned with a counterfactual drift toward the model's frequent completion. Both are \emph{subtractive}, since they weaken an undesired direction. Our failure mode is the mirror image. The late-segment concept is explicitly requested and present in the conditioning, yet outcompeted by the early-scene prior and under-expressed where it should appear. Removing a direction cannot recover an omitted signal, so we instead add the suppressed direction back when the alignment of the update with it falls below a threshold, and we schedule this threshold over two dimensions (diffusion timestep $\times$ video frame) rather than the single-axis schedules of these formulations.

\subsection{Diffusion Models for Image and Video Generation}
Denoising diffusion~\cite{ho2020denoising} and latent diffusion~\cite{rombach2022high} now extend across a wide range of video architectures~\cite{ho2022video, blattmann2023align, blattmann2023stable, ho2022imagen, singer2022make, bar2024lumiere, girdhar2311emu, lian2023llm, wang2023modelscope, henschel2024streamingt2v, he2022latent, hong2022cogvideo, villegas2022phenaki, ge2023preserve, wang2024lavie}, alongside proprietary systems such as Sora~\cite{openai}, Veo~\cite{veo2024, veo2}, and Movie Gen~\cite{polyak2025moviegencastmedia}. Recent open backbones, namely CogVideoX~\cite{yang2025cogvideox}, HunyuanVideo~\cite{kong2024hunyuanvideo, wu2025hunyuanvideo}, and Mochi~\cite{mochi}, serve as the frozen backbones in our study. Scale alone does not remove the tendency to anchor a clip to its early-scene prior, and the same suppression recurs across distinct architectures, text encoders, and schedulers, indicating a property of the current model class. Because our correction is defined entirely in classifier-free guidance space, it intervenes only at inference time and applies to any backbone with a CFG update.

\section{Method}
\label{sec:method}

\begin{figure*}[t]
    \centering
    \includegraphics[width=0.8\linewidth]{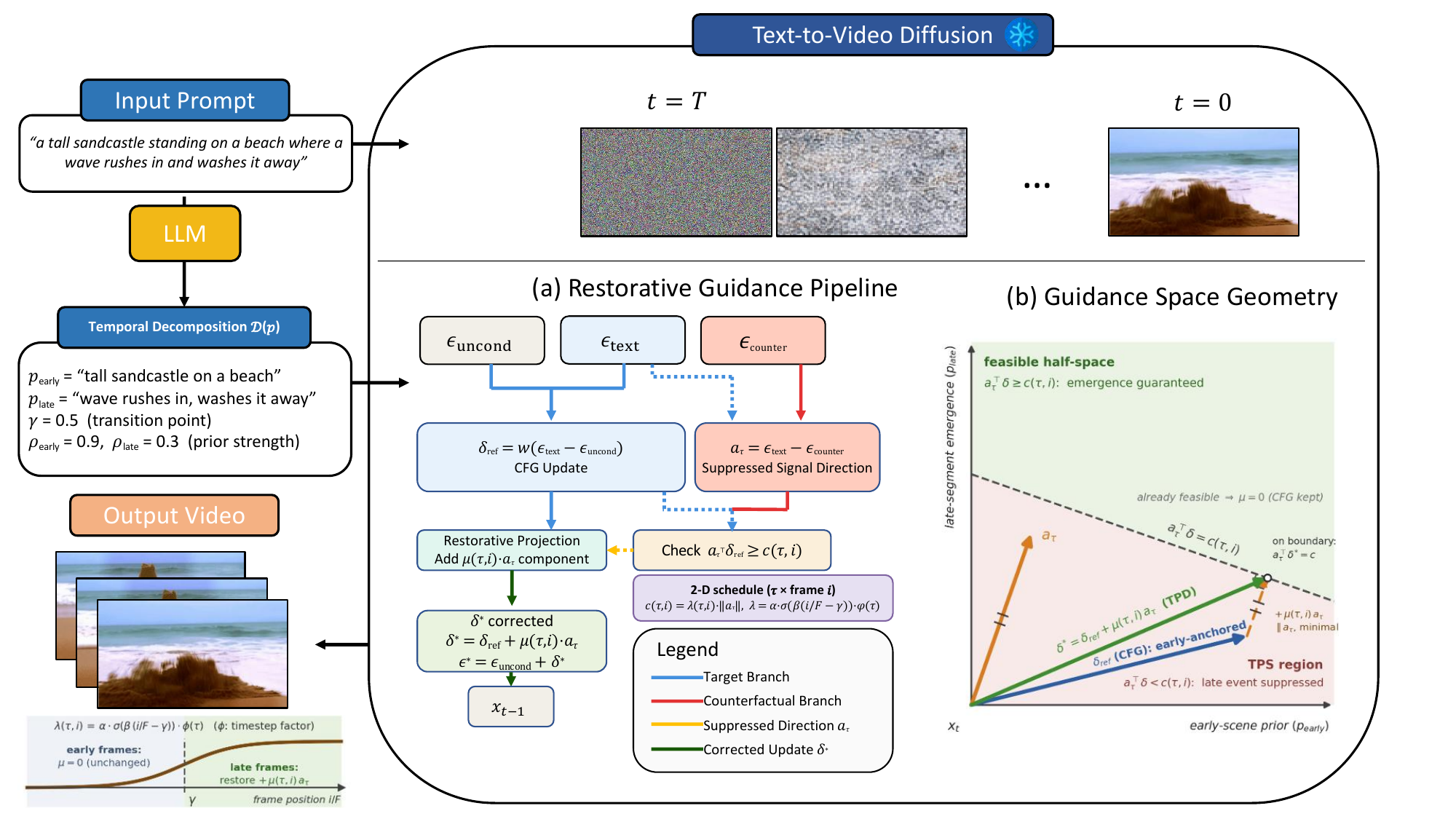}
    \caption{\textbf{Overview of Temporal Prior Decoupling (TPD).} An LLM decomposes the unsegmented prompt $p$ into the early segment $p_{\text{early}}$, the late segment $p_{\text{late}}$, the transition point $\gamma$, and the prior strengths $\rho_{\text{early}}, \rho_{\text{late}}$. \textbf{(a) Restorative Guidance Pipeline:} Three branches of the frozen denoiser are evaluated (unconditional $\epsilon_{\text{uncond}}$, full-prompt $\epsilon_{\text{text}}$, and temporal counterfactual $\epsilon_{\text{counter}}$ conditioned on $p_{\text{early}}$ alone), yielding the CFG update $\delta_{\text{ref}} = w(\epsilon_{\text{text}} - \epsilon_{\text{uncond}})$ and the suppressed signal direction $a_\tau = \epsilon_{\text{text}} - \epsilon_{\text{counter}}$. If $a_\tau^\top \delta_{\text{ref}} \geq c(\tau,i)$, where the threshold is scheduled over denoising step $\tau$ and video frame $i$, the update is left untouched; otherwise the minimal-energy projection yields $\delta^\star = \delta_{\text{ref}} + \mu(\tau,i)\,a_\tau$. \textbf{(b) Guidance Space Geometry:} The threshold splits guidance space into a half-space in which the alignment of the update with $a_\tau$ reaches $c(\tau,i)$ and a half-space in which it falls short. Standard CFG lands in the latter, and TPD moves it onto the boundary along $a_\tau$ by the smallest displacement, applying no correction where the threshold is already met ($\mu = 0$). Reaching the threshold is a property of the guidance update at one step and does not by itself guarantee that the late concept appears in the generated video.}
    \label{fig:method_figure}
\end{figure*}

TPD corrects the guidance update using a threshold on its alignment with a suppressed late-segment direction. At each denoising timestep and video frame, we compute the smallest change to the reference guidance update that raises this alignment to a scheduled threshold. The threshold and its schedule are design choices rather than quantities derived from the desired video, so the correction is a heuristic that pushes the guidance toward the late segment by a controlled amount. Figure~\ref{fig:method_figure} overviews the pipeline and the underlying guidance-space geometry.

\subsection{Problem Setup}
\label{subsec:problem_setup}

Let $x_t \in \mathbb{R}^{F \times H \times W \times C}$ denote the latent at diffusion timestep $t \in [0,1]$, with $F$ video frames, $t=1$ pure noise, and $t=0$ the data sample. We write $\epsilon_\theta(x_t, t, c)$ for the denoiser prediction under conditioning $c$, and $\epsilon_{\text{uncond}} := \epsilon_\theta(x_t, t, \varnothing)$. Given a temporally structured prompt $p = [p_{\text{early}},\, p_{\text{late}}]$, the standard classifier-free guidance (CFG) update is
\begin{equation}
\delta_{\text{ref}} = w \left( \epsilon_{\text{text}} - \epsilon_{\text{uncond}} \right),
\label{eq:cfg}
\end{equation}
where $\epsilon_{\text{text}} := \epsilon_\theta(x_t, t, \Phi(p))$, $\Phi(\cdot)$ is the pretrained text encoder, and $w > 1$ is the guidance scale; the guided prediction is $\epsilon_{\text{target}} = \epsilon_{\text{uncond}} + \delta_{\text{ref}}$. Under TPS, $\delta_{\text{ref}}$ is dominated by the $p_{\text{early}}$ prior across the entire temporal axis, and the update at late frames fails to reflect $p_{\text{late}}$ despite its presence in the prompt. Our goal is to strengthen, at late frames $i$, the component of the corrected update $\delta^*(\tau, i)$ along the direction associated with $p_{\text{late}}$, while leaving the update at early frames unchanged.

\subsection{Temporal Decomposition via LLM}
\label{subsec:decomposition}

A deterministic LLM-based stage extracts
\begin{equation}
\mathcal{D}(p) = \left\{ p_{\text{early}},\; p_{\text{late}},\; \gamma \in [0,1],\; \rho_{\text{early}} \in [0,1],\; \rho_{\text{late}} \in [0,1] \right\},
\end{equation}
where $\gamma$ is the normalized transition point and $\rho_{\text{early}}, \rho_{\text{late}}$ are the LLM's estimates of how strongly a video model trained on internet-scale data favors each segment. These scores are heuristic judgments of the LLM rather than measurements of the backbone, and we use them only as a proxy for TPS risk, with a higher $\rho_{\text{early}}$ relative to $\rho_{\text{late}}$ indicating a higher risk. For the sandcastle prompt, this yields $p_{\text{early}}$ = ``tall sandcastle on a beach'', $p_{\text{late}}$ = ``wave rushes in, washes it away'', $\gamma = 0.5$, $\rho_{\text{early}} = 0.8$, $\rho_{\text{late}} = 0.25$, and the large gap indicates a high risk of TPS. \textbf{Implementation details are provided in Appendix~\ref{supp:implementation_decomposition}.}

\subsection{Temporal Counterfactual Construction}
\label{subsec:counterfactual}

We construct a \emph{temporal counterfactual} branch by conditioning the denoiser on $p_{\text{early}}$ alone:
\begin{equation}
\epsilon_{\text{counter}} := \epsilon_\theta(x_t, t, \Phi(p_{\text{early}})),
\end{equation}
the prediction the model would produce were the late segment absent, which is the prediction that TPS drives the trajectory toward. Unlike DCR~\cite{anonymous2026dcr}, which \emph{substitutes} a frequent-completion prompt, this counterfactual is obtained by \emph{truncating} the prompt, so the two constructions probe different model tendencies. We define the \emph{suppressed signal direction} as
\begin{equation}
a_\tau = \epsilon_{\text{text}} - \epsilon_{\text{counter}},
\label{eq:suppressed_direction}
\end{equation}
the incremental contribution that $p_{\text{late}}$ induces in the full-prompt denoiser beyond what $p_{\text{early}}$ alone would produce. Under TPS this contribution is present in $\epsilon_{\text{text}}$ but too weak to overcome the dominance of $p_{\text{early}}$. Because $p_{\text{early}} \neq p$, $a_\tau$ is in general a distinct direction in guidance space and cannot be written as a scalar multiple of $\delta_{\text{ref}}$. The correction below therefore adds a specific semantic component rather than rescaling guidance magnitude along a fixed direction as in $\delta_{\text{ref}} \mapsto c\,\delta_{\text{ref}}$. Prior projection methods construct their directions analogously from the same pretrained denoiser, namely $\epsilon_{\text{neg}} - \epsilon_{\text{uncond}}$ toward a forbidden concept in NEGATE~\cite{anonymous2026negate} and $\epsilon_{\text{probe}} - \epsilon_{\text{target}}$ toward a frequent completion in DCR, but keep the update \emph{below} a threshold along them. Here $a_\tau$ instead identifies a \emph{desired but weakly represented} direction, which we strengthen when the alignment of the update with it is low.

\subsection{Thresholded Restorative Projection}
\label{subsec:projection}

Writing $\phi_\tau(\delta) = a_\tau^\top \delta$ for the alignment of an update $\delta$ with $a_\tau$, TPD requires
\begin{equation}
\phi_\tau(\delta(\tau, i)) \geq c(\tau, i),
\label{eq:lowerbound}
\end{equation}
where $c(\tau, i)$ is a threshold whose form is given in Sec.~\ref{subsec:schedule}. The threshold is chosen heuristically to be large in late-segment frames and structure-forming steps and close to zero elsewhere. It is not derived from the content of the target video and does not certify any property of it. The constraint defines a closed half-space in guidance space, with the opposite orientation to the constraint of~\cite{anonymous2026negate}:
\begin{align}
\text{\cite{anonymous2026negate}:} &\quad a_t^\top \delta \leq b_t \quad \text{(upper bound, subtractive)} \label{eq:negate_constraint} \\
\text{TPD:} &\quad a_\tau^\top \delta \geq c(\tau, i) \quad \text{(threshold from below, restorative)} \label{eq:tpd_constraint}
\end{align}
The constraint in~\eqref{eq:negate_constraint} caps the alignment of the update with the negated direction, and the constraint in~\eqref{eq:tpd_constraint} requires a minimum alignment with the suppressed direction, which TPD reaches by adding that direction when the alignment is too low. Both act on the guidance update at a single denoising step, and neither guarantees the semantics of the decoded video, which also depends on the remaining denoising steps, the scheduler, and whether the backbone can render the concept at all. In TPD, an update that meets the threshold carries a stronger late-segment signal but does not by itself ensure that the late concept appears in the decoded video.

\subsection{Minimal-Energy Projection}
\label{subsec:minimal_energy}

The corrected update is the solution of the quadratic program
\begin{equation}
\delta^*(\tau, i) = \arg\min_{\delta} \frac{1}{2} \|\delta - \delta_{\text{ref}}\|_2^2 \quad \text{s.t.} \quad a_\tau^\top \delta \geq c(\tau, i),
\label{eq:qp}
\end{equation}
the update closest to the original CFG direction that satisfies the constraint. The objective is strictly convex and the constraint set a closed half-space, so the solution exists, is unique, and follows in closed form from the KKT conditions (derivation in Appendix~\ref{supp:derivation}):
\begin{equation}
\delta^*(\tau, i) = \delta_{\text{ref}} + \mu(\tau, i)\, a_\tau,
\label{eq:solution}
\end{equation}
with
\begin{equation}
\mu(\tau, i) = \frac{\max\!\left\{0,\; c(\tau, i) - a_\tau^\top \delta_{\text{ref}}\right\}}{\|a_\tau\|_2^2}.
\label{eq:mu}
\end{equation}
Correction occurs only when~\eqref{eq:lowerbound} is violated. If $a_\tau^\top \delta_{\text{ref}} \geq c(\tau, i)$ then $\mu = 0$ and $\delta_{\text{ref}}$ is unchanged; otherwise the added term is the smallest perturbation that reaches the threshold, with magnitude $\mu(\tau,i)\|a_\tau\|$ proportional to the shortfall. The final prediction is $\epsilon_t^*(\tau, i) = \epsilon_{\text{uncond}} + \delta^*(\tau, i)$. Compared with NEGATE's correction $\delta^* = \delta_{\text{ref}} - \lambda_t a_t$, TPD adds rather than removes signal along its direction, since the direction comes from a temporal counterfactual and the constraint binds from below. Projection onto a closed half-space is non-expansive, so the corrected update is Lipschitz in the reference and introduces no stiffness into the reverse-time ODE; a fixed $\varepsilon = 10^{-8}$ added to $\|a_\tau\|_2^2$ ensures numerical stability when $\|a_\tau\|$ is small. Wherever the threshold is already met, the original CFG update is used unchanged (full stability analysis in Appendix~\ref{supp:derivation}).

\subsection{Heuristic Threshold Schedule}
\label{subsec:schedule}

We set the threshold to
\begin{equation}
c(\tau, i) = \lambda(\tau, i) \cdot \|a_\tau\|,
\label{eq:bound}
\end{equation}
with the frame- and timestep-dependent weight
\begin{equation}
\lambda(\tau, i) = \alpha \cdot \sigma\!\left(\beta \cdot \left(\frac{i}{F-1} - \gamma\right)\right) \cdot \phi(\tau).
\label{eq:lambda}
\end{equation}
Because $c(\tau, i)$ is proportional to $\|a_\tau\|$, the constraint~\eqref{eq:lowerbound} is equivalent to requiring that the component of the update along the unit vector $a_\tau / \|a_\tau\|$ be at least $\lambda(\tau, i)$, so $\lambda$ directly sets how far the update is pushed along the suppressed direction. Both the product form of~\eqref{eq:lambda} and its individual factors are hand-designed. They encode where and when we want the correction to act, and their hyperparameters $\beta$, $q$, and $\alpha_0$ are fixed by hand rather than fitted. The \textbf{frame position factor}, a sigmoid of the normalized frame index centered at the LLM-estimated transition point $\gamma$, is close to zero for early frames ($i/(F-1) \ll \gamma$), so the threshold there is near zero and the early segment is left largely unchanged, and close to one for late frames, where the full weight $\alpha\,\phi(\tau)$ applies. The sharpness $\beta > 0$ controls how abruptly the weight switches on. Here $i \in \{0, \dots, F-1\}$ indexes the $F$ latent frames, and $\tau \in \{T-1, \dots, 0\}$ counts the remaining denoising steps, so $\tau = T-1$ at the first, noisiest step. The \textbf{denoising timestep factor} $\phi(\tau) = \big(\tau / (T-1)\big)^q$ with $q = 2.0$, similar to the scheduling of~\cite{anonymous2026negate}, puts more weight on the early, structure-forming denoising steps and less on the fine-detail steps, where a change would mainly affect texture. The \textbf{prior strength factor}
\begin{equation}
\alpha = \alpha_0 \cdot \max\{0,\, \rho_{\text{early}} - \rho_{\text{late}}\}
\label{eq:alpha}
\end{equation}
scales the threshold with the LLM's estimate of how strongly the early segment dominates. When $\rho_{\text{early}} \approx \rho_{\text{late}}$, $\alpha \approx 0$ and TPD reduces to standard CFG. Because $\rho$ comes from an LLM judgment rather than from the backbone, we discuss this choice in Sec.~\ref{sec:limitation_future}. This two-dimensional schedule extends the one-dimensional temporal scheduling of prior projection methods~\cite{anonymous2026negate, anonymous2026dcr} to the joint $(\text{diffusion timestep}) \times (\text{video frame})$ space. The video-frame axis has no analogue in their single-axis schedules and is what localizes the correction to the late-segment frames. We isolate its contribution, together with that of the adaptive projection, through an always-add (attribute-translation) variant in our ablations.

\subsection{Unified Treatment}
\label{subsec:unified}

The complete update, given by~\eqref{eq:solution} with the threshold~\eqref{eq:bound}--\eqref{eq:alpha}, reduces to standard CFG when the alignment already meets the threshold, applies the minimal-energy correction otherwise, and applies no correction to early frames or fine-detail steps regardless of the other axis. Together with NEGATE~\cite{anonymous2026negate} and DCR~\cite{anonymous2026dcr}, TPD belongs to a family of guidance corrections defined by half-space constraints on the CFG update:

\begin{table}[t]
\centering
\renewcommand{\arraystretch}{1.3}
\setlength{\tabcolsep}{4pt}
\footnotesize
\begin{tabularx}{\columnwidth}{l
  >{\raggedright\arraybackslash}X
  >{\raggedright\arraybackslash}X
  c}
\toprule
\textbf{Method} & \textbf{Direction} $a_t$ & \textbf{Constraint} & \textbf{Operation} \\
\midrule
NEGATE & $\epsilon_{\text{neg}} - \epsilon_{\text{uncond}}$ & $a_t^\top\delta \leq b_t$ & Subtractive \\
DCR & $\epsilon_{\text{probe}} - \epsilon_{\text{target}}$ & Remove aligned component & Subtractive \\
TPD (Ours) & $\epsilon_{\text{text}} - \epsilon_{\text{counter}}$ & $a_\tau^\top\delta \geq c(\tau,i)$ & Restorative \\
\bottomrule
\end{tabularx}
\label{tab:unified_family}
\end{table}

All three operate entirely within classifier-free guidance space, require no retraining, and compute their directions from the same pretrained denoiser. NEGATE and TPD are closed-form projections onto half-spaces of opposite orientation, addressing negation enforcement and temporal prior suppression respectively, while DCR removes a drift-aligned component to suppress compositional bias.

\section{Experiments and Results}
\subsection{Implementation Details}
TPD operates inside the sampling loop of a frozen backbone. At each denoising step we evaluate the unconditional, full-prompt, and temporal-counterfactual branches, and apply the smallest correction that raises the alignment $a_\tau^\top \delta$ to the threshold $c(\tau,i)$, leaving frames that already meet it unchanged. The threshold uses a sigmoid over normalized frame position with sharpness $\beta = 10.0$ and a polynomial schedule with $q = 2.0$; the overall strength uses $\alpha_0 = 1.0$, with stabilization $\varepsilon = 10^{-8}$. Temporal decomposition is a single deterministic GPT-4o query returning $p_{\text{early}}$, $p_{\text{late}}$, $\gamma$, and the two prior strengths; the input remains one unsegmented sentence, with no timestamps or scene boundaries required. Backbone parameters follow default configurations~\cite{mochi}, and no learnable parameters are introduced. We use Mochi as the primary backbone since its guidance interface admits the three-branch decomposition cleanly and each schedule component can be disabled in isolation; the formulation depends only on the classifier-free guidance update and therefore transfers without modification, and the targeted failure mode is shared across HunyuanVideo~\cite{kong2024hunyuanvideo} and CogVideoX~\cite{yang2025cogvideox} (Appendices~\ref{sec:detailed_qualitative}, \ref{supp:backbone}).

\subsection{Benchmarking Datasets}
Existing video benchmarks are \textbf{insufficient for evaluating temporal emergence}. Corpora such as MSR-VTT, WebVid, or MS-COCO pair one caption with one visual state, and a generation that renders only the opening scene throughout can still align well with a caption mentioning the later event, so the failure we study is invisible to aggregate clip--text similarity. Attribute-transition suites such as CAT-Bench~\cite{lo2025prompt} do vary content over time, but slide a persistent subject's single attribute between values and never test whether a \emph{new} entity or event can arise over a scene that continues to hold. We therefore construct a suite targeting \emph{position-resolved realization}, that is, whether the late-segment concept is present in the frames where it was requested. Our dataset consists of single, unsegmented prompts distributed evenly across eight categories (ENTRY, SCENE, ONSET, STATE, CAUSAL, WEATHER\_NIGHT, ACTIVITY, STRUCTURAL), grouped into four emergence mechanisms (entity entry, gathering, phenomenon onset, and causal transformation) rather than surface scene types. Every prompt pairs an early scene chosen to be a strong, self-consistent completion with a late concept that must emerge over it; the early setting \textbf{persists rather than is replaced}, and the late segment introduces a new entity, phenomenon, or state change instead of a single-attribute morph, keeping our dataset disjoint from attribute-translation settings. The suite is evaluation-only. \textbf{Full construction details, category definitions, and representative prompts are in Appendix~\ref{supp:detailed_dataset}}.

\begin{figure}[t]
    \centering
    \includegraphics[width=0.875\linewidth]{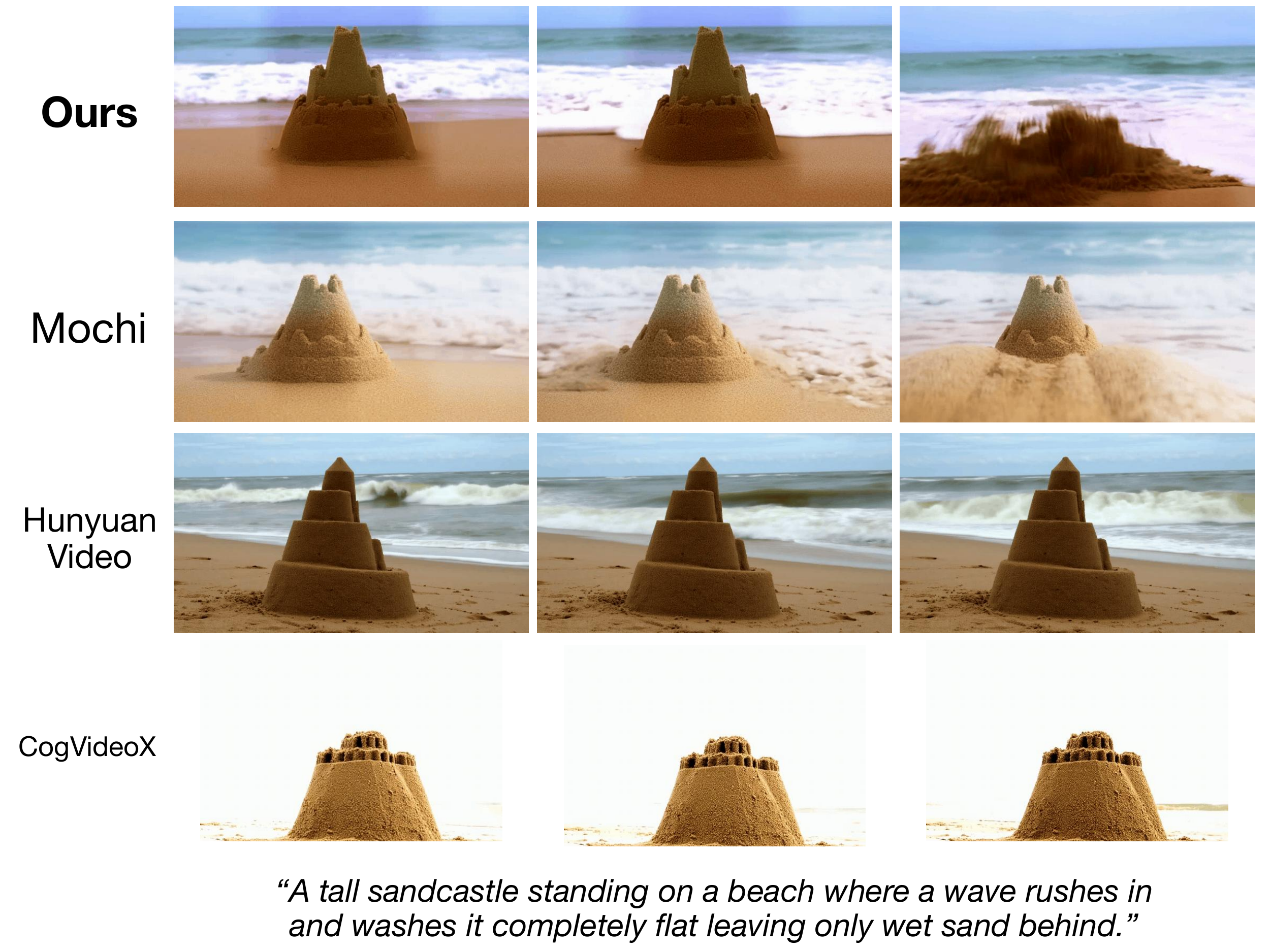}
    \caption{\textbf{Representative qualitative comparison (STRUCTURAL).}
    Prompt: ``a tall sandcastle standing on a beach where a wave rushes in and washes it completely flat leaving only wet sand behind''; frames in temporal order. All baselines carry the intact structure to the final frame. HunyuanVideo renders breaking waves that never reach the castle, and CogVideoX degenerates into a washed-out frame that loses the beach context as well. Our method keeps the sandcastle intact through roughly the first half of the clip. The water then reaches its base, erodes it from the sides, and the castle collapses into a low heap of wet sand by the final frame, so both segments appear in their temporal positions.}
    \label{fig:qualitative_main}
\end{figure}

\subsection{Qualitative Results}
\label{sec:qualitative}
Figure~\ref{fig:qualitative_main} shows a representative STRUCTURAL comparison; \textbf{extended comparisons across all eight categories are in Appendix~\ref{sec:detailed_qualitative}}. Across categories the baselines share one behavior. The opening frame satisfies the early segment, and subsequent frames repeat it with little change. In ENTRY, ACTIVITY, and SCENE the requested subjects or crowds never arrive (the shoreline stays empty, the highway stays clear), and in ONSET and WEATHER\_NIGHT the requested event never develops (a moonlit bay is rendered accurately, but the storm never forms). The hardest cases ask the model to destroy an intact object. For the sandcastle prompt, every baseline carries the structure to the final frame, one even placing an approaching wave in the background without letting it reach the castle. The backbones fail in distinguishable ways. CogVideoX often degenerates into washed-out frames that lose the early scene as well, whereas HunyuanVideo renders the opening with high visual quality and simply holds it. TPD corrects the update mainly in late-segment frames and structure-forming steps, so its outputs diverge from the baselines where the emergence belongs, and the sandcastle stays intact until the water reaches its base and only then collapses. We observe the same pattern across all eight categories.

\begin{table}[t]
\begin{small}
    \centering
    \resizebox{0.5\textwidth}{!}{%
    \begin{tabular}{lcccccc}
        \toprule
        \textbf{Method} & \textbf{CLIPScore} $\uparrow$ & \textbf{CLIP-early} $\downarrow$ & \textbf{BLIP} $\uparrow$ & \textbf{TCS} $\uparrow$ & \textbf{TVR} $\downarrow$ & \textbf{EPS} $\uparrow$ \\
        \midrule
        Mochi            & 0.3017 & 0.2865 & 0.6883 & 3.1850 & 0.4950 & 4.7500 \\
        HunyuanVideo     & 0.2964 & 0.2795 & 0.7098 & 2.8250 & 0.5550 & 4.6750 \\
        CogVideoX        & 0.2821 & 0.2822 & 0.6620 & 2.4050 & 0.6875 & 3.7475 \\
        \midrule
        CFG Scale-Up     & 0.3019 & 0.2863 & 0.6921 & 3.2250 & 0.5275 & 4.7650 \\
        Always-Add       & 0.3027 & 0.2850 & 0.7074 & 3.3325 & 0.4900 & 4.7675 \\
        \midrule
        \textbf{Ours}    & \textbf{0.3109} & \textbf{0.2580} & \textbf{0.7140} & \textbf{3.8775} & \textbf{0.3225} & \textbf{4.8800} \\
        \midrule
        Ours w/o Frame Axis        & 0.3085 & 0.2691 & 0.7096 & 3.4175 & 0.4775 & 4.8075 \\
        Ours w/o Timestep Schedule & 0.3083 & 0.2588 & 0.7082 & 3.4050 & 0.4475 & 4.8325 \\
        Ours w/o $\rho$-Adaptive $\alpha$ & 0.3084 & 0.2592 & 0.7044 & 3.2600 & 0.4725 & 4.8075 \\
        \midrule
    \end{tabular}}
    \caption{\textbf{Quantitative evaluation.} All metrics are averaged over the full evaluation set. Mochi, HunyuanVideo, and CogVideoX denote standard CFG baselines. CFG Scale-Up applies standard CFG on Mochi with the guidance scale raised to $9.0$, and Always-Add replaces the correction of~\eqref{eq:solution} by $\delta = \delta_{\text{ref}} + \lambda(\tau,i)\,a_\tau$ at every step and frame, with $\lambda$ from~\eqref{eq:lambda} but without the threshold test, corresponding to attribute-translation style guidance. The three ``w/o'' variants each set one factor of~\eqref{eq:lambda} to a constant (Sec.~\ref{sec:ablation}). Ablation variants are based on Mochi. CLIP-early measures the similarity between late-segment frames and $p_{\text{early}}$, so lower values indicate that the early-scene prior no longer dominates the late frames. TCS denotes the Temporal Coverage Score, TVR the Temporal-suppression Violation Rate, and EPS the Early Preservation Score, all obtained from a GPT-4o vision-language judge.}
    \label{tab:tps_main}
\end{small}
\end{table}

\subsection{Quantitative Evaluation}
\label{sec:quantitative_evaluation}
We evaluate with six metrics covering overall alignment, residual early-scene dominance, and position-resolved realization. Because the question is \emph{when} content appears, sampled frames are split at the transition point $\gamma$ into early and late sets, and position-sensitive metrics use the late frames alone. CLIPScore measures alignment between all frames and $p$; \emph{CLIP-early} measures the similarity between the \emph{late} frames and $p_{\text{early}}$, where a high value means the closing frames still resemble the opening scene and lower is therefore better; BLIP caption similarity provides a text-level check that bypasses the CLIP image encoder. Since similarity cannot separate a faint emergence from an absent one, a multimodal judge additionally scores the \emph{Temporal Coverage Score (TCS)}, rating on a 1--5 scale how fully the late concept is realized within the late frames; the \emph{Temporal-suppression Violation Rate (TVR)}, counting how often it fails to appear at all; and the \emph{Early Preservation Score (EPS)}, rating whether the opening setting survives as coherent context. EPS is a guard rather than a ranking metric. It exposes the degenerate solution of discarding the early scene to render only the late concept, and is informative only alongside TCS, since a baseline that never transitions preserves the early scene by doing nothing. To limit circularity from using GPT-4o in both decomposition and judging, the judge receives a fixed rubric and the explicit late-frame window, runs at temperature 0, and issues the three judgments as independent queries; \textbf{judge-based compliance scores agree with human rankings on closely related tasks~\cite{anonymous2026negate} (full rubric and protocol in Appendix~\ref{sec:detailed_quantitative_evaluation})}. As shown in Table~\ref{tab:tps_main}, TPD attains the best value on all six metrics. The late frames leave the early scene behind (lowest CLIP-early) without loss of overall fidelity (highest CLIPScore and BLIP), roughly a third fewer clips fail to produce the requested event (TVR $0.4950 \to 0.3225$), and EPS remains at the top of its range, confirming the gains are not obtained by abandoning the early scene. The two non-projection baselines show why a directional correction is needed. Enlarging the guidance scale leaves CLIP-early essentially unchanged and even raises TVR, which indicates that the suppression is a matter of direction rather than magnitude, and adding the suppressed direction unconditionally recovers only part of the gap. \textbf{Category-wise breakdowns are in Appendix~\ref{sec:detailed_quantitative_evaluation}}.

\begin{table}[t]
\begin{small}
    \centering
    \resizebox{0.5\textwidth}{!}{%
    \begin{tabular}{lcccccc}
        \toprule
        \textbf{Method} & \textbf{SC} $\uparrow$ & \textbf{BC} $\uparrow$ & \textbf{TF} $\uparrow$ & \textbf{DD} & \textbf{AQ} $\uparrow$ & \textbf{IQ} $\uparrow$ \\
        \midrule
        Mochi            & 0.9515 & 0.9484 & 0.9864 & 0.3400 & 0.5280 & 0.5510 \\
        HunyuanVideo     & 0.9716 & 0.9668 & 0.9913 & 0.3125 & 0.5582 & 0.6196 \\
        CogVideoX        & 0.9707 & 0.9700 & 0.9885 & 0.4000 & 0.4768 & 0.4555 \\
        \midrule
        CFG Scale-Up     & 0.9523 & 0.9519 & 0.9858 & 0.3450 & 0.5504 & 0.5639 \\
        Always-Add       & 0.9548 & 0.9475 & 0.9865 & 0.3600 & 0.5246 & 0.5441 \\
        \midrule
        \textbf{Ours}    & 0.9553 & 0.9505 & 0.9875 & 0.3650 & 0.5307 & 0.5399 \\
        \midrule
        Ours w/o Frame Axis        & 0.9549 & 0.9483 & 0.9862 & 0.3275 & 0.5331 & 0.5454 \\
        Ours w/o Timestep Schedule & 0.9522 & 0.9481 & 0.9855 & 0.3750 & 0.5296 & 0.5425 \\
        Ours w/o $\rho$-Adaptive $\alpha$ & 0.9552 & 0.9507 & 0.9857 & 0.3400 & 0.5275 & 0.5495 \\
        \bottomrule
    \end{tabular}}
    \caption{\textbf{Reference-free perceptual quality (VBench).} Six VBench quality dimensions evaluated on the same generated videos in custom-input mode. SC, subject consistency; BC, background consistency; TF, temporal flickering; DD, dynamic degree; AQ, aesthetic quality; IQ, imaging quality (normalized to $[0,1]$). DD measures the share of videos with large motion and carries no preferred direction; on this benchmark, whose prompts request an emergent event, a higher DD accompanies the requested transition. SC and BC reward frame-to-frame invariance, so a baseline that never leaves the opening scene gains on them by doing nothing.}
    \label{tab:tps_vbench}
\end{small}
\end{table}

Restoring the suppressed event would mean little if it came at the cost of perceptual quality, so Table~\ref{tab:tps_vbench} additionally reports reference-free quality scores from VBench~\cite{huang2024vbench} on the same generated videos, covering six quality dimensions evaluated in custom-input mode. The consistency dimensions are the demanding ones for TPD. Subject and background consistency reward frame-to-frame invariance, which places a method that makes substantially more clips actually transition (TVR $0.4950 \to 0.3225$, Table~\ref{tab:tps_main}) at a structural disadvantage against baselines that hold the opening scene throughout. TPD nevertheless matches or exceeds Mochi on both ($0.9553$ against $0.9515$ and $0.9505$ against $0.9484$), which reflects how the correction operates. The early scene persists as context while the event develops over it, the same property EPS measures, so the transition reads as an evolution of one scene rather than a cut to another. Temporal flickering and aesthetic quality are likewise at parity, dynamic degree is higher in line with the requested emergence, and imaging quality is $0.011$ lower, within the spread of the Mochi-based rows. The correction therefore leaves perceptual quality intact while changing when content appears, which is the intended division of labor.

\begin{figure}[htb!]
\centering
\includegraphics[width=0.875\linewidth]{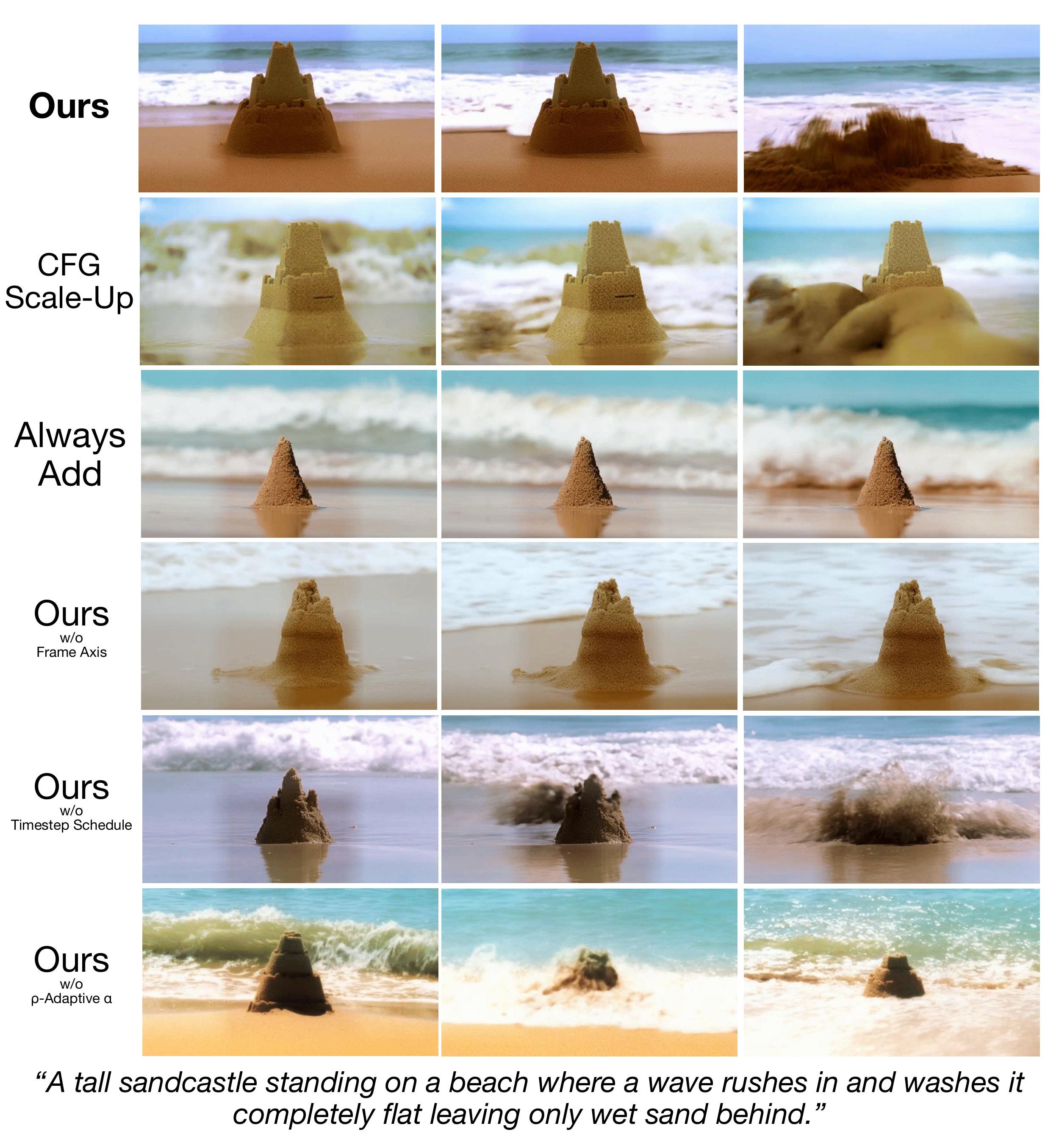}
\caption{\textbf{Ablation study} on the sandcastle prompt, frames in temporal order. \textbf{CFG Scale-Up}: the structure never falls, since scaling an update does not redirect it. \textbf{Always-Add} ($\delta = \delta_{\text{ref}} + \lambda a_\tau$, no threshold test): the castle stays intact throughout, as a fixed additive push does not overcome a strong early-scene prior. \textbf{w/o Frame Axis} ($\sigma(\cdot) \equiv 1$ in~\eqref{eq:lambda}): water already intrudes in the opening frame and a clean early state is never established, since the correction is no longer restricted to late frames. \textbf{w/o Timestep Schedule} ($\phi(\tau) \equiv 1$): the castle bursts apart while the wave is still offshore, rendering the collapse before its cause. \textbf{w/o $\rho$-Adaptive $\alpha$} ($\alpha \equiv \alpha_0$ in~\eqref{eq:alpha}): the stronger fixed threshold brings the wave onto the castle within the first third of the clip, well before the transition point, and part of the castle survives as a small mound. Only the full model produces a physically ordered emergence, in which the water first reaches the castle, then erodes its base, and only then brings it down.}
\label{results:ablation}
\end{figure}

\subsection{Ablation Study}
\label{sec:ablation}
Each ablation modifies exactly one part of the threshold schedule $\lambda(\tau,i) = \alpha \cdot \sigma(\beta(i/(F-1)-\gamma)) \cdot \phi(\tau)$ in~\eqref{eq:lambda} and leaves the rest of TPD unchanged. \emph{w/o Frame Axis} sets the frame factor $\sigma(\beta(i/(F-1)-\gamma)) \equiv 1$, so $\lambda = \alpha\,\phi(\tau)$ for every frame; \emph{w/o Timestep Schedule} sets $\phi(\tau) \equiv 1$, so $\lambda = \alpha\,\sigma(\beta(i/(F-1)-\gamma))$ at every step; and \emph{w/o $\rho$-Adaptive $\alpha$} replaces~\eqref{eq:alpha} by the constant $\alpha \equiv \alpha_0 = 1$. Since $\rho_{\text{early}}, \rho_{\text{late}} \in [0,1]$, this constant is never smaller than the adaptive $\alpha$, so the variant tests whether a uniformly applied strength can replace the per-prompt one. \emph{Always-Add} keeps $\lambda(\tau,i)$ intact but drops the threshold test of~\eqref{eq:mu}, always using $\delta = \delta_{\text{ref}} + \lambda(\tau,i)\,a_\tau$. The two variants therefore differ in magnitude as well as in the test. The added term $\lambda\,a_\tau$ has a component of $\lambda\|a_\tau\|$ along the unit vector $a_\tau/\|a_\tau\|$, whereas TPD raises that component only up to $\lambda$, so Always-Add pushes harder than TPD when $\|a_\tau\| > 1$ and less when $\|a_\tau\| < 1$. CFG Scale-Up uses no counterfactual branch and only raises $w$ in~\eqref{eq:cfg}. Results are in Table~\ref{tab:tps_main} and Figure~\ref{results:ablation}. Discarding the frame axis produces the largest CLIP-early degradation and nearly doubles its variance. The restorative signal is no longer confined to the frames it should clear, and water disturbs the structure already in the opening frame. Removing the timestep schedule leaves coverage largely intact but lowers CLIPScore and BLIP, with a specific qualitative failure in which the collapse is rendered before the wave that should cause it arrives, breaking the causal order of the event. Fixing $\alpha$ yields the largest TCS drop of any ablation, even though it applies at least as strong a threshold to every prompt, indicating that a larger correction is not uniformly better and that the per-prompt scaling matters. Always-Add raises TCS partially but leaves CLIP-early near the baseline, since a fixed increment cannot adapt to how strongly a given early scene dominates. Each factor therefore controls a different aspect of the correction (\emph{where}, \emph{when}, and \emph{how much}), and removing any one degrades a different metric. \textbf{Detailed analysis is in Appendix~\ref{sec:detailed_ablation}}.

\section{Conclusion}
We identified temporal prior suppression, in which a text-to-video model renders the opening scene of a temporally structured prompt faithfully and never produces the event that should emerge over it, and addressed it with a heuristic correction of the guidance update. The discrepancy between the full-prompt prediction and a temporal counterfactual conditioned on the early segment gives a suppressed direction, and when the alignment of the CFG update with this direction falls below a scheduled threshold, a closed-form, minimal-energy correction raises it to the threshold. Unlike prior projection methods, which remove signal, the correction adds signal, and because the threshold is scheduled jointly over denoising step and video frame, it acts mainly on the frames where the emergence belongs. On our evaluation dataset, built around emergence mechanisms rather than surface scene types, the method improves late-concept realization and reduces outright suppression while leaving the early scene intact, with the frame axis, timestep schedule, and prior-strength weighting each controlling a distinct aspect of the correction. Because the correction acts in guidance space, it makes the late concept substantially more likely to be rendered rather than guaranteeing it.

\section{Limitations and Future Works}
\label{sec:limitation_future}
TPD is a heuristic and offers no guarantee that the late-segment concept appears. The threshold of~\eqref{eq:lowerbound} is met in guidance space, but the decoded video also depends on the remaining denoising steps and on whether the backbone can render the concept, and the judge still reports a missing late concept in 32\% of the clips (TVR of $0.3225$, Table~\ref{tab:tps_main}), with causal transformation the weakest case. The schedule of~\eqref{eq:lambda} is hand-designed, and its inputs $\gamma$, $\rho_{\text{early}}$, and $\rho_{\text{late}}$ are LLM predictions not calibrated to the backbone. TPD also assumes one persisting scene and one emergent event, and our quantitative study is confined to Mochi (Appendix~\ref{supp:backbone}). \textbf{Appendix~\ref{supp:detailed_limitation_future} discusses these limitations and the extensions they suggest in detail.}

{
    \small
    \bibliographystyle{ieeenat_fullname}
    \bibliography{main}
}


\newpage
\onecolumn
\appendix

\section{Ethics Statement}
\label{supplementary:ethics}
\begin{tcolorbox}[breakable, title=Ethics Statement]
This paper presents Temporal Prior Decoupling (TPD), a training-free correction applied to the guidance signal of text-to-video diffusion models at sampling time. We record the following considerations. \textbf{Intended Use.} TPD targets a specific reliability gap. When a description asks for a new event to emerge partway through a clip, current models frequently hold the opening scene for the full duration and never produce the requested event. Restoring that suppressed content is intended to make generated video correspond more closely to what was actually asked for, which is relevant to research on controllable generation as well as to downstream settings such as storyboarding, visual prototyping, and simulation, where a mismatch between prompt and output is a functional defect. \textbf{Potential Misuse.} Any method that improves text-to-video generation can also improve the production of deceptive or harmful video. We note that TPD does not extend what a backbone is capable of synthesizing; it reallocates the guidance update so that content already specified in the prompt is more likely to be realized in its intended temporal position. The risk profile is accordingly inherited from the underlying model rather than enlarged by our correction, and mitigations appropriate to the backbone (content filtering, provenance marking, usage policies) remain applicable and necessary when TPD is used. \textbf{Dataset and Data.} Our dataset was assembled exclusively for evaluation. Its prompts describe physically realizable natural and everyday scenes, were authored without reference to identifiable individuals, and contain no offensive, sensitive, or privacy-implicating material. No human subjects were involved and no personal data was collected at any stage. We release the suite so that the reported numbers can be reproduced and so that temporal emergence can be measured consistently in subsequent work. \textbf{Use of External Language Models.} TPD's decomposition stage and our judged metrics both query a proprietary multimodal model through a commercial API. The only content transmitted is the prompts themselves and, for scoring, frames from videos generated in this study; no user data, private material, or third-party content is sent. We acknowledge that this introduces a dependency on a service whose behavior may change over time, and we therefore fix the decoding temperature to zero and record the model version used, so that the protocol is specified precisely even if the endpoint is later updated. The decomposition itself is a lightweight parsing step and could be replaced by an open model or by manual annotation without altering the method. \textbf{Computational Resources.} No training or fine-tuning is performed. Relative to standard classifier-free guidance, the additional cost per denoising step is one further forward pass of the frozen denoiser to obtain the temporal counterfactual branch; the projection itself reduces to an inner product and a norm and is negligible in comparison. All experiments ran on standard research GPU hardware, and the absence of a training stage keeps the energy cost of this work substantially below that of approaches requiring model adaptation. \textbf{Broader Impact.} Generative video systems are increasingly evaluated on appearance quality, yet a clip can look convincing while silently omitting part of what was requested. By targeting this omission with a correction of the guidance update and measuring it at the temporal position where it occurs, we hope to encourage evaluation practices that separate rendering quality from faithfulness, and to make inference-time correction a more tractable alternative to retraining when a model's distributional habits conflict with a user's intent. \textbf{Artifact Licensing and Compatibility.} Every pretrained backbone examined here was retrieved from its official distribution and used under the terms its authors declare. Mochi~1~\cite{mochi} and CogVideoX~\cite{yang2025cogvideox} carry Apache~2.0 licenses, which permit research use, modification, and redistribution without restriction. HunyuanVideo~\cite{kong2024hunyuanvideo} is distributed under the Tencent Hunyuan Community License, which allows academic research within the licensed territory; the experiments described in this paper fall wholly inside that scope. The perception models used by our embedding metrics, CLIP and BLIP, are released under MIT and BSD-3-Clause respectively, and both permit the inference-only research use made of them here. At no point are weights retrained, adapted, altered, or redistributed, and each backbone is invoked strictly as a frozen sampler, which is what allows our usage to remain consistent with the research-oriented terms attached to all of these artifacts.
\end{tcolorbox}

\section{Detailed Limitations and Future Works}
\label{supp:detailed_limitation_future}
TPD is a heuristic and offers no guarantee that the late-segment concept appears. The threshold in~\eqref{eq:lowerbound} is met in guidance space at each step, but the decoded video also depends on the remaining denoising steps and on whether the backbone can render the concept, and the judge still reports a missing late concept in 32\% of the clips (TVR of $0.3225$, Table~\ref{tab:tps_main}). The effect is weakest on causal transformation, where an intact structure must be destroyed, since a guidance-space intervention reweights the backbone's own predictions and cannot create a trajectory to which the backbone assigns negligible probability. The threshold schedule in~\eqref{eq:lambda} is hand-designed, and its inputs $\gamma$, $\rho_{\text{early}}$, and $\rho_{\text{late}}$ are predicted by an LLM. These predictions have no ground truth, and $\rho$ in particular reflects the LLM's judgment of how common each segment is rather than a measurement of the backbone, so this stage is not calibrated to the backbone. The frame factor is smooth and the strength depends only on the difference between the two estimates, which limits the effect of small errors, but a poor estimate still leads to under- or over-correction. Learning these parameters with preference optimization such as DPO or GRPO would be more principled, but it requires a reward signal and an optimization stage, so the method would no longer be training-free. Estimating them from signals of the backbone itself, such as the relative magnitudes of $a_\tau$ and $\delta_{\text{ref}}$, is a training-free alternative that we leave to future work. TPD also assumes a two-part description with one persisting scene and one emergent event, so prompts specifying several ordered or overlapping events fall outside its present scope, and our quantitative study is confined to Mochi (Appendix~\ref{supp:backbone}). Future work may combine upper-bound suppression, drift removal, and threshold-based restoration in a single correction, and extend the schedule to multi-event descriptions in which each emergent concept has its own temporal window.

\section{Detailed Benchmarking Dataset}
\label{supp:detailed_dataset}

Whereas conventional video benchmarks score whether a generated clip matches a static description, our goal is to measure something the standard protocols never test, namely whether a requested concept is realized \emph{at the temporal position where it should emerge}. This shift from holistic clip--text agreement to position-resolved emergence changes what a dataset must contain and motivates a dedicated suite.

\paragraph{Why Standard Datasets Are Insufficient.}
Widely used corpora such as MSR-VTT, WebVid, or MS-COCO pair a single caption with a single visual state and were assembled to support retrieval and captioning. Descriptions that specify a within-clip transition, in which an early scene persists while a new event arrives over it, are essentially absent and are never organized in a controlled way. In addition, these benchmarks quantify \emph{aggregate} embedding agreement between a clip and its caption, which is blind to \emph{when} content appears, since a video that renders only the early scene throughout can still score highly against a caption that mentions the later event. Attribute-transition suites such as CAT-Bench~\cite{lo2025prompt} do introduce change over time, but their prompts are built around a persistent subject whose single attribute slides from one value to another (e.g., young to old), and therefore do not probe the appearance of a new entity or event over a continuing scene. Formally, retrieval-style evaluation reduces to a similarity ordering
\[
S(V, T_{\text{match}}) > S(V, T_{\text{mismatch}}),
\]
whereas the quantity we care about is the probability that the late-segment concept is actually generated in its intended temporal window,
\begin{equation}
\mathbb{P}_{x_0 \sim p_\theta(\cdot \mid p)}
\!\left[\, \mathcal{R}_{\text{late}}(x_0; p) = 1 \,\right],
\label{eq:tps_objective}
\end{equation}
where $\mathcal{R}_{\text{late}}(x_0; p)$ indicates that the emergent concept specified by the late segment of $p$ is present in the corresponding frames. No existing benchmark supplies prompts structured to estimate Eq.~\eqref{eq:tps_objective}, so we assemble a controlled suite for this purpose.

\paragraph{Dataset Scale.}
Our dataset comprises eight categories of 50 prompts each, for a total of 400 single-prompt evaluation samples. Every prompt describes one continuous scene in which an early state should hold and a later concept should emerge within the same clip. The suite is used exclusively for evaluation; no training or fine-tuning is performed.

\paragraph{Temporal Emergence Formulation.}
Each prompt $p$ is parsed into an early segment $p_{\text{early}}$ and a late segment $p_{\text{late}}$. A faithful generation must satisfy two conditions jointly. The early scene is preserved in the opening frames, and the late concept is realized in the closing frames. Writing $\mathcal{P}_{\text{early}}$ and $\mathcal{R}_{\text{late}}$ for these two events,
\begin{equation}
\mathcal{T}(x_0; p) =
\mathbf{1}\!\left[\mathcal{P}_{\text{early}}(x_0; p_{\text{early}})\right]
\cdot
\mathbf{1}\!\left[\mathcal{R}_{\text{late}}(x_0; p_{\text{late}})\right].
\end{equation}
Temporal prior suppression corresponds to the case in which the early scene is preserved but the late concept never appears,
\begin{equation}
\mathcal{P}_{\text{early}}(x_0; p_{\text{early}}) = 1,
\qquad
\mathcal{R}_{\text{late}}(x_0; p_{\text{late}}) = 0,
\end{equation}
i.e., the trajectory remains anchored to the dominant early-scene prior across the whole clip. Each category is designed to drive the model into exactly this regime, so that the gap between $\mathcal{P}_{\text{early}}$ and $\mathcal{R}_{\text{late}}$ exposes the failure.

\paragraph{Design Principles.}
All prompts follow four rules:
\begin{itemize}
    \item \textbf{Scene Persistence:} the early scene is a coherent setting that should remain visible, not be replaced, when the late concept arrives.
    \item \textbf{Genuine Emergence:} the late segment introduces a new entity, phenomenon, or state change rather than a single-attribute morph of the early scene.
    \item \textbf{Mechanism Isolation:} each category targets one mode by which new content enters a scene, so failures can be attributed to a specific emergence type.
    \item \textbf{Prior Dominance:} the early scene is chosen to be a strong, high-frequency configuration, making it a natural attractor that competes against the late concept.
\end{itemize}

The eight categories are defined at the level of surface scene type and group into four higher-level emergence mechanisms, namely \emph{entity entry} (ENTRY, ACTIVITY), \emph{gathering} (SCENE), \emph{phenomenon onset} (ONSET, WEATHER\_NIGHT), and \emph{causal transformation} (STATE, CAUSAL, STRUCTURAL). The grouping is intentionally non-uniform. We prioritize coverage of distinct scene types that elicit emergence, together with a clean separation from single-attribute transition settings, over an even count per mechanism. Causal transformation is represented by more categories because it is the regime in which the early-scene prior is strongest and emergence is hardest to elicit, making it the most informative for diagnosing suppression.

\subsubsection*{(1) ENTRY: Single-Subject Entry}

A still, unpopulated natural setting into which a discrete animal subject (or small group) enters and acts:
\begin{quote}
``A quiet mountain lake at dawn with still water and empty shoreline where a herd of deer walks down to the bank and begins drinking.'' \\
``A bare snowy riverbank at dawn where a group of elk descends the slope and wades into the shallows.'' \\
``A silent wetland at dawn with flat water and empty reeds where a flock of herons glides in and lands along the bank.''
\end{quote}
The early scene establishes an empty environment, and the late concept is the arrival of living subjects. Because the empty natural vista is itself a highly probable completion, the model tends to keep the shoreline empty for the full clip.

\subsubsection*{(2) SCENE: Crowd Gathering}

A vacant human venue at an off-hour into which people and activity progressively accumulate:
\begin{quote}
``An empty city plaza at dawn where crowds of people gradually fill the streets and market stalls appear.'' \\
``A lonely train station platform before dawn where a train arrives and dozens of commuters flood onto the platform.'' \\
``A quiet airport gate area before boarding where passengers stream in and the area becomes packed with travelers.''
\end{quote}
The constraint is a transition from emptiness to density. The deserted-venue prior is strong, so generations frequently leave the space sparsely populated rather than realizing the crowd.

\subsubsection*{(3) ONSET: Phenomenon Onset}

A calm landscape over which a large-scale natural phenomenon develops and sweeps across the frame:
\begin{quote}
``A dry desert basin under a pale sky where dust begins to swirl and a towering sandstorm sweeps across the plain.'' \\
``A quiet dry riverbed under a pale blue sky where a flash flood surge rushes in from upstream and fills the channel.'' \\
``A still open rangeland under a pale sky where a tornado funnel descends from a rotating cloud and touches the ground.''
\end{quote}
The early scene is a quiescent environment and the late concept is a transient meteorological or geophysical event. The placid-landscape prior dominates, and the dramatic onset is often omitted.

\subsubsection*{(4) STATE: Surface State Change}

A clean, undisturbed surface that is altered by an external action, leaving a visible change:
\begin{quote}
``A pristine white snow field under clear skies where a flock of dark birds flies across the field in moving black shapes.'' \\
``A still pond with a glass-smooth surface where a stone drops in and ripples spread outward across the entire surface.'' \\
``A blank white canvas on an easel where a painter begins applying bold strokes of red and blue paint.''
\end{quote}
The early scene is a pristine state and the late concept is its disturbance. The smooth, untouched configuration is a strong prior, so the disturbance frequently fails to register.

\subsubsection*{(5) CAUSAL: Ice Fracture}

A frozen surface in winter stillness that fractures and breaks apart under a causal trigger:
\begin{quote}
``A still sea covered with smooth ice in winter light where cracks spread across the surface and the ice breaks apart into drifting shards.'' \\
``A smooth sheet of ice covering a city pond where a stone is thrown and a web of cracks radiates outward from the impact.'' \\
``A flat ice sheet on a bay in winter stillness where pressure ridges begin to form and slabs buckle upward along a fault line.''
\end{quote}
This category isolates causal physical change, in which an intact frozen surface must visibly fracture. The intact-ice prior is overwhelmingly dominant, making the fracture among the hardest emergent events to elicit.

\subsubsection*{(6) WEATHER\_NIGHT: Nocturnal Weather Onset}

A tranquil nighttime scene under stars or moonlight over which a violent weather event breaks:
\begin{quote}
``A calm bay with boats anchored under a clear night sky where a storm rolls in and lightning illuminates the churning water.'' \\
``A still moonlit field where a thunderstorm rolls in from the horizon and lightning strikes the open ground.'' \\
``A calm starlit mountain valley where a blizzard descends and the valley disappears in swirling snow.''
\end{quote}
The night setting persists while the storm emerges. This category pairs phenomenon onset with a low-light context that further biases the model toward the calm initial state, compounding suppression of the late event.

\subsubsection*{(7) ACTIVITY: Traffic Inflow}

An empty transit setting into which vehicles or moving participants progressively appear and fill the space:
\begin{quote}
``A completely empty highway at dawn where cars begin appearing and the road fills with flowing traffic.'' \\
``A still airport runway before dawn where planes begin taxiing in from the gates and line up for departure.'' \\
``A quiet mountain road at sunrise where a cycling peloton appears over the crest and streams downhill.''
\end{quote}
The constraint is a transition from a clear route to one filled with directed motion. The empty-thoroughfare prior is strong, and generations often keep the route deserted.

\subsubsection*{(8) STRUCTURAL: Structural Collapse}

An intact built or assembled structure that is destabilized and collapses into a disordered state:
\begin{quote}
``A tall sandcastle standing on a beach where a wave rushes in and washes it completely flat leaving only wet sand behind.'' \\
``A row of dominoes standing on a table that begins to fall from one end and cascades to the other.'' \\
``A completed Jenga tower on a table that a player pulls the wrong block from and brings down.''
\end{quote}
The early scene is an ordered structure and the late concept is its destruction. The intact-structure prior is strong, so the collapse, which requires the model to abandon a stable, coherent object, is frequently not realized.

\paragraph{Summary.}
ENTRY, SCENE, ONSET, STATE, CAUSAL, WEATHER\_NIGHT, ACTIVITY, and STRUCTURAL together span the principal ways a new concept can arise within a continuing scene, namely a subject entering, a crowd gathering, a phenomenon developing, a surface being altered, a structure fracturing, weather breaking at night, traffic flowing in, and a structure collapsing. Each category fixes a persistent early scene and a distinct emergent late concept, so that performance can be read as the degree to which the late concept is realized in its temporal window rather than suppressed by the early-scene prior. By organizing prompts around emergence mechanisms and scoring position-resolved realization as in Eq.~\eqref{eq:tps_objective}, our dataset provides a targeted benchmark for diagnosing temporal prior suppression that existing video benchmarks do not offer.

\begin{figure*}[t] 
    \centering 
    \includegraphics[width=0.7\linewidth]{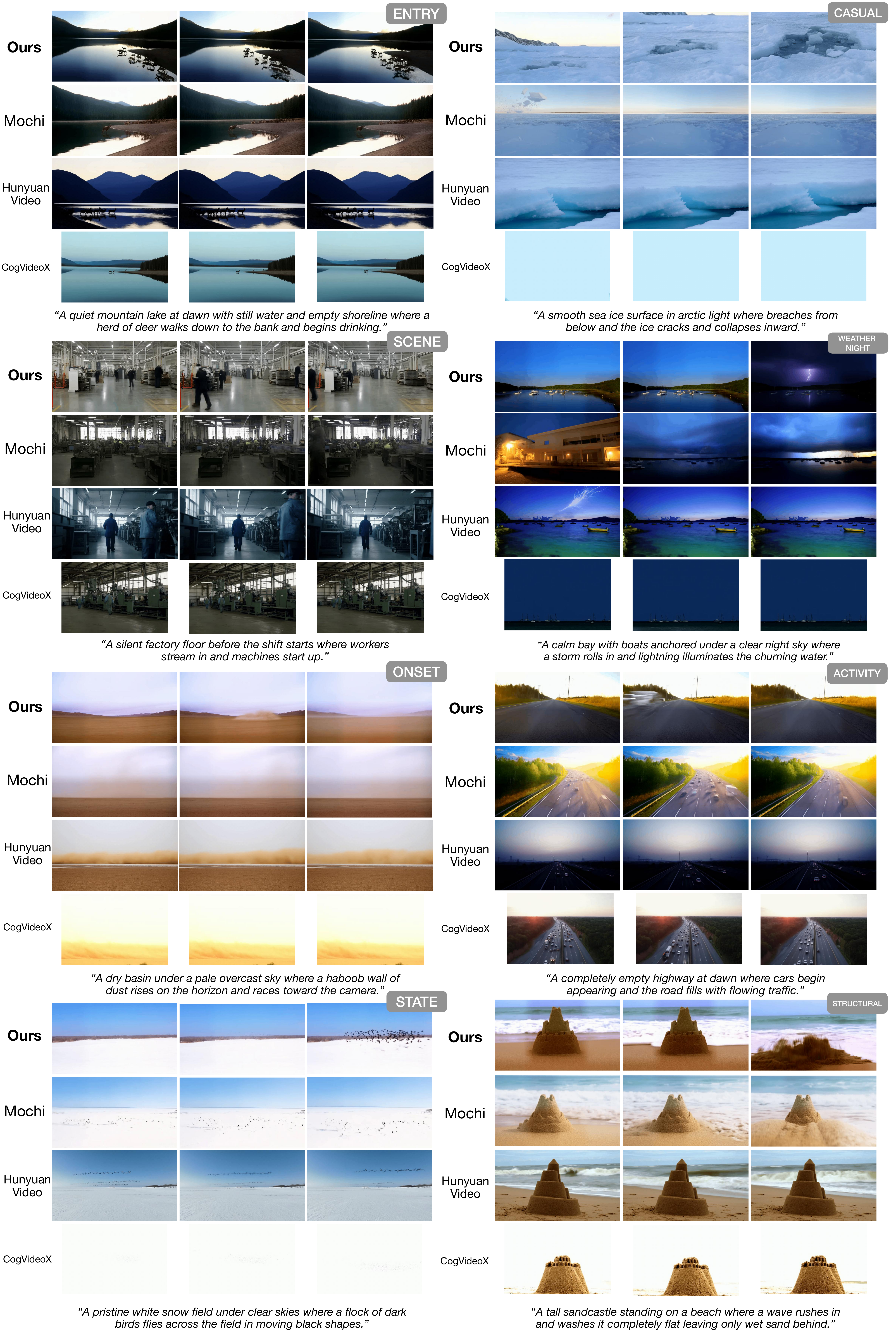} 
    \caption{\textbf{Qualitative comparison across categories (Set 1).} We evaluate our method (TPD) against state-of-the-art video diffusion models (Mochi, HunyuanVideo, CogVideoX) on eight temporal emergence categories (ENTRY, SCENE, ONSET, STATE, CAUSAL, WEATHER\_NIGHT, ACTIVITY, and STRUCTURAL). Each row shows three frames sampled from generated videos for a representative prompt per category. Baseline models consistently fail to realize the late-segment concept, for instance keeping a shoreline empty instead of showing arriving deer, or preserving an intact sandcastle instead of letting the wave reach it, whereas in these examples TPD realizes both the early scene and the emergent late concept in their respective temporal positions.} 
    \label{fig:detailed_qualitative_tps_supp1} 
\end{figure*}

\begin{figure*}[t]
    \centering
    \includegraphics[width=0.7\linewidth]{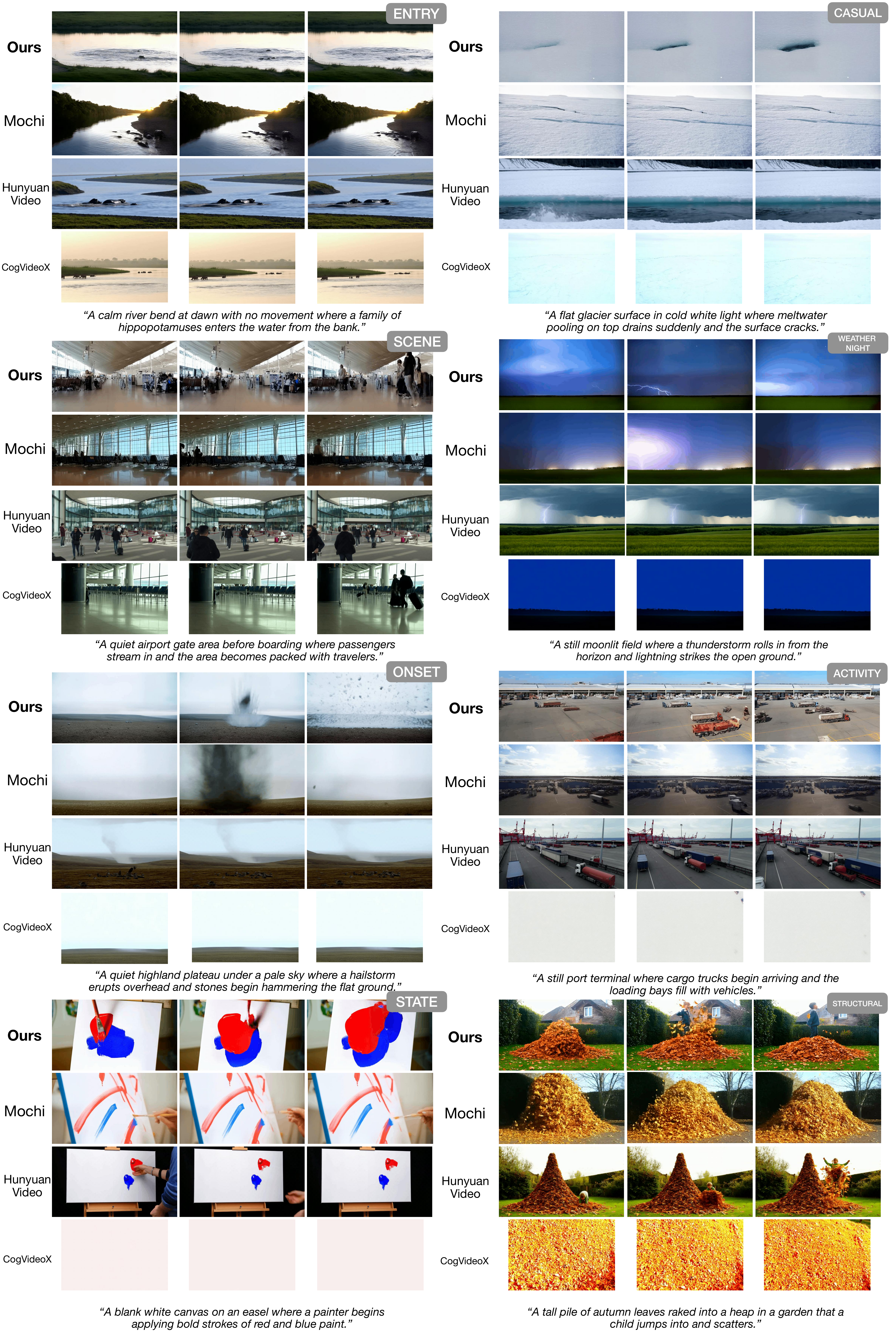}
    \caption{\textbf{Qualitative comparison across categories (Set 2).} This figure presents a second set of representative prompts covering all eight temporal emergence categories (ENTRY, SCENE, ONSET, STATE, CAUSAL, WEATHER\_NIGHT, ACTIVITY, STRUCTURAL). These examples include more challenging emergence scenarios such as \emph{a family of hippopotamuses entering a calm river bend}, \emph{a hailstorm erupting over a highland plateau}, and \emph{a child scattering a pile of autumn leaves}. While baseline models frequently anchor the entire clip to the dominant early-scene prior and fail to realize the emergent event, TPD restores the suppressed late-segment signal in its intended temporal window while preserving early-scene coherence.}
    \label{fig:detailed_qualitative_tps_supp2}
\end{figure*}

\begin{figure*}[t]
    \centering
    \includegraphics[width=0.7\linewidth]{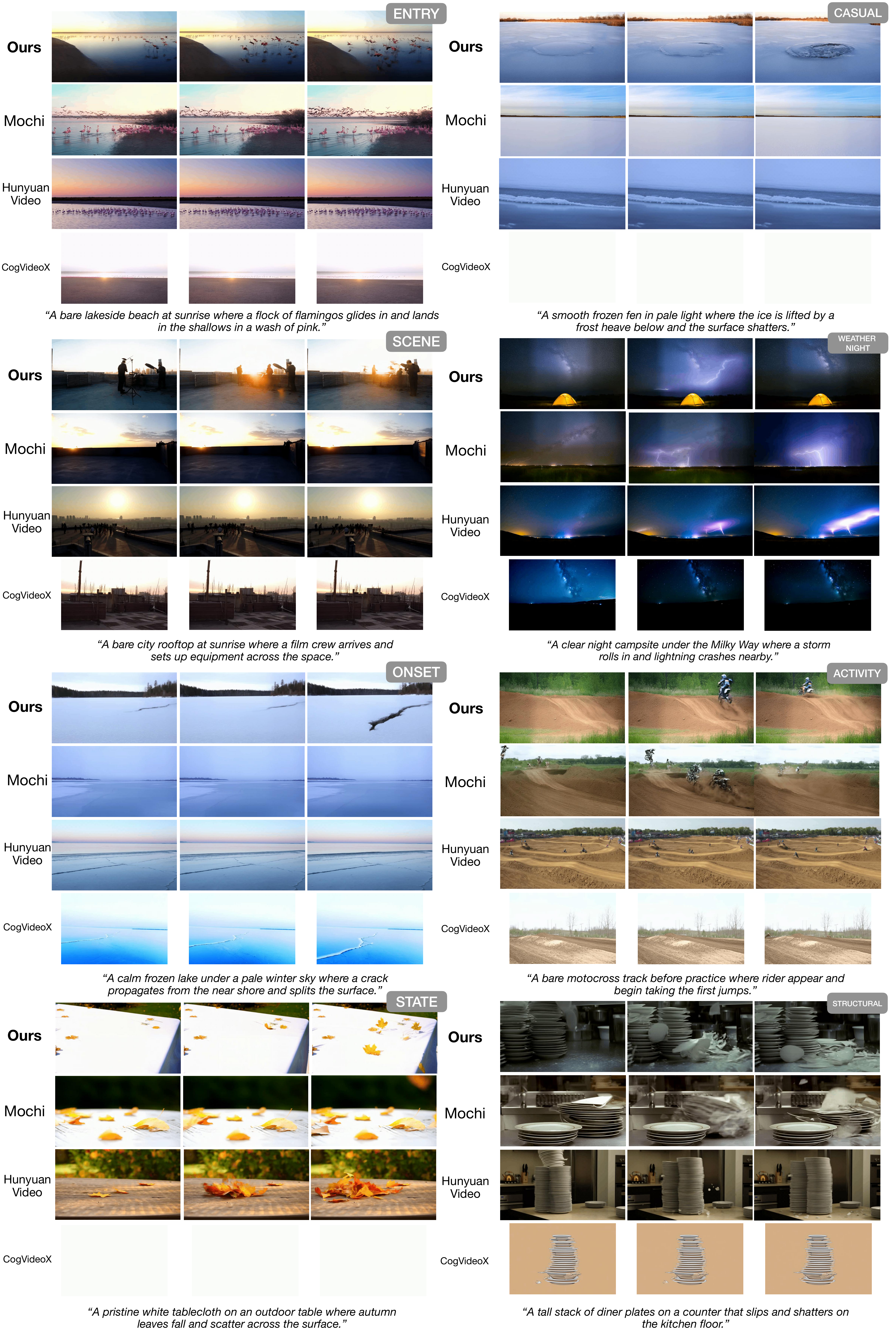}
    \caption{\textbf{Qualitative comparison across categories (Set 3).} This figure presents a third set of prompts spanning all eight temporal emergence categories, including \emph{a flock of flamingos landing in a bare lakeside beach}, \emph{a lightning storm rolling into a Milky Way campsite}, and \emph{a tall stack of diner plates shattering on the kitchen floor}. These cases probe stronger early-scene priors (frozen surfaces, bare natural settings, and tranquil nocturnal scenes) that exhibit particularly pronounced temporal prior suppression. Across all settings, baseline models exhibit trajectory anchoring toward the high-prior early scene, whereas TPD strengthens the late-segment signal in the guidance update and realizes the late concept more often without disrupting early-segment fidelity.}
    \label{fig:detailed_qualitative_tps_supp3}
\end{figure*}

\section{Detailed Qualitative Results}
\label{sec:detailed_qualitative}

Figures~\ref{fig:detailed_qualitative_tps_supp1}, \ref{fig:detailed_qualitative_tps_supp2}, and~\ref{fig:detailed_qualitative_tps_supp3} collect three independent sets of prompts, each covering all eight categories. Every row displays three frames sampled at equal intervals along the clip, so the leftmost frame corresponds to the early segment and the rightmost falls inside the window where the late concept is expected to appear; reading a row from left to right therefore traces whether the requested emergence ever occurs. TPD is applied to Mochi and compared against Mochi, HunyuanVideo, and CogVideoX under their default sampling configurations. Since our dataset is organized by the mechanism through which new content enters a scene, we discuss the results along those four mechanisms rather than by surface scene type.

\paragraph{Entity Entry and Gathering.} In ENTRY, ACTIVITY, and SCENE the early segment is defined by \emph{absence} (an unoccupied shoreline, a clear roadway, a venue before opening), and the late segment asks for that emptiness to be populated. This makes the early state an unusually attractive completion, because a deserted landscape is already a self-consistent image that requires no further change. The baselines default to this completion. For \emph{a quiet mountain lake at dawn with still water and empty shoreline where a herd of deer walks down to the bank} (Set 1), the water and bank remain undisturbed from the first frame to the last, and the clip reads as a static landscape rather than an arrival. The same holds for \emph{a silent factory floor before the shift starts} (Set 1), where the hall stays dim and unstaffed, and for \emph{a quiet airport gate area before boarding} (Set 2), where the concourse never reaches the density the prompt asks for. TPD instead introduces the subjects within the late frames (deer at the waterline, workers moving between machines, travelers filling the gate), while the opening frame remains close to the corresponding baseline output, which is the behavior the frame-dependent threshold is designed to produce.

\paragraph{Phenomenon Onset.} ONSET and WEATHER\_NIGHT require a transient large-scale event to develop over a quiescent landscape. Here the baselines often render the early scene with considerable fidelity, and the event never arrives. For \emph{a calm bay with boats anchored under a clear night sky where a storm rolls in and lightning illuminates the churning water} (Set 1), HunyuanVideo produces an attractive moonlit bay with anchored boats that persists unchanged, so the early segment is rendered accurately while the storm requested in the late segment never appears. Comparable behavior appears for \emph{a dry basin under a pale overcast sky where a haboob wall of dust rises on the horizon} (Set 1) and \emph{a quiet highland plateau under a pale sky where a hailstorm erupts overhead} (Set 2). Nocturnal prompts compound the difficulty, since an emergent event must be realized at low contrast against a dark background; TPD nevertheless produces visible lightning and storm structure in the late frames of both Set 1 and Set 3 while retaining the night setting, rather than brightening the whole clip to accommodate the event.

\paragraph{Causal Transformation.} STATE, CAUSAL, and STRUCTURAL are the most demanding, because realizing the late concept requires destroying an object or surface that is already stable and coherent. The representative case is \emph{a tall sandcastle standing on a beach where a wave rushes in and washes it completely flat} (Set 1). All three baselines carry an intact sandcastle through to the final frame, in one case even rendering an approaching wave in the background without allowing it to reach the structure, whereas TPD keeps the castle intact until the water reaches it and then shows it eroding at the base and collapsing into a low heap of sand. Fracture prompts behave similarly. For \emph{a smooth sea ice surface in arctic light where the ice cracks and collapses inward} (Set 1) and \emph{a smooth frozen fen where the ice is lifted by a frost heave below and the surface shatters} (Set 3), the baselines hold a continuous white sheet, while TPD opens dark leads and displaced fragments in the late frames. \emph{A tall stack of diner plates on a counter that slips and shatters} (Set 3) and \emph{a tall pile of autumn leaves that a child jumps into and scatters} (Set 2) follow the same pattern. That this group is the hardest is consistent with the category-wise scores in Table~\ref{tab:tps_category}, where CAUSAL records the lowest TCS and the highest TVR of any category for every method.

\paragraph{Backbone-Specific Degeneracies.} The three backbones fail in qualitatively different ways, which is informative for how the quantitative results should be read. CogVideoX frequently degenerates into near-uniform or washed-out frames, visible for the snow field and sea ice prompts in Set 1, the port terminal and blank canvas prompts in Set 2, and the tablecloth prompt in Set 3, losing not only the emergent event but the early scene as well. This is the source of its markedly low EPS and its highest TVR in Table~\ref{tab:tps_main}. HunyuanVideo shows the opposite profile. Its early scenes are rendered with strong visual quality, yet the clip remains anchored to them, which is why it can score competitively on appearance-oriented metrics while still trailing on TCS. Visual fidelity and temporal faithfulness are therefore separable properties, and only the latter is what our dataset is built to measure.

\paragraph{Overall Observations.} Two patterns recur across the three sets. First, baseline clips are typically \emph{correct at the beginning and unchanged thereafter}, as the opening frame satisfies the early segment, and the trajectory then changes little. This explains why early-scene preservation alone does not separate the methods, since most baselines preserve the early scene because they do nothing else, and why the informative contrast appears in TCS and TVR, with EPS serving to confirm that improvements are not obtained by discarding the early setting. Second, TPD's divergence from the baselines is concentrated in the later frames, matching the restorative correction being gated by both the frame position and the denoising step; the corresponding drop in CLIP-early in Table~\ref{tab:tps_main} is the quantitative counterpart of this visual behavior. Taken together, the qualitative evidence indicates that temporal prior suppression is not a rendering deficiency that scale alone resolves, but a directional imbalance in the guidance signal that can be substantially reduced at inference time by strengthening the suppressed component.

\section{Detailed Quantitative Evaluation}
\label{sec:detailed_quantitative_evaluation}

Our evaluation asks a question that aggregate video--text agreement cannot answer, namely not only whether a clip matches its prompt overall but also whether the requested content appears \emph{at the temporal position where it was requested}. A clip that renders only the early scene for its entire duration can still align reasonably well with a caption that mentions the later event, so a protocol that averages over all frames is blind to the failure we study. We therefore adopt six metrics, three in embedding space and three from a vision-language judge, and make the temporal partition explicit wherever the position of content is what is being measured.

\paragraph{Temporal Partition of Frames.}
For a clip of $T$ uniformly sampled frames $\{I_t\}_{t=1}^{T}$ generated from prompt $p$, let $\gamma \in [0,1]$ be the transition point produced by the decomposition stage (Sec.~\ref{subsec:decomposition}). We split the sampled frames into an early set and a late set according to normalized position,
\[
\mathcal{E} = \Big\{ t \;\Big|\; \tfrac{t-1}{T-1} < \gamma \Big\},
\qquad
\mathcal{L} = \Big\{ t \;\Big|\; \tfrac{t-1}{T-1} \ge \gamma \Big\},
\]
with the convention that both sets are non-empty. Metrics concerned with \emph{where} content appears are computed on $\mathcal{L}$, while metrics concerned with overall agreement use all $T$ frames. The same $\gamma$ drives generation and evaluation, so no additional annotation is introduced at scoring time.

\paragraph{CLIPScore.}
Global agreement between the clip and the complete prompt is measured as the mean frame-wise cosine similarity in CLIP space,
\[
\text{CLIPScore}
= \frac{1}{T} \sum_{t=1}^{T}
\cos\!\big(E_{\text{img}}(I_t),\, E_{\text{text}}(p)\big).
\]
This quantity is insensitive to temporal ordering by construction; we report it to verify that restoring the late concept does not come at the expense of overall prompt fidelity, not as evidence of emergence.

\paragraph{Residual Early-Scene Dominance (CLIP-early).}
To quantify how strongly the early-scene prior persists into the frames it should have vacated, we compare only the late frames against the early segment $p_{\text{early}}$:
\[
\text{CLIP-early}
= \frac{1}{|\mathcal{L}|} \sum_{t \in \mathcal{L}}
\cos\!\big(E_{\text{img}}(I_t),\, E_{\text{text}}(p_{\text{early}})\big).
\]
A high value means the closing frames still resemble the opening scene, which is the signature of temporal prior suppression; lower is better. Restricting the sum to $\mathcal{L}$ is essential, because averaged over the whole clip the statistic would be dominated by frames in which resembling $p_{\text{early}}$ is the correct behavior. Because $p_{\text{early}}$ is already available as the conditioning of the counterfactual branch, this metric probes the same quantity the method acts on, viewed from the output side.

\paragraph{BLIP Caption Agreement.}
As an independent check that does not route through CLIP's image encoder, we caption each frame with a pretrained BLIP model and compare the caption text to the prompt text:
\[
\text{BLIP}
= \frac{1}{T} \sum_{t=1}^{T}
\cos\!\big(E_{\text{text}}(\hat{C}_t),\, E_{\text{text}}(p)\big).
\]
Since the comparison is text-to-text, agreement here reflects whether an external captioner describes the frames in terms that match the prompt, rather than whether the frames lie near the prompt embedding.

These three quantities are complementary but share a limitation. All are similarity scores, and similarity does not distinguish a clip in which the emergent event is faint from one in which it is absent but the scene is otherwise well rendered. We therefore complement them with judged measurements.

\paragraph{Vision-Language Judging of Temporal Emergence.}
Faithful emergence requires two conditions to hold simultaneously, namely that the early scene is preserved as context and that the late concept is realized in its window, which is the criterion $\mathcal{T}(x_0;p) = \mathbf{1}[\mathcal{P}_{\text{early}}]\cdot\mathbf{1}[\mathcal{R}_{\text{late}}]$ formalized in Appendix~\ref{supp:detailed_dataset}. Embedding scores collapse both conditions into a single number and can be satisfied by degenerate solutions, most notably by abandoning the early scene entirely and rendering only the late concept. We therefore query a multimodal judge for each condition separately.

\textbf{Judge input.} The judge receives (i) the early segment $p_{\text{early}}$, (ii) the late segment $p_{\text{late}}$, (iii) frames sampled at uniform intervals in temporal order, and (iv) an explicit statement of which frame indices constitute the early portion and which constitute the late portion under the partition above. Supplying the partition makes the judgment position-aware. Without it the judge can only report whether the late concept appears somewhere, which is not the property in question. Frames are resized with aspect ratio preserved, and no cropping or masking is applied.

\textbf{Judge configuration.} We use GPT-4o through the OpenAI API with deterministic decoding (temperature $=0$) and a fixed rubric supplied verbatim on every call, so that scores are comparable across methods and categories. Each of the three judgments below is issued as a separate query with its own rubric, which keeps the criteria independent rather than letting one score anchor another.

\paragraph{Temporal Coverage Score (TCS).}
The judge assigns an integer $s_i \in \{1,\dots,5\}$ recording how fully the late-segment concept is realized within the late frames:
\begin{itemize}
    \item 1: the late concept never appears; the clip remains on the early scene throughout.
    \item 2: only a faint trace of the late concept is discernible.
    \item 3: the late concept is partially or ambiguously present in the late frames.
    \item 4: the late concept emerges, with minor incompleteness or weak intensity.
    \item 5: the late concept emerges clearly and completely, as described.
\end{itemize}
Averaging over the $N$ evaluated clips gives
\[
\text{TCS} = \frac{1}{N}\sum_{i=1}^{N} s_i ,
\]
with higher values indicating stronger late-concept realization. TCS plays the role that compliance scoring plays in related constrained-generation evaluations~\cite{anonymous2026negate, anonymous2026dcr}, redirected from concept presence to concept \emph{placement}.

\paragraph{Temporal-Suppression Violation Rate (TVR).}
Separately, the judge emits a binary indicator of outright suppression, set when the clip stays anchored to the early scene and the late concept never meaningfully materializes:
\[
v_i =
\begin{cases}
1 & \text{if the late concept is judged absent},\\
0 & \text{otherwise},
\end{cases}
\qquad
\text{TVR} = \frac{1}{N}\sum_{i=1}^{N} v_i .
\]
Where TCS grades the degree of realization, TVR counts how often realization fails altogether, giving the incidence of the failure mode rather than its average severity. Lower is better.

\paragraph{Early Preservation Score (EPS).}
The third judgment scores the complementary condition, whether the opening setting survives as coherent context once the late concept arrives, on the same five-point scale, from 1 (the early scene is discarded or replaced outright) to 5 (the early scene remains recognizable and coherent throughout):
\[
\text{EPS} = \frac{1}{N}\sum_{i=1}^{N} e_i .
\]
EPS is not intended to rank methods. Its function is to close a loophole. A method could drive TCS up and TVR down by simply overwriting the scene with the late concept, which would satisfy every other metric while violating the definition of emergence. EPS makes that shortcut visible, and should be read alongside TCS rather than on its own. In particular, an untreated baseline that never transitions will score well on EPS because it does nothing, so a high EPS is informative only when paired with a high TCS.

\textbf{Aggregation.} All scores are averaged over the complete set of prompts with no thresholding, exclusion, or post-hoc adjustment; ambiguous mid-scale judgments are retained as issued. Because TVR is a per-clip binary quantity, its dispersion follows the Bernoulli form, and the standard deviations reported in Table~\ref{tab:tps_supp} should be interpreted accordingly.

\paragraph{Overall Performance.}
Table~\ref{tab:tps_supp} reports every metric with standard deviations. TPD attains the best value on all six. The embedding results show that the improvement is directional rather than merely stronger conditioning. CLIP-early falls from $0.2865$ (Mochi) to $0.2580$, indicating that the closing frames leave the early scene behind, while CLIPScore rises from $0.3017$ to $0.3109$ and BLIP from $0.6883$ to $0.7140$, so overall fidelity is not traded away to obtain it. The judged metrics move considerably further. TCS increases from $3.1850$ to $3.8775$ and TVR drops from $0.4950$ to $0.3225$, meaning roughly a third fewer clips fail to produce the requested event at all.

The two non-projection baselines isolate why this works. Raising the guidance scale leaves the failure essentially untouched. CFG Scale-Up changes CLIP-early by less than $0.0003$ and in fact records a \emph{higher} TVR ($0.5275$) than plain Mochi, confirming that a larger guidance scale amplifies the early-scene-dominated update without changing its direction. Always-Add, which retains our two-dimensional schedule but adds $\lambda(\tau,i)\,a_\tau$ unconditionally instead of applying the threshold test, does help, raising TCS to $3.3325$ and lowering CLIP-early to $0.2850$. It nonetheless trails the full method by $0.545$ TCS and $0.1675$ TVR, and its CLIP-early remains close to the untreated baseline. Adding a fixed amount of the suppressed direction is thus a real but partial remedy, and the adaptive threshold, which intervenes only where the alignment falls short and by the smallest amount that reaches the threshold, accounts for the remaining gap.

The ablations separate the roles of the three factors of~\eqref{eq:lambda}. Removing the $\rho$-adaptive weighting produces the largest TCS degradation ($3.8775 \to 3.2600$), since that factor governs \emph{how much} correction a prompt receives; the fixed $\alpha = \alpha_0$ is at least as large as the adaptive value for every prompt, so the drop reflects the loss of per-prompt scaling rather than a weaker correction. Removing the frame axis instead produces the largest CLIP-early degradation ($0.2580 \to 0.2691$, against $0.2588$ and $0.2592$ for the other two ablations), and its standard deviation nearly doubles ($0.0759$ versus roughly $0.039$ elsewhere). This matches the component's role, since the frame factor determines \emph{where} the correction is applied, so discarding it spreads the restorative signal across the whole clip and the late frames are no longer specifically cleared of the early prior, with the effect varying substantially from prompt to prompt. CLIP-early, computed on the late frames, is where this contribution is measurable.

Finally, EPS behaves as intended for a guard metric. All Mochi-based variants cluster between $4.7500$ and $4.8800$, so it does not separate them, and the one method it does flag is CogVideoX at $3.7475$ with a standard deviation of $1.8199$, a backbone that, as the qualitative results show, frequently loses the scene altogether rather than transitioning within it. Every projection variant remains near the ceiling, so the gains in TCS and TVR were not obtained by sacrificing the early scene. HunyuanVideo attains a higher BLIP score than Mochi ($0.7098$ versus $0.6883$) while recording a lower TCS ($2.8250$ versus $3.1850$), consistent with its profile of rendering the opening scene well and then holding it (Appendix~\ref{sec:detailed_qualitative}).

\paragraph{Category-wise Analysis.}
Table~\ref{tab:tps_category} breaks the judged metrics down by emergence category, comparing Mochi, Always-Add, and TPD on a common backbone. TPD records the highest TCS in all eight categories, the lowest TVR in seven, and the highest EPS in seven.

The largest gains occur where the early scene is a strongly self-consistent image that the model has no incentive to disturb. ONSET improves most sharply, with TCS rising from $3.26$ to $4.60$ and TVR falling from $0.54$ to $0.10$, while Always-Add is \emph{worse} than the untreated baseline in this category on both metrics ($3.22$ and $0.58$), so a fixed additive push is not simply a weaker version of the correction but can actively destabilize prompts where a transient phenomenon must be introduced without disrupting a quiescent landscape. SCENE and ENTRY follow with TCS gains of $+0.98$ and $+0.70$ and TVR reductions of $0.30$ and $0.06$; in ENTRY, TPD reaches an EPS of $5.00$, meaning subjects were introduced without perturbing the setting at all.

Causal transformation remains the hardest regime, as our dataset design anticipated. CAUSAL records the lowest TCS and the highest TVR of any category for every method, and although TPD leads on all three metrics ($2.36$, $0.80$, $5.00$), the absolute level of realization stays low, since fracturing an intact frozen surface requires the trajectory to abandon a highly probable configuration, and a guidance-space correction mitigates rather than eliminates that pull. STRUCTURAL shows a similar pattern with more headroom, improving TCS from $2.84$ to $3.42$ and TVR from $0.56$ to $0.42$, while Always-Add again raises TCS ($3.12$) at the cost of a worse TVR ($0.62$).

We note two categories. In STATE, Mochi attains a marginally higher EPS than TPD ($4.96$ versus $4.84$) even though TPD leads decisively on TCS ($4.04 \to 4.70$) and TVR ($0.30 \to 0.10$); introducing a disturbance onto a pristine surface necessarily alters that surface, so a small EPS cost here is consistent with the task rather than evidence of degradation. In WEATHER\_NIGHT, TPD improves TCS ($2.68 \to 3.00$) and EPS ($4.94 \to 4.96$) but records a slightly higher TVR ($0.58 \to 0.62$), the single category in which the violation rate does not improve. This category pairs phenomenon onset with a low-light setting in which an emergent event is inherently low-contrast, making the binary presence judgment least stable; the graded TCS, which tolerates partial realization, still moves in the expected direction.

\paragraph{On the Reliability of Automated Judging.}
The judged metrics carry the interpretive weight of our evaluation, so their protocol is designed to limit variance rather than to maximize scores. The rubric is fixed and identical across methods, decoding is deterministic, the temporal partition is stated explicitly so that the judge is not left to infer the intended window, and each condition is scored by an independent query so that a favorable judgment on one does not propagate to the others. Every clip in our evaluation set is scored, with no filtering. Prior work on constrained video generation has found that judge-based compliance scores and human perceptual rankings agree on method ordering for closely related tasks~\cite{anonymous2026negate}, supporting the use of these scores as a proxy for human assessment of temporal faithfulness. The qualitative comparisons in Sec.~\ref{sec:detailed_qualitative} are drawn from the same generations and exhibit the same ordering, providing a direct visual check on the judged results.

\begin{table*}[t]
\begin{small}
    \centering
    \resizebox{\textwidth}{!}{%
    \begin{tabular}{lcccccc}
        \toprule
        \textbf{Method} & \textbf{CLIPScore} $\uparrow$ & \textbf{CLIP-early} $\downarrow$ & \textbf{BLIP} $\uparrow$ & \textbf{TCS} $\uparrow$ & \textbf{TVR} $\downarrow$ & \textbf{EPS} $\uparrow$ \\
        \midrule
        Mochi            & 0.3017 $\pm$ 0.0267 & 0.2865 $\pm$ 0.0300 & 0.6883 $\pm$ 0.1011 & 3.1850 $\pm$ 1.8396 & 0.4950 $\pm$ 0.5006 & 4.7500 $\pm$ 0.8088 \\
        HunyuanVideo     & 0.2964 $\pm$ 0.0273 & 0.2795 $\pm$ 0.0335 & 0.7098 $\pm$ 0.0862 & 2.8250 $\pm$ 1.8810 & 0.5550 $\pm$ 0.4976 & 4.6750 $\pm$ 0.9781 \\
        CogVideoX        & 0.2821 $\pm$ 0.0489 & 0.2822 $\pm$ 0.0390 & 0.6620 $\pm$ 0.1143 & 2.4050 $\pm$ 1.7970 & 0.6875 $\pm$ 0.4641 & 3.7475 $\pm$ 1.8199 \\
        \midrule
        CFG Scale-Up     & 0.3019 $\pm$ 0.0259 & 0.2863 $\pm$ 0.0305 & 0.6921 $\pm$ 0.1006 & 3.2250 $\pm$ 1.8743 & 0.5275 $\pm$ 0.4999 & 4.7650 $\pm$ 0.7288 \\
        Always-Add       & 0.3027 $\pm$ 0.0255 & 0.2850 $\pm$ 0.0304 & 0.7074 $\pm$ 0.0829 & 3.3325 $\pm$ 1.8440 & 0.4900 $\pm$ 0.5005 & 4.7675 $\pm$ 0.7279 \\
        \midrule
        \textbf{Ours}    & \textbf{0.3109 $\pm$ 0.0255} & \textbf{0.2580 $\pm$ 0.0393} & \textbf{0.7140 $\pm$ 0.0857} & \textbf{3.8775 $\pm$ 1.6353} & \textbf{0.3225 $\pm$ 0.4680} & \textbf{4.8800 $\pm$ 0.4808} \\
        \midrule
        Ours w/o Frame Axis        & 0.3085 $\pm$ 0.0252 & 0.2691 $\pm$ 0.0759 & 0.7096 $\pm$ 0.0914 & 3.4175 $\pm$ 1.8525 & 0.4775 $\pm$ 0.5001 & 4.8075 $\pm$ 0.7118 \\
        Ours w/o Timestep Schedule & 0.3083 $\pm$ 0.0254 & 0.2588 $\pm$ 0.0391 & 0.7082 $\pm$ 0.0840 & 3.4050 $\pm$ 1.8370 & 0.4475 $\pm$ 0.4979 & 4.8325 $\pm$ 0.6485 \\
        Ours w/o $\rho$-Adaptive $\alpha$ & 0.3084 $\pm$ 0.0263 & 0.2592 $\pm$ 0.0391 & 0.7044 $\pm$ 0.0916 & 3.2600 $\pm$ 1.8428 & 0.4725 $\pm$ 0.4999 & 4.8075 $\pm$ 0.7223 \\
        \midrule
    \end{tabular}}
    \caption{\textbf{Quantitative evaluation results with standard deviations.} All metrics are reported as mean $\pm$ standard deviation over the full evaluation set. Mochi, HunyuanVideo, and CogVideoX denote standard CFG baselines. CFG Scale-Up applies standard CFG with an enlarged guidance scale on Mochi, and Always-Add applies the scheduled direction $a_\tau$ additively at every step without the threshold test. Ablation variants are based on Mochi. CLIP-early is computed on late-segment frames against $p_{\text{early}}$, so lower is better. TCS denotes the Temporal Coverage Score, TVR the Temporal-suppression Violation Rate, and EPS the Early Preservation Score, all obtained from a GPT-4o vision-language judge. Since TVR is a binary per-clip indicator, its standard deviation follows the Bernoulli form and is reported for completeness.}
    \label{tab:tps_supp}
\end{small}
\end{table*}

\begin{table*}[t]
\centering
\small
\begin{tabular}{llccc}
    \toprule
    \textbf{Category} & \textbf{Method} & \textbf{TCS} $\uparrow$ & \textbf{TVR} $\downarrow$ & \textbf{EPS} $\uparrow$ \\
    \midrule
    \multirow{3}{*}{ENTRY}
    & Mochi         & 4.1000 & 0.1400 & 4.9200 \\
    & Always-Add    & 4.3600 & 0.2000 & 4.7400 \\
    & \textbf{Ours} & \textbf{4.8000} & \textbf{0.0800} & \textbf{5.0000} \\
    \midrule
    \multirow{3}{*}{SCENE}
    & Mochi         & 3.3800 & 0.5000 & 4.5600 \\
    & Always-Add    & 3.7600 & 0.4000 & 4.5200 \\
    & \textbf{Ours} & \textbf{4.3600} & \textbf{0.2000} & \textbf{4.7400} \\
    \midrule
    \multirow{3}{*}{ONSET}
    & Mochi         & 3.2600 & 0.5400 & 4.7400 \\
    & Always-Add    & 3.2200 & 0.5800 & 4.8200 \\
    & \textbf{Ours} & \textbf{4.6000} & \textbf{0.1000} & \textbf{4.8600} \\
    \midrule
    \multirow{3}{*}{STATE}
    & Mochi         & 4.0400 & 0.3000 & \textbf{4.9600} \\
    & Always-Add    & 4.1400 & 0.2200 & 4.8400 \\
    & \textbf{Ours} & \textbf{4.7000} & \textbf{0.1000} & 4.8400 \\
    \midrule
    \multirow{3}{*}{CAUSAL}
    & Mochi         & 1.8200 & 0.8600 & 4.8600 \\
    & Always-Add    & 1.9600 & 0.8600 & 4.9600 \\
    & \textbf{Ours} & \textbf{2.3600} & \textbf{0.8000} & \textbf{5.0000} \\
    \midrule
    \multirow{3}{*}{WEATHER\_NIGHT}
    & Mochi         & 2.6800 & \textbf{0.5800} & 4.9400 \\
    & Always-Add    & 2.7000 & 0.6000 & 4.7600 \\
    & \textbf{Ours} & \textbf{3.0000} & 0.6200 & \textbf{4.9600} \\
    \midrule
    \multirow{3}{*}{ACTIVITY}
    & Mochi         & 3.3600 & 0.4800 & 4.6400 \\
    & Always-Add    & 3.4000 & 0.4400 & 4.8000 \\
    & \textbf{Ours} & \textbf{3.7800} & \textbf{0.2600} & \textbf{4.9000} \\
    \midrule
    \multirow{3}{*}{STRUCTURAL}
    & Mochi         & 2.8400 & 0.5600 & 4.3800 \\
    & Always-Add    & 3.1200 & 0.6200 & 4.7000 \\
    & \textbf{Ours} & \textbf{3.4200} & \textbf{0.4200} & \textbf{4.7400} \\
    \bottomrule
\end{tabular}
\caption{\textbf{Category-wise quantitative evaluation.}
We report TCS ($\uparrow$), TVR ($\downarrow$), and EPS ($\uparrow$)
across all eight emergence categories.
All methods use Mochi as the backbone.
Always-Add applies the scheduled direction additively without the threshold test.
Categories are defined in Appendix~\ref{supp:detailed_dataset}.}
\label{tab:tps_category}
\end{table*}

\section{Detailed Ablation Study}
\label{sec:detailed_ablation}
We provide an extended analysis of the baseline and ablation variants reported in the main paper, summarized quantitatively in Table~\ref{tab:tps_supp} and illustrated in Figure~\ref{results:ablation}. The two non-projection baselines (CFG Scale-Up, Always-Add) isolate whether the correction is needed at all, while the three component removals (frame axis, timestep schedule, $\rho$-adaptive weighting) isolate what each part of the two-dimensional schedule contributes.
 
\paragraph{CFG Scale-Up.}
Enlarging the guidance scale under standard classifier-free guidance leaves the failure essentially unchanged. CLIP-early moves by less than $0.0003$ relative to the untreated backbone ($0.2863$ versus $0.2865$), and TVR is in fact \emph{higher} ($0.5275$ versus $0.4950$), with a modest CLIPScore of $0.3019$. Qualitatively, the sandcastle remains standing in every frame, and the water that reaches its base in the second half of the clip does not bring it down. This is the expected outcome, since under TPS the CFG update already points toward the early scene, so enlarging its scale reinforces the anchoring rather than overcoming it.
 
\paragraph{Always-Add.}
Retaining the full schedule~\eqref{eq:lambda} but replacing~\eqref{eq:solution}--\eqref{eq:mu} by $\delta = \delta_{\text{ref}} + \lambda(\tau,i)\,a_\tau$, applied unconditionally without the threshold test, recovers part of the suppressed signal but not reliably. On the full evaluation set, it raises TCS to $3.3325$ and lowers CLIP-early to $0.2850$, yet it trails the full method by $0.545$ TCS and $0.1675$ TVR, and its CLIP-early stays close to the untreated baseline. The figure shows the limiting case, in which the sandcastle remains intact for the entire clip despite the additive push. Because the fixed increment is applied whether or not the constraint is already met, it cannot adapt its magnitude to a strongly self-consistent early scene, and on the STRUCTURAL and ONSET categories it even records a higher TVR than the untreated backbone (Table~\ref{tab:tps_category}). Adding a constant amount of the suppressed direction is thus a partial and sometimes destabilizing remedy, in contrast to the adaptive threshold, which intervenes only where the alignment falls short and only by the amount needed to reach it.
 
\paragraph{Effect of Removing the Frame Axis.}
Setting the frame-position factor $\sigma(\beta(i/(F-1)-\gamma))$ in~\eqref{eq:lambda} to unity, so that $\lambda(\tau,i) = \alpha\,\phi(\tau)$, applies the correction equally to every frame, discarding the localization that confines restoration to the late segment. This produces the largest CLIP-early degradation among the ablations ($0.2580 \to 0.2691$) and nearly doubles its standard deviation ($0.0759$ versus roughly $0.039$ for the other two), while TCS falls to $3.4175$. Qualitatively, the correction bleeds into the opening frames, and water already disturbs the base of the sandcastle in the first frame, so the clip never establishes a clean early state and the transition loses its temporal anchor. The frame axis determines where along the clip the correction acts; CLIP-early, computed on the late frames, is where its removal is measurable.
 
\paragraph{Effect of Removing the Timestep Schedule.}
Fixing the denoising-step factor $\phi(\tau)$ in~\eqref{eq:lambda} to unity, so that $\lambda(\tau,i) = \alpha\,\sigma(\beta(i/(F-1)-\gamma))$, applies the correction uniformly across all sampling steps, including the fine-detail steps where high-frequency content is resolved. On the dataset this yields a CLIPScore and BLIP slightly below the full model, consistent with a loss of visual coherence rather than of coverage. The qualitative failure is more specific, as the sandcastle bursts apart while the wave is still offshore. Because the restorative direction is injected even at the late denoising steps that fix local appearance, the model renders the \emph{outcome} of the event, the disintegrating structure, before it has synthesized the \emph{cause}, the arriving water. The collapse therefore appears without a wave to produce it, breaking the causal ordering that the schedule is meant to preserve by concentrating correction in the early, structure-forming steps. This distinguishes the timestep factor from the frame factor. The frame factor governs \emph{where} in the clip the correction acts, whereas the timestep factor governs \emph{when} in the denoising process it acts, and removing the latter corrupts the temporal logic of the event rather than its spatial placement.
 
\paragraph{Effect of Removing $\rho$-Adaptive Weighting.}
Replacing $\alpha = \alpha_0 \max(0, \rho_{\text{early}} - \rho_{\text{late}})$ in~\eqref{eq:alpha} with the constant $\alpha_0 = 1$ removes the method's ability to scale its intervention to the estimated severity of suppression. Because $\rho_{\text{early}}, \rho_{\text{late}} \in [0,1]$, the constant is never smaller than the adaptive $\alpha$, so this variant applies an equal or stronger threshold to every prompt. It nonetheless produces the largest TCS drop of any ablation ($3.8775 \to 3.2600$), so a uniformly stronger correction does not substitute for the per-prompt one. For the sandcastle prompt ($\rho_{\text{early}} - \rho_{\text{late}} = 0.55$, so the fixed $\alpha$ is about $1.8\times$ the adaptive value), the stronger correction brings the wave onto the castle within the first third of the clip, well before the estimated transition point $\gamma = 0.5$, and part of the structure survives as a small mound. A uniformly stronger correction therefore moves the event earlier rather than realizing it more completely, which is consistent with the drop in TCS. The $\rho$-adaptive factor thus controls how much correction each prompt receives. Removing any one component degrades a different metric, and only the combination produces a correctly ordered emergence.

\section{Implementation Details of the Decomposition Stage}
\label{supp:implementation_decomposition}

The decomposition stage converts a single unsegmented prompt $p$ into the structured representation $\mathcal{D}(p) = \{p_{\text{early}}, p_{\text{late}}, \gamma, \rho_{\text{early}}, \rho_{\text{late}}\}$ consumed by the sampler. It runs once per prompt before generation begins, adds no cost to the denoising loop, and requires nothing from the user beyond the prompt itself.

\paragraph{Segment Extraction.}
Prompts in our dataset describe a persisting setting followed by an emergent event, and this structure is carried by a subordinating clause. The early segment $p_{\text{early}}$ is therefore the matrix clause describing the initial scene, and the late segment $p_{\text{late}}$ is the subordinate clause describing what should emerge over it. For \emph{``a tall sandcastle standing on a beach where a wave rushes in and washes it away,''} this yields $p_{\text{early}} =$ \emph{``a tall sandcastle standing on a beach''} and $p_{\text{late}} =$ \emph{``a wave rushes in and washes it away.''} Only $p_{\text{early}}$ is passed to the denoiser, as the conditioning of the temporal counterfactual branch; $p_{\text{late}}$ is used solely for evaluation and is never supplied to the model during generation. The full prompt is conditioned on as-is, so the model always receives the complete description and the segmentation never reaches it as a segmented input.

\paragraph{Estimating the Scalar Parameters.}
The three scalars are obtained from a single GPT-4o query. The model is instructed, through a fixed system prompt, to estimate:
\begin{itemize}
    \item $\gamma \in [0,1]$: the normalized point in the clip at which the scene should begin transitioning from the early to the late concept, expressed as a fraction of clip duration.
    \item $\rho_{\text{early}} \in [0,1]$: how strongly a video diffusion model trained on internet-scale data is expected to favor the early concept given this prompt, where a larger value implies a greater tendency to suppress what follows.
    \item $\rho_{\text{late}} \in [0,1]$: the corresponding strength of the late concept under the same distribution, where a smaller value implies greater susceptibility to suppression.
\end{itemize}
The system prompt states explicitly that $\rho_{\text{early}} \gg \rho_{\text{late}}$ signals a high risk of temporal prior suppression, while $\rho_{\text{early}} \ll \rho_{\text{late}}$ indicates a prompt likely to succeed without intervention, so the two quantities are elicited as \emph{relative} distributional strengths rather than absolute frequencies. The user message supplies the full prompt and the extracted early segment on separate lines. The model is required to answer with exactly three tab-separated numbers and no accompanying text, which makes parsing deterministic; values are clamped to $[0,1]$ after parsing. If a response fails to parse, the prompt falls back to neutral defaults $(\gamma, \rho_{\text{early}}, \rho_{\text{late}}) = (0.5, 0.7, 0.3)$. The resulting triples are stored one per line, aligned with the prompt files, and read directly by the generation script.

\paragraph{Sensitivity and Failure Behavior.}
The schedule limits, but does not remove, the effect of imprecise estimates. The frame factor is a sigmoid rather than a hard cut, so an error in $\gamma$ shifts the onset of correction smoothly instead of misplacing it discretely, and the sharpness $\beta$ controls how quickly the weighting saturates on either side. The strength factor $\alpha = \alpha_0 \max(0, \rho_{\text{early}} - \rho_{\text{late}})$ depends only on the \emph{difference} between the two prior strengths, so a systematic bias affecting both estimates equally leaves the correction unchanged, and a prompt judged to have comparable segment priors receives $\alpha \approx 0$ and reverts to standard classifier-free guidance. A poor estimate can still mislead the correction, causing under-correction when the gap $\rho_{\text{early}} - \rho_{\text{late}}$ is underestimated or over-correction when it is overestimated. Because these values come from an LLM judgment rather than from the backbone, we discuss this stage further in Sec.~\ref{sec:limitation_future}.

\section{Backbone Choice and Method Generality}
\label{supp:backbone}

\paragraph{What this section establishes.}
The contribution of this paper is twofold, namely identifying temporal prior suppression as a failure mode of current text-to-video systems, and proposing a closed-form heuristic correction of the guidance update that reduces it. Supporting these claims calls for two different kinds of evidence. The first is that the failure is not confined to one model, which requires breadth across backbones. The second is that our correction is responsible for the improvement, which requires depth on a single backbone where every schedule component can be switched off in isolation and the resulting change attributed unambiguously. We supply the former by evaluating three independent backbones under identical conditions, and the latter through the ablations conducted on Mochi. Adapting the correction's magnitude to a new backbone is a separate, downstream matter, discussed at the end of this section.

\paragraph{Temporal prior suppression recurs across independent backbones.}
A restoration method is only useful if the suppression occurs beyond a single system. Our measurements support this directly. Under standard classifier-free guidance, Mochi, HunyuanVideo, and CogVideoX all fail to place the requested late-segment concept in its intended window. Coverage scores remain between $2.4$ and $3.2$ on a five-point scale, and between roughly half and two-thirds of clips never realize the emergent event at all (Table~\ref{tab:tps_main}). These systems differ in transformer design, text encoder, scheduler family, and pretraining corpus, yet all three anchor the trajectory to the opening scene across every emergence mechanism in our dataset. Their failures are not even identical in character, since one tends to render the early scene with high fidelity and simply hold it while another degenerates into frames that lose the scene entirely (Appendix~\ref{sec:detailed_qualitative}), so the shared outcome is unlikely to be an artifact of a single architecture or training recipe. We therefore regard the deficiency as a characteristic of the present model class and argue for explicit temporal control at that level.

\paragraph{The correction uses only the classifier-free guidance update.}
For any text-conditioned diffusion model that exposes a classifier-free guidance update, our method is fully specified by four operations. First, three evaluations of the same frozen denoiser, conditioned on $\varnothing$, the full prompt $p$, and the early segment $p_{\text{early}}$, producing $\epsilon_{\text{uncond}}$, $\epsilon_{\text{text}}$, and $\epsilon_{\text{counter}}$ with no parameter touched. Second, two differences in noise-prediction space, the reference update $\delta_{\text{ref}} = w(\epsilon_{\text{text}} - \epsilon_{\text{uncond}})$ and the suppressed direction $a_\tau = \epsilon_{\text{text}} - \epsilon_{\text{counter}}$. Third, a scalar threshold $c(\tau,i) = \lambda(\tau,i)\|a_\tau\|$ formed from the schedule. Fourth, the projection of $\delta_{\text{ref}}$ onto the half-space $\{\delta : a_\tau^\top \delta \ge c(\tau,i)\}$, whose solution $\delta^\star = \delta_{\text{ref}} + \mu(\tau,i)\,a_\tau$ needs one inner product, one squared norm, and one truncation at zero. None of these steps consults the transformer's structure, the scheduler family, the noise parameterization (noise prediction or flow matching), or the identity of the text encoder. The correction is thus as architecture-neutral as the guidance update it modifies.

\paragraph{The frame axis depends only on the latent's temporal dimension.}
The correction is indexed by video frame as well as by denoising step. This does not introduce a backbone dependency. The frame factor is evaluated at the \emph{normalized} position $i/(F-1)$ within the latent temporal axis, so it is invariant to how many latent frames a given backbone allocates and to the temporal compression ratio of its autoencoder. Any model whose latent carries a temporal dimension, which video generation already requires, admits the same weighting without modification, and a model with a single temporal slot reduces to the one-dimensional case.

\paragraph{Per-backbone differences reduce to calibration.}
The one quantity our method inherits from the host model is the magnitude regime of $\epsilon_{\text{text}} - \epsilon_{\text{uncond}}$, which sets a sensible range for the guidance scale $w$ and for the global strength $\alpha_0$; the sigmoid sharpness $\beta$ and the schedule exponent $q$ shape the profile of the correction rather than its size. Backbones already ship with different default guidance scales because these magnitudes differ, quite apart from any correction applied on top. Retuning these values for a new model is therefore the ordinary calibration that every guidance-based method carries with it, and not a test of whether the thresholded projection itself transfers. Two properties limit how much calibration matters here. The threshold is relative, since $c(\tau,i)$ is proportional to $\|a_\tau\|$ and therefore rescales with the guidance magnitude instead of being fixed in absolute units, and the projection is inactive wherever the threshold is already met, so a conservative setting leaves the sampler untouched rather than distorting it.

\paragraph{Why Mochi serves as the controlled testbed.}
Among current open video backbones we adopt Mochi for three reasons. It is reachable through the standard \texttt{diffusers} interface, so the three-branch evaluation and the projection sit entirely on top of the existing sampler without altering the transformer, the scheduler, or the conditioning path. Its configuration, T5 conditioning with a flow-matching scheduler, is representative of contemporary open systems rather than unusual, so conclusions drawn on it are unlikely to be artifacts of an atypical design. And its inference stack is public and self-contained, which is what allows each reported ablation to be reproduced exactly, including the removals that separate the frame axis, the timestep schedule, and the prior-strength weighting.

\paragraph{Outlook.}
A full quantitative study of the correction applied to HunyuanVideo and CogVideoX, and further to image backbones where the frame axis collapses and only the timestep schedule remains, would require per-backbone calibration of $(w, \alpha_0, \beta, q)$ together with the corresponding evaluation pipelines. That effort is engineering rather than formulation, and orthogonal to the claim established here, which is that the failure is shared across the model class, and the correction is defined entirely in guidance space. We regard this extension as the natural continuation of the mechanism, the decomposition stage, and the dataset introduced in this work.

\section{Derivation and Stability of the Minimal-Energy Projection}
\label{supp:derivation}

\paragraph{KKT Derivation.}
The Lagrangian of the quadratic program~\eqref{eq:qp} is
\begin{equation}
\mathcal{L}(\delta, \mu) = \frac{1}{2}\|\delta - \delta_{\text{ref}}\|_2^2 - \mu\left(a_\tau^\top \delta - c(\tau, i)\right),
\end{equation}
with dual variable $\mu \geq 0$. Setting $\nabla_\delta \mathcal{L} = 0$ gives $\delta = \delta_{\text{ref}} + \mu a_\tau$. Substituting into the complementary slackness condition $\mu(a_\tau^\top \delta - c(\tau,i)) = 0$ and the dual feasibility condition $\mu \geq 0$ yields the closed-form solution~\eqref{eq:solution} with multiplier~\eqref{eq:mu}. When the constraint is violated, the correction magnitude is
\begin{equation}
\|\delta^*(\tau,i) - \delta_{\text{ref}}\| = \mu(\tau,i)\|a_\tau\| = \frac{\max\{0,\, c(\tau,i) - a_\tau^\top\delta_{\text{ref}}\}}{\|a_\tau\|},
\end{equation}
proportional to the degree of violation and inversely proportional to the norm of the suppressed signal direction.

\paragraph{Stability and Convergence.}
Because the constraint set defined by~\eqref{eq:lowerbound} is a closed half-space and the objective in~\eqref{eq:qp} is strictly convex, the solution exists and is unique. Projection onto a closed half-space is non-expansive:
\begin{equation}
\|\delta^*(\tau,i) - \delta^{*\prime}(\tau,i)\| \leq \|\delta_{\text{ref}} - \delta_{\text{ref}}'\|,
\end{equation}
ensuring Lipschitz continuity of the corrected update with respect to the reference, so small perturbations in the denoising trajectory do not amplify through the correction mechanism. Unlike steep repulsive potentials or large guidance-scale amplification, the projection introduces no stiffness into the reverse-time ODE. The correction magnitude satisfies
\begin{equation}
\|\delta^*(\tau,i) - \delta_{\text{ref}}\| = \mu(\tau,i) \|a_\tau\| \leq \frac{|c(\tau,i) - a_\tau^\top \delta_{\text{ref}}|}{\|a_\tau\|},
\end{equation}
so no oscillatory amplification occurs. When the suppressed direction is weakly expressed, $\|a_\tau\|$ becomes small; the fixed $\varepsilon = 10^{-8}$ added to $\|a_\tau\|_2^2$ in~\eqref{eq:mu} ensures numerical stability, and we empirically verify that $\|a_\tau\|$ remains non-negligible in the late-segment frames where restoration is applied. When $c(\tau,i) \leq a_\tau^\top \delta_{\text{ref}}$, the threshold is already met and $\mu = 0$, so the method intervenes only when the alignment falls below the threshold and otherwise preserves the original CFG update.

\section{Computation Time and Memory Consumption}
\label{supp:computation_text}
\begin{table}[htb!]
    \small
    \centering
    \resizebox{0.5\textwidth}{!}{
    \begin{tabular}{l|l|c|c}
    \toprule
    \textbf{Model} & \textbf{Step} &
    \textbf{Memory Consumption} &
    \textbf{Computation Time} \\
    \midrule
    \textbf{Ours} & Inference & 26.3057 GB / 80.0000 GB (26,937 MiB) & 248 sec \\
    \midrule
    Ours w/o Frame Axis        & Inference & 25.4111 GB / 80.0000 GB (26,021 MiB) & 168 sec \\
    Ours w/o Timestep Schedule & Inference & 25.4209 GB / 80.0000 GB (26,031 MiB) & 168 sec \\
    Ours w/o $\rho$-Adaptive $\alpha$ & Inference & 25.4111 GB / 80.0000 GB (26,021 MiB) & 168 sec \\
    \midrule
    CFG Scale-Up & Inference & 24.4541 GB / 80.0000 GB (25,041 MiB) & 117 sec \\
    Always-Add   & Inference & 26.2998 GB / 80.0000 GB (26,931 MiB) & 168 sec \\
    \midrule
    Mochi & Inference & 24.4385 GB / 80.0000 GB (25,025 MiB) & 114 sec \\
    HunyuanVideo & Inference & 38.5186 GB / 80.0000 GB (39,443 MiB) & 396 sec \\
    CogVideoX & Inference & 15.6318 GB / 80.0000 GB (16,007 MiB) & 61 sec \\
    \bottomrule
    \end{tabular}
    }
    \caption{\textbf{Inference-time memory and runtime on a single NVIDIA H100 (80GB).} Peak GPU memory and end-to-end inference time are reported for our full method, ablation variants, and all baselines. All methods share the same backbone (Mochi) except HunyuanVideo and CogVideoX. All measurements are conducted under identical hardware and evaluation conditions.}
    \label{tab:computation}
\end{table}
Table~\ref{tab:computation} reports peak GPU memory and end-to-end inference time on a single NVIDIA H100 (80GB) for all evaluated methods.

\paragraph{Computational overhead of TPD.}
Our full method evaluates three denoiser branches per step (unconditional, full-prompt, and the temporal counterfactual conditioned on $p_{\text{early}}$) against the two branches of standard classifier-free guidance. Relative to Mochi (114 sec), the full method takes 248 sec. Part of this reflects the additional counterfactual forward pass, and the remainder comes from the frame-wise thresholded projection, which checks and corrects a per-frame alignment threshold at every denoising step. Peak memory rises only marginally, from 24.44 GB (Mochi) to 26.31 GB (Ours), since the three branches share the same frozen weights and only the counterfactual activations are added; the projection introduces no parameters and negligible memory, as it operates on flattened latent slices with a single inner product and norm per frame. The cost is therefore dominated by repeated evaluation and per-frame projection at inference time, and requires no training or architectural change.

\paragraph{Ablation variants.}
The three component removals all retain the full three-branch structure and differ only in how the correction is scheduled and solved, yet each runs at 168 sec, substantially faster than the full method (248 sec). The gap isolates the cost of the complete two-dimensional thresholded projection. when the frame axis is removed the correction no longer varies per frame, when the timestep schedule is removed it no longer varies per step, and when the $\rho$-adaptive weighting is removed a fixed strength is used, and in each case the per-frame constraint solve is correspondingly simplified. Memory is essentially unchanged across these variants (25.41--25.42 GB), consistent with the projection contributing computation rather than storage. Always-Add (168 sec, 26.30 GB) matches this group, since it keeps all three branches but adds the scheduled direction unconditionally, skipping the threshold check, which confirms that the additional runtime of the full method stems specifically from the thresholded projection rather than from the counterfactual branch alone.

\paragraph{Baselines across backbones.}
CFG Scale-Up (117 sec, 24.45 GB) runs at essentially Mochi's cost, as it is standard two-branch guidance with an enlarged scale and no counterfactual branch; the 3 sec difference is measurement noise. Among the backbones, HunyuanVideo is the heaviest at 396 sec and 38.52 GB, and CogVideoX the lightest at 61 sec and 15.63 GB, spanning a wide efficiency range that is independent of our correction. Our method's overhead is incurred entirely at inference and buys a substantial improvement in temporal-emergence fidelity without retraining or modifying the backbone; where inference latency is critical, the ablation timings indicate that a reduced schedule recovers most of the runtime (168 vs.\ 248 sec) at a cost of 0.46--0.62 TCS (Table~\ref{tab:tps_supp}).

\begin{figure*}[t]
    \centering
    \includegraphics[width=0.95\linewidth]{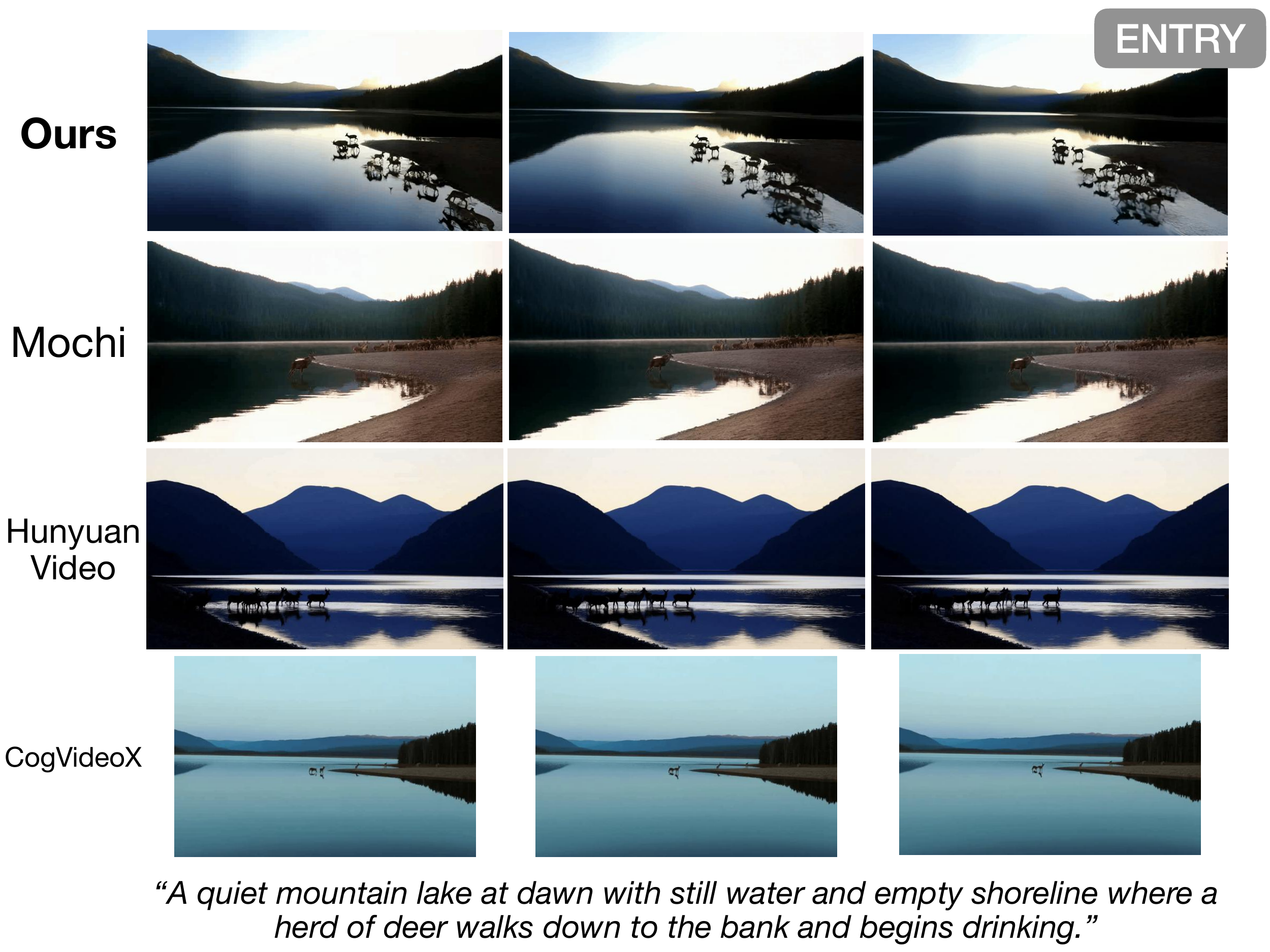}
    \caption{\textbf{ENTRY.}
    Prompt: ``A quiet mountain lake at dawn with still water and empty shoreline where a herd of deer walks down to the bank and begins drinking.''}
\end{figure*}

\begin{figure*}[t]
    \centering
    \includegraphics[width=0.95\linewidth]{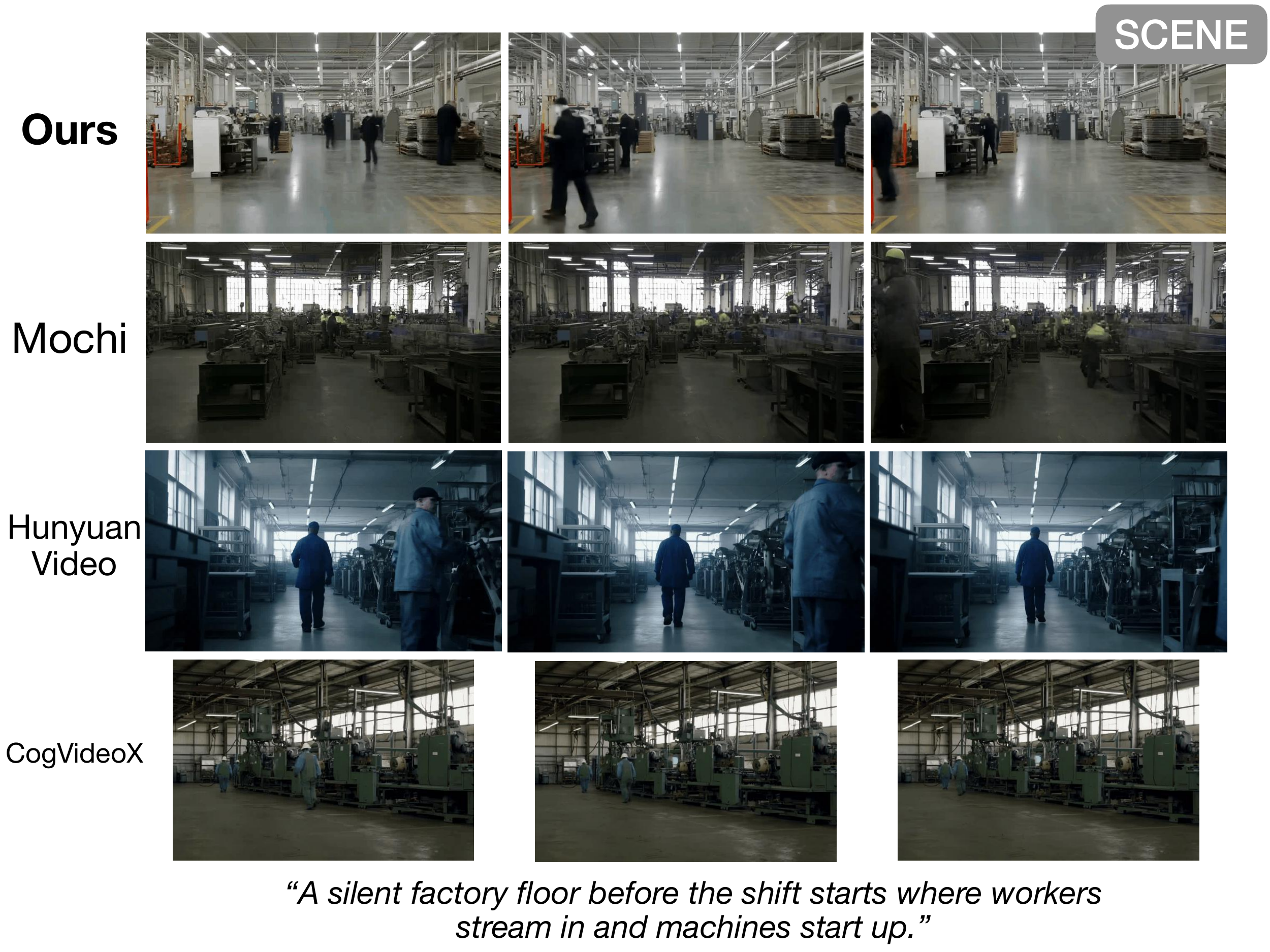}
    \caption{\textbf{SCENE.}
    Prompt: ``A silent factory floor before the shift starts where workers stream in and machines start up.''}
\end{figure*}

\begin{figure*}[t]
    \centering
    \includegraphics[width=0.95\linewidth]{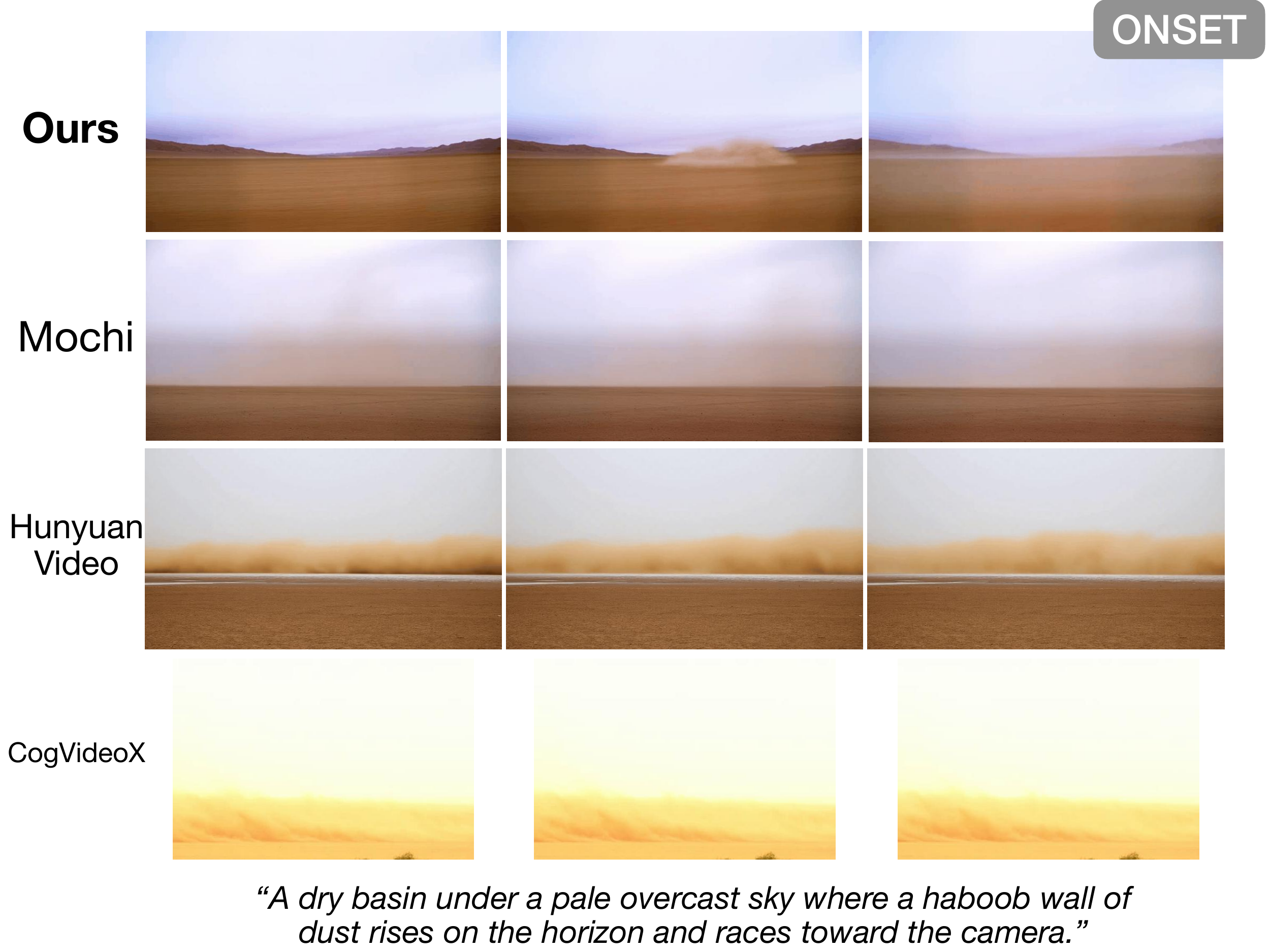}
    \caption{\textbf{ONSET.}
    Prompt: ``A dry basin under a pale overcast sky where a haboob wall of dust rises on the horizon and races toward the camera.''}
\end{figure*}

\begin{figure*}[t]
    \centering
    \includegraphics[width=0.95\linewidth]{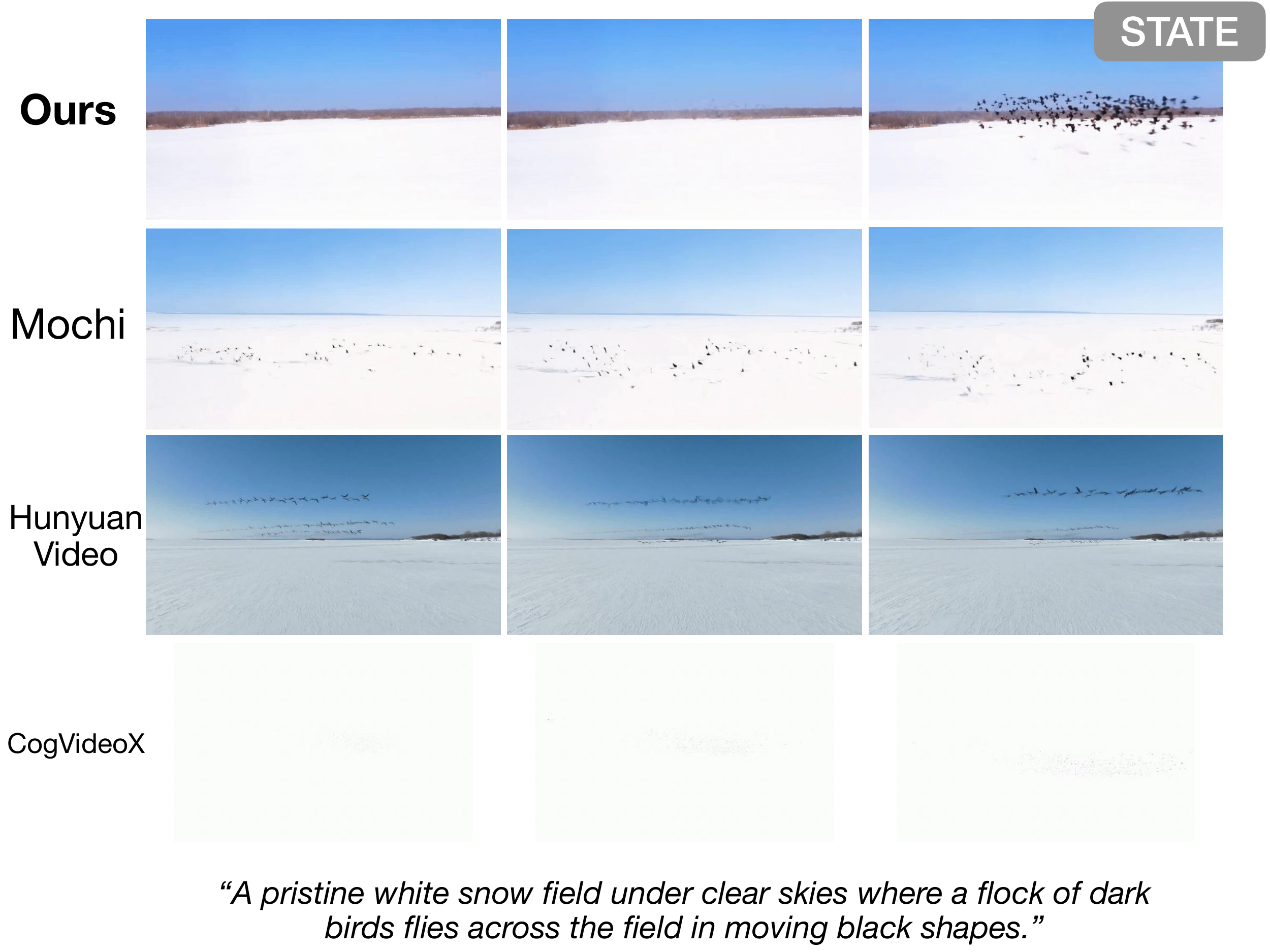}
    \caption{\textbf{STATE.}
    Prompt: ``A pristine white snow field under clear skies where a flock of dark birds flies across the field in moving black shapes.''}
\end{figure*}

\begin{figure*}[t]
    \centering
    \includegraphics[width=0.95\linewidth]{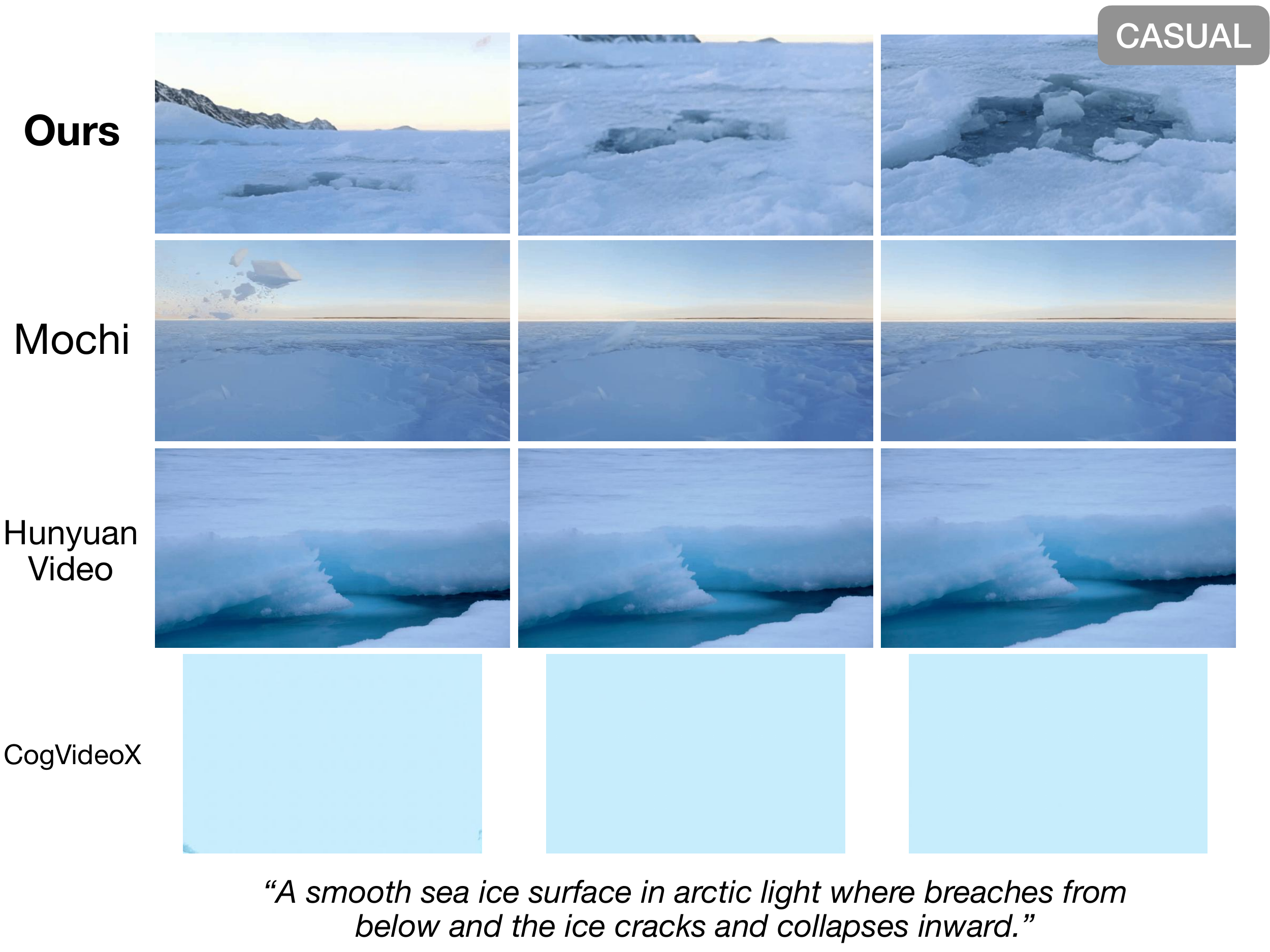}
    \caption{\textbf{CAUSAL.}
    Prompt: ``A smooth sea ice surface in arctic light where breaches from below and the ice cracks and collapses inward.''}
\end{figure*}

\begin{figure*}[t]
    \centering
    \includegraphics[width=0.95\linewidth]{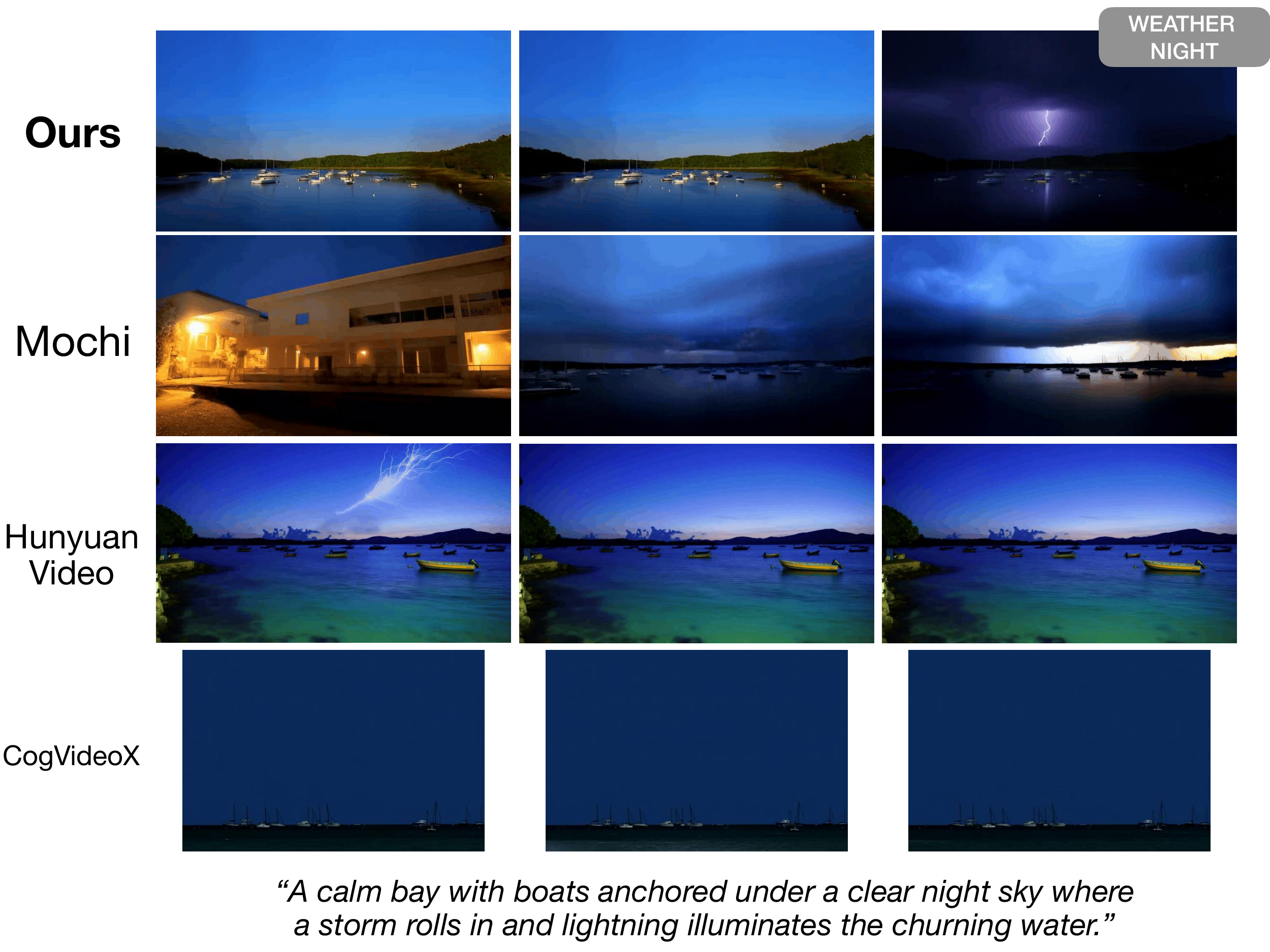}
    \caption{\textbf{WEATHER\_NIGHT.}
    Prompt: ``A calm bay with boats anchored under a clear night sky where a storm rolls in and lightning illuminates the churning water.''}
\end{figure*}

\begin{figure*}[t]
    \centering
    \includegraphics[width=0.95\linewidth]{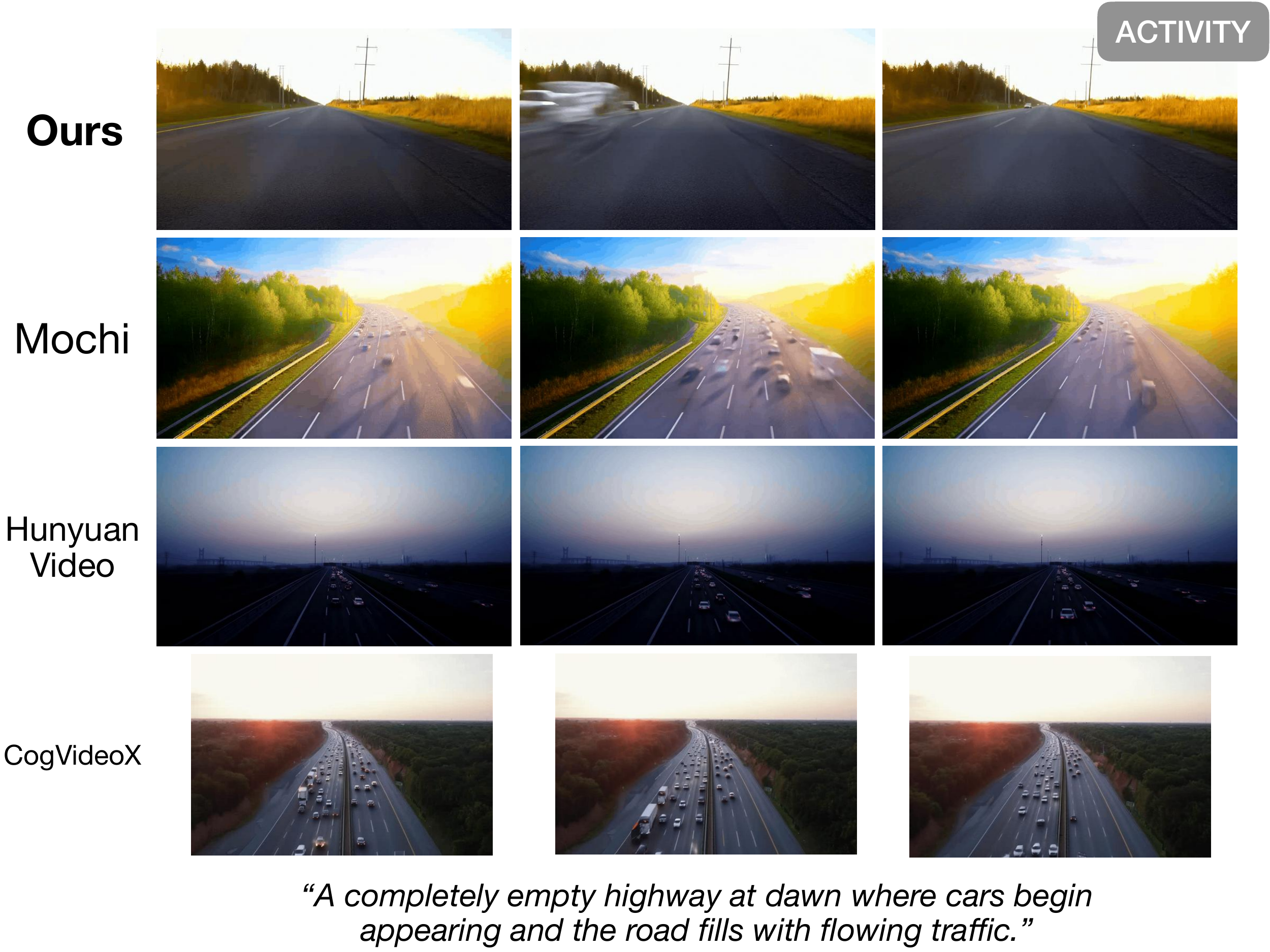}
    \caption{\textbf{ACTIVITY.}
    Prompt: ``A completely empty highway at dawn where cars begin appearing and the road fills with flowing traffic.''}
\end{figure*}

\begin{figure*}[t]
    \centering
    \includegraphics[width=0.95\linewidth]{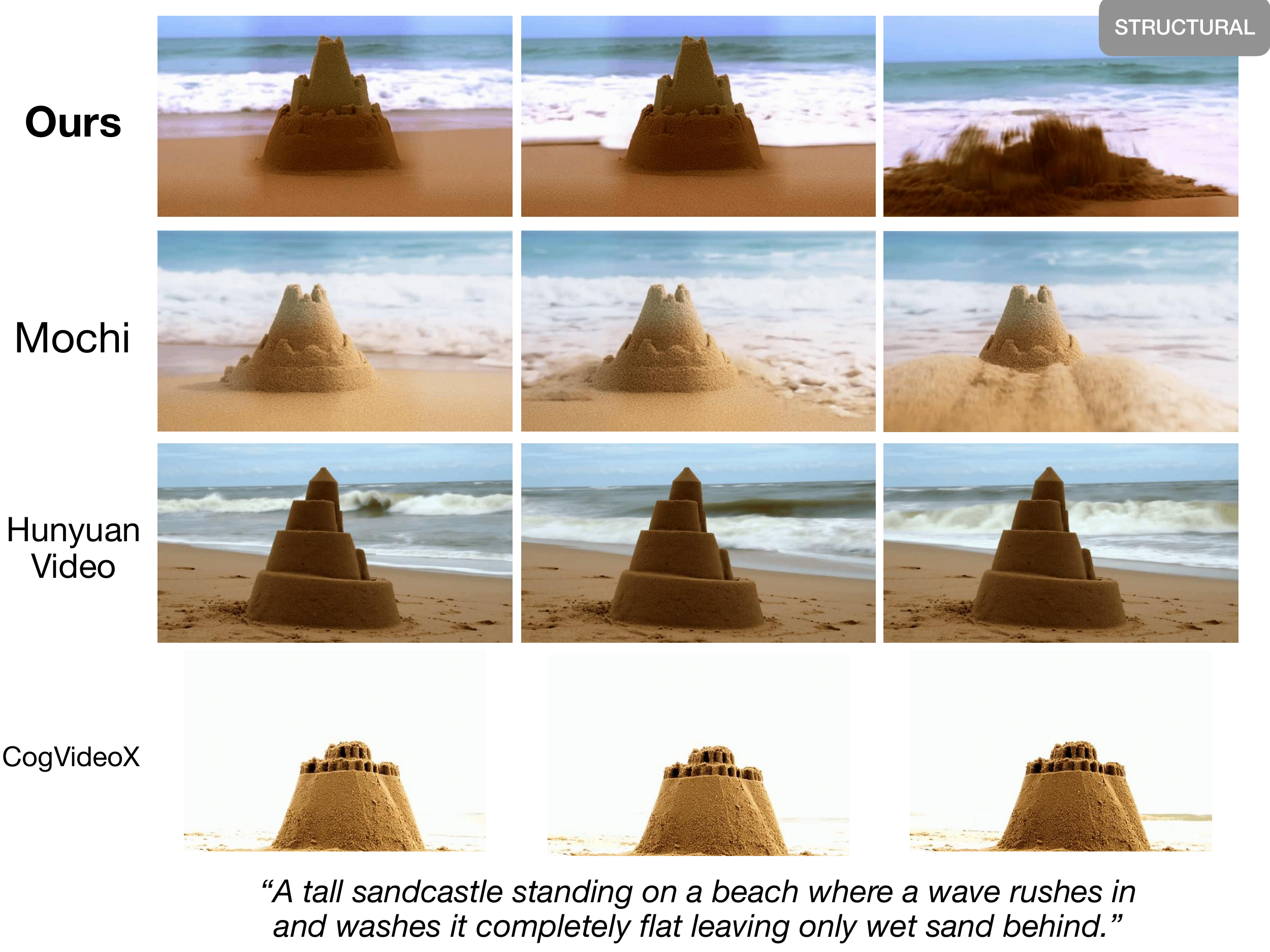}
    \caption{\textbf{STRUCTURAL.}
    Prompt: ``A tall sandcastle standing on a beach where a wave rushes in and washes it completely flat leaving only wet sand behind.''}
\end{figure*}

\begin{figure*}[t]
    \centering
    \includegraphics[width=0.95\linewidth]{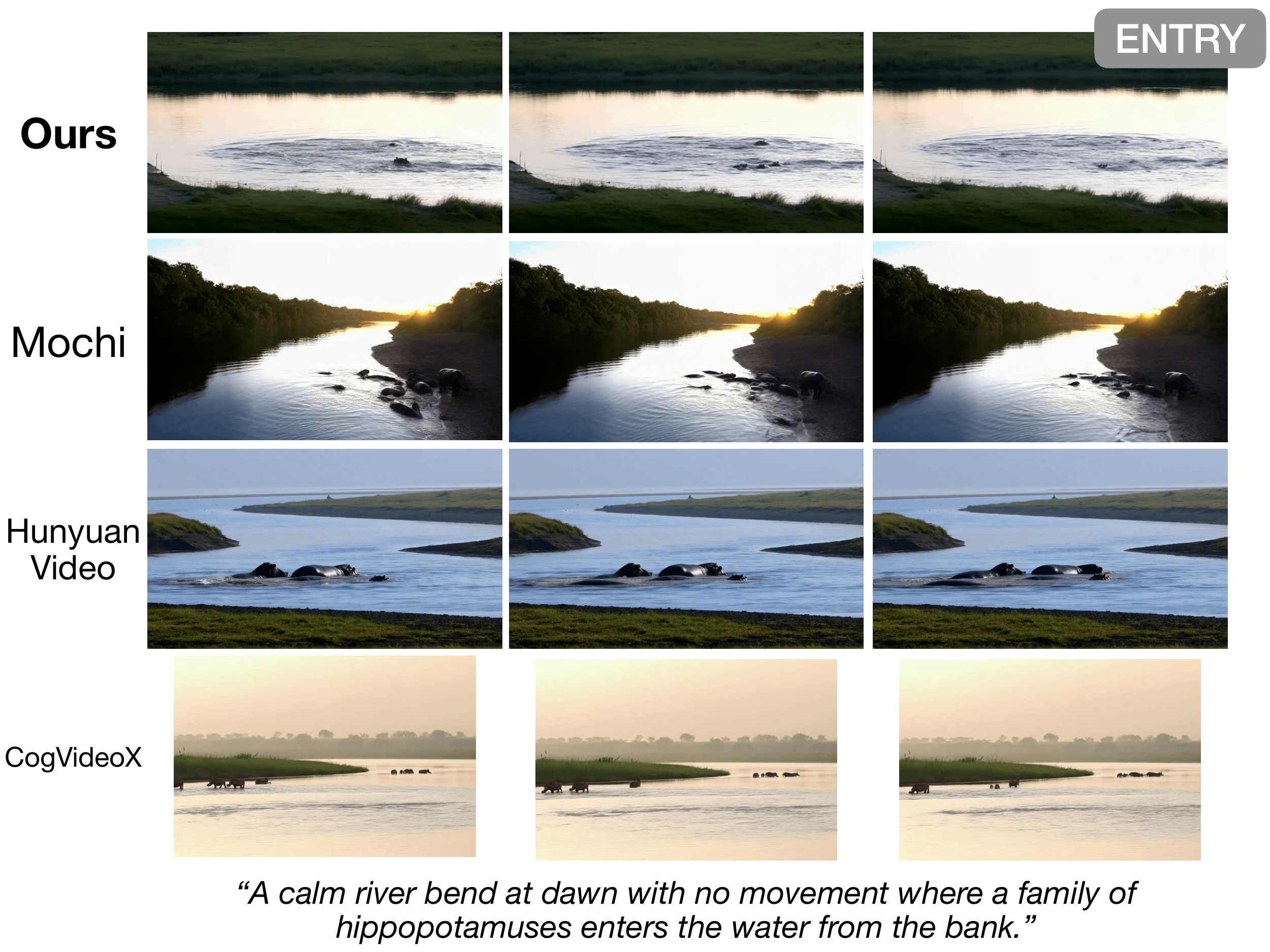}
    \caption{\textbf{ENTRY.}
    Prompt: ``A calm river bend at dawn with no movement where a family of hippopotamuses enters the water from the bank.''}
\end{figure*}

\begin{figure*}[t]
    \centering
    \includegraphics[width=0.95\linewidth]{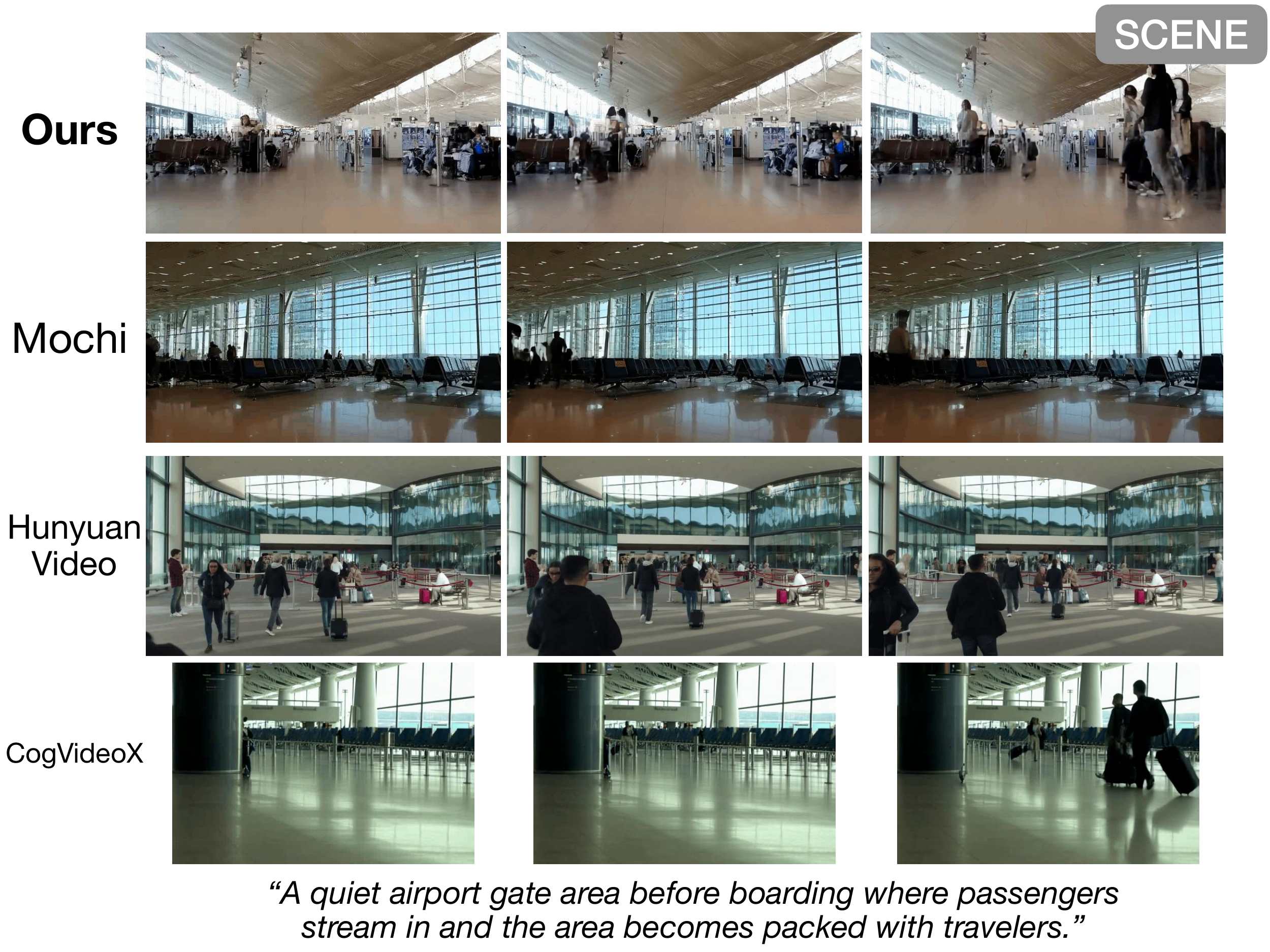}
    \caption{\textbf{SCENE.}
    Prompt: ``A quiet airport gate area before boarding where passengers stream in and the area becomes packed with travelers.''}
\end{figure*}

\begin{figure*}[t]
    \centering
    \includegraphics[width=0.95\linewidth]{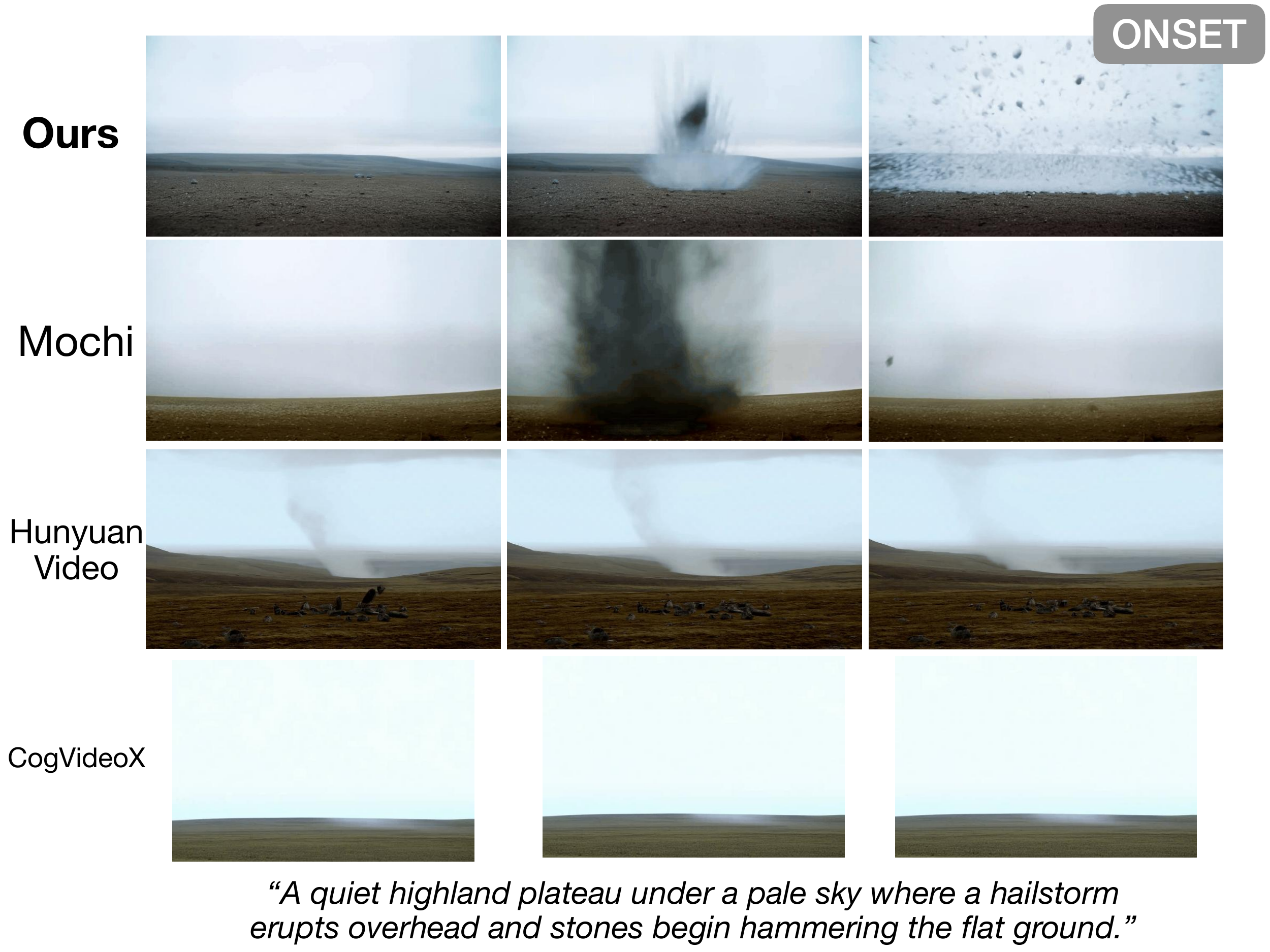}
    \caption{\textbf{ONSET.}
    Prompt: ``A quiet highland plateau under a pale sky where a hailstorm erupts overhead and stones begin hammering the flat ground.''}
\end{figure*}

\begin{figure*}[t]
    \centering
    \includegraphics[width=0.95\linewidth]{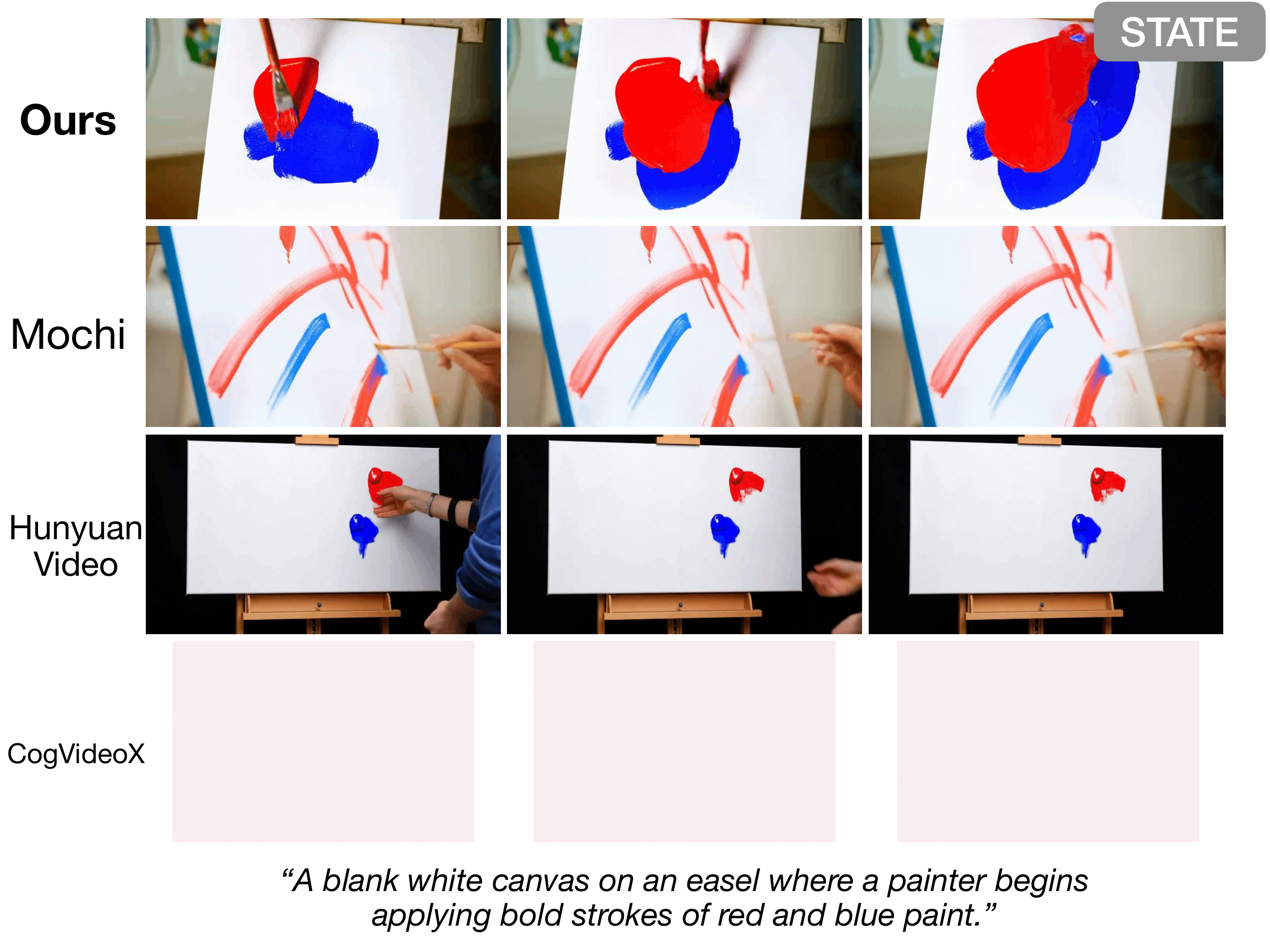}
    \caption{\textbf{STATE.}
    Prompt: ``A blank white canvas on an easel where a painter begins applying bold strokes of red and blue paint.''}
\end{figure*}

\begin{figure*}[t]
    \centering
    \includegraphics[width=0.95\linewidth]{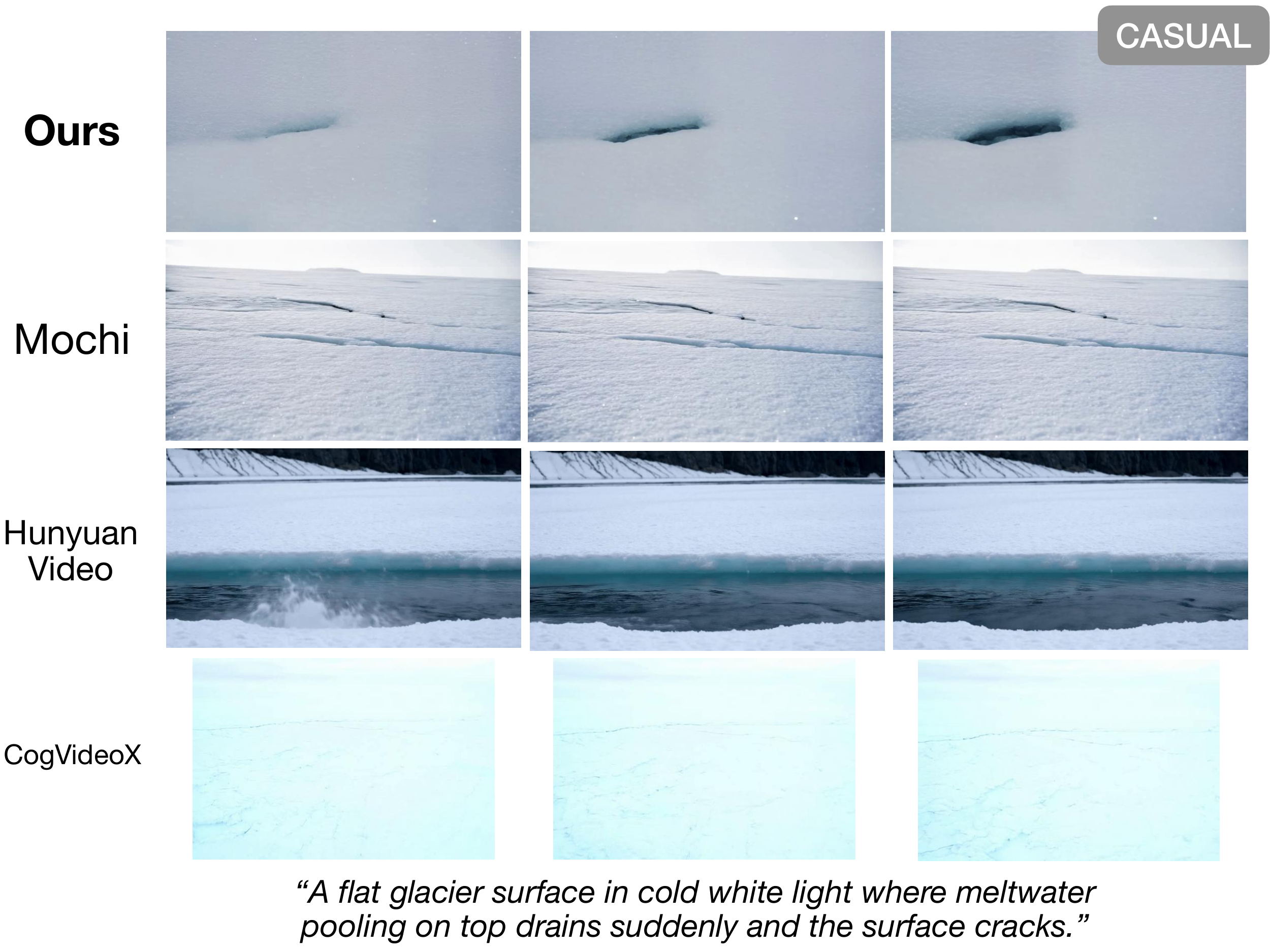}
    \caption{\textbf{CAUSAL.}
    Prompt: ``A flat glacier surface in cold white light where meltwater pooling on top drains suddenly and the surface cracks.''}
\end{figure*}

\begin{figure*}[t]
    \centering
    \includegraphics[width=0.95\linewidth]{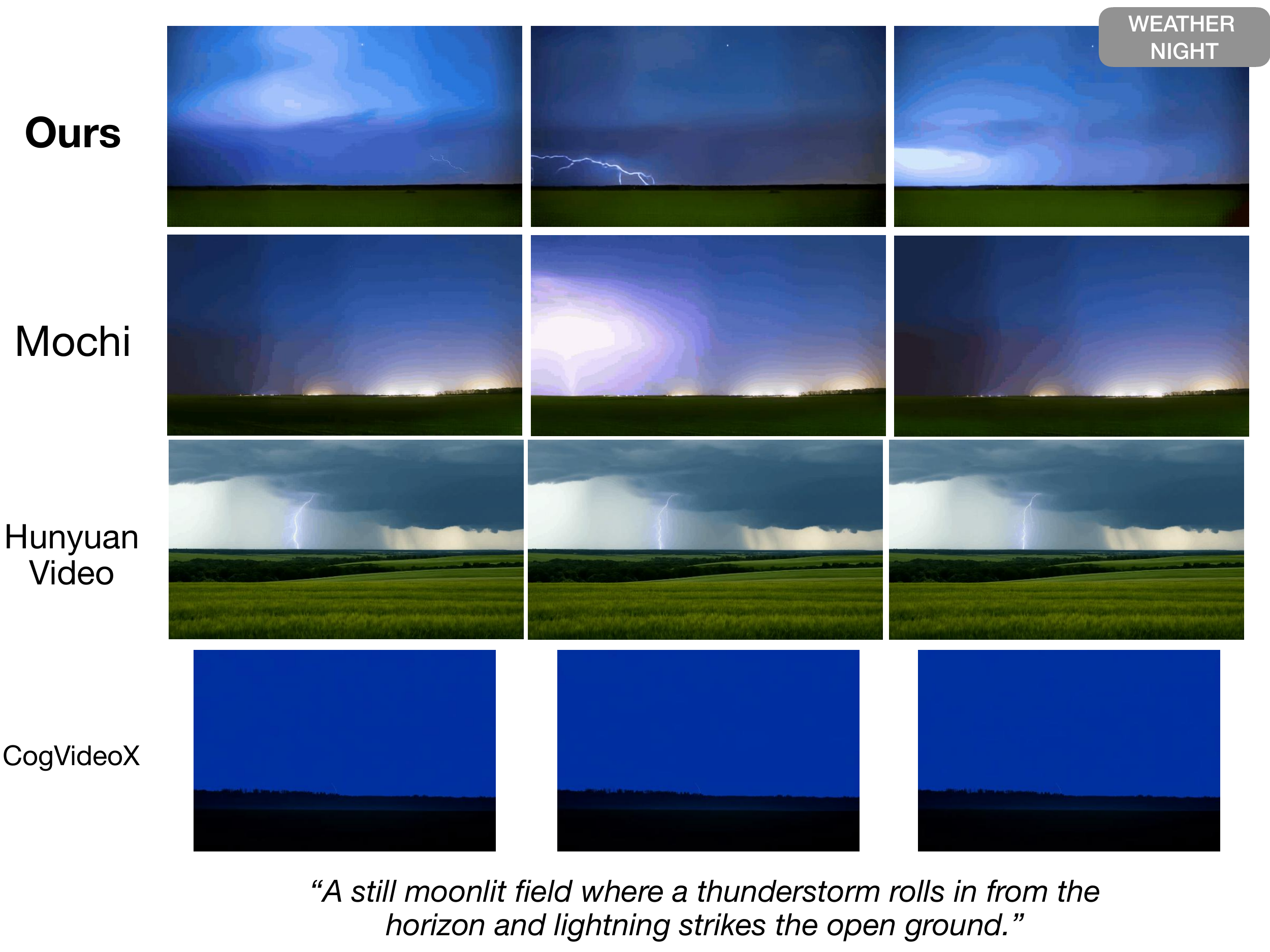}
    \caption{\textbf{WEATHER\_NIGHT.}
    Prompt: ``A still moonlit field where a thunderstorm rolls in from the horizon and lightning strikes the open ground.''}
\end{figure*}

\begin{figure*}[t]
    \centering
    \includegraphics[width=0.95\linewidth]{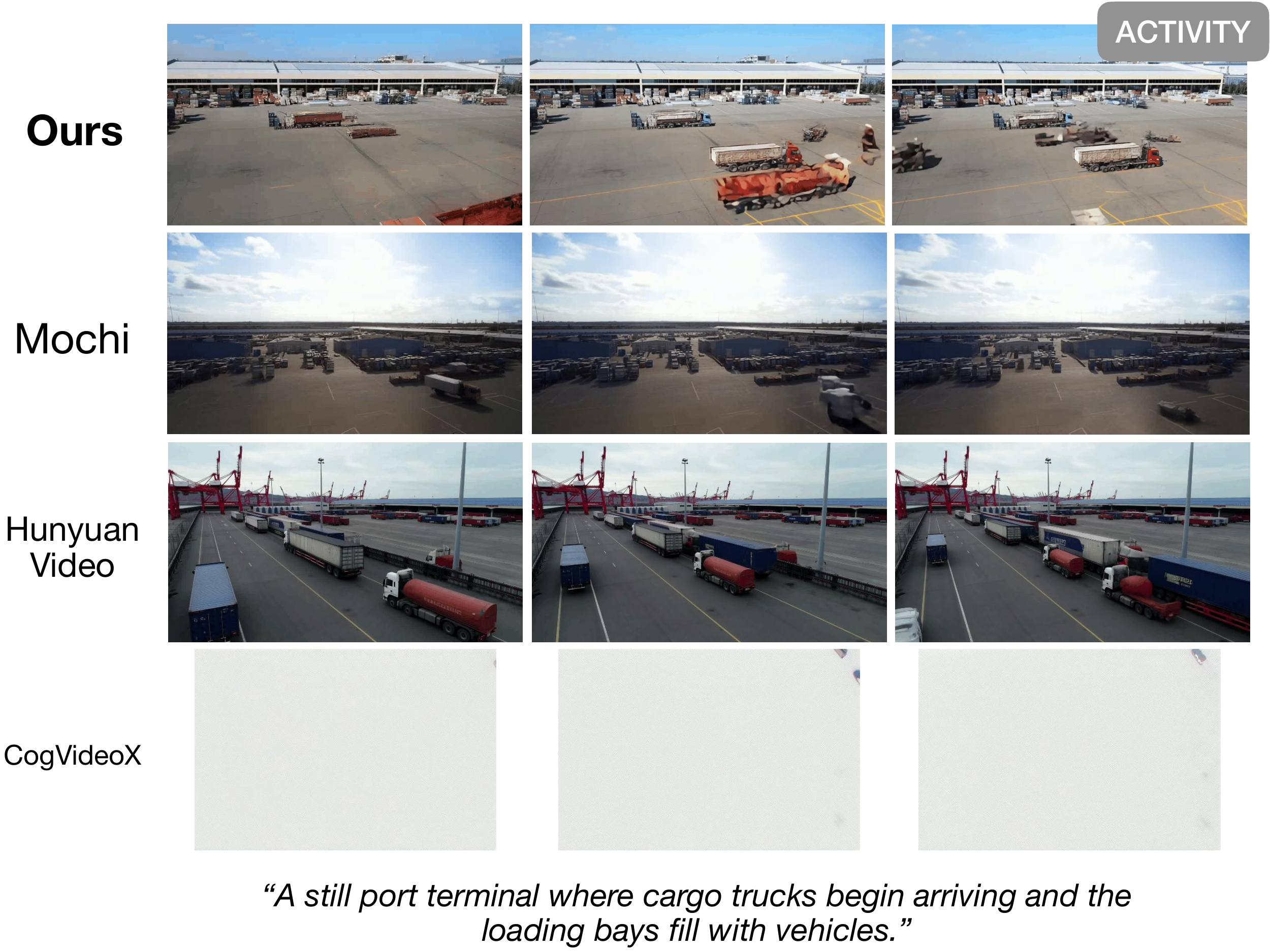}
    \caption{\textbf{ACTIVITY.}
    Prompt: ``A still port terminal where cargo trucks begin arriving and the loading bays fill with vehicles.''}
\end{figure*}

\begin{figure*}[t]
    \centering
    \includegraphics[width=0.95\linewidth]{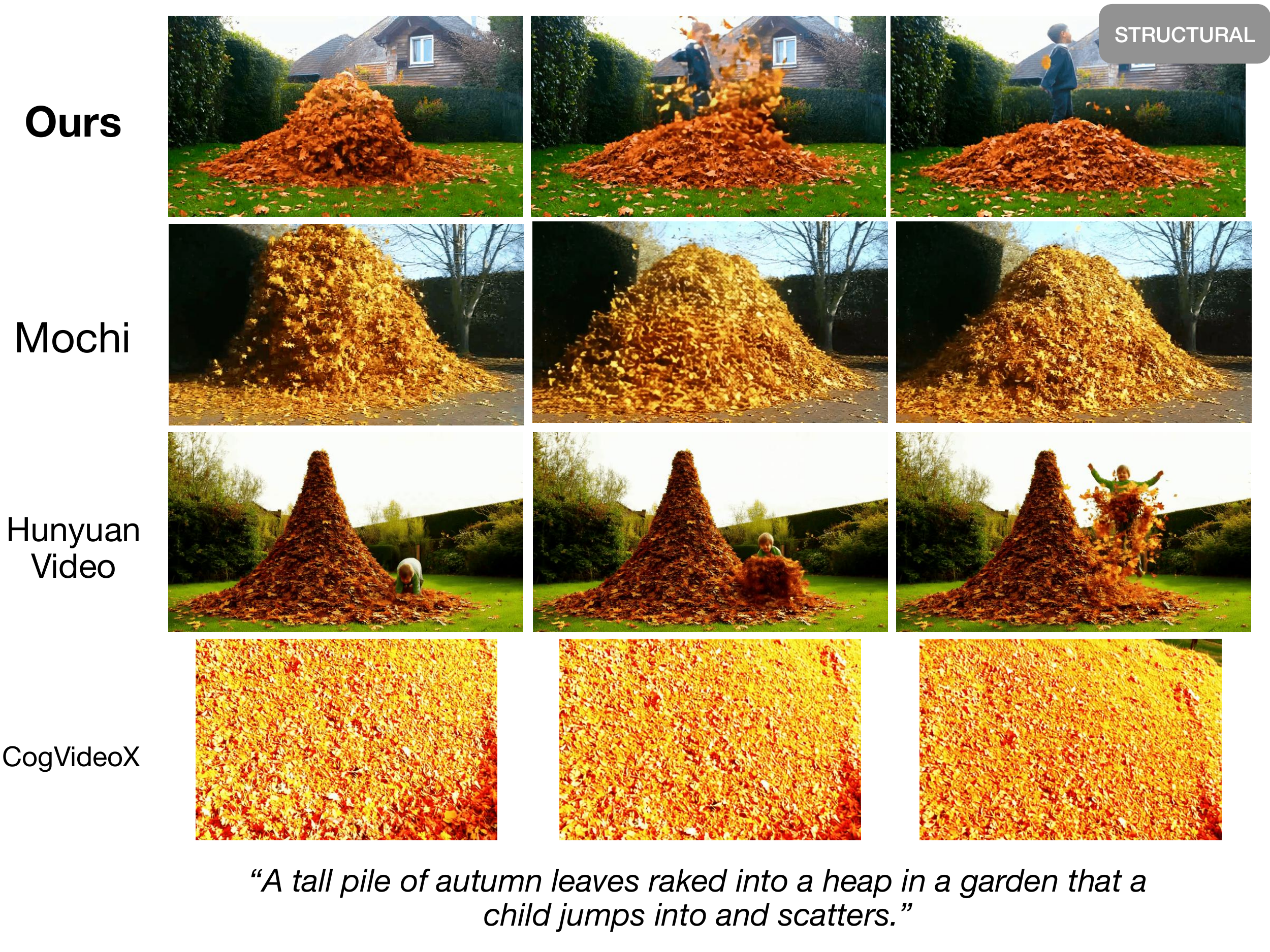}
    \caption{\textbf{STRUCTURAL.}
    Prompt: ``A tall pile of autumn leaves raked into a heap in a garden that a child jumps into and scatters.''}
\end{figure*}

\begin{figure*}[t]
    \centering
    \includegraphics[width=0.95\linewidth]{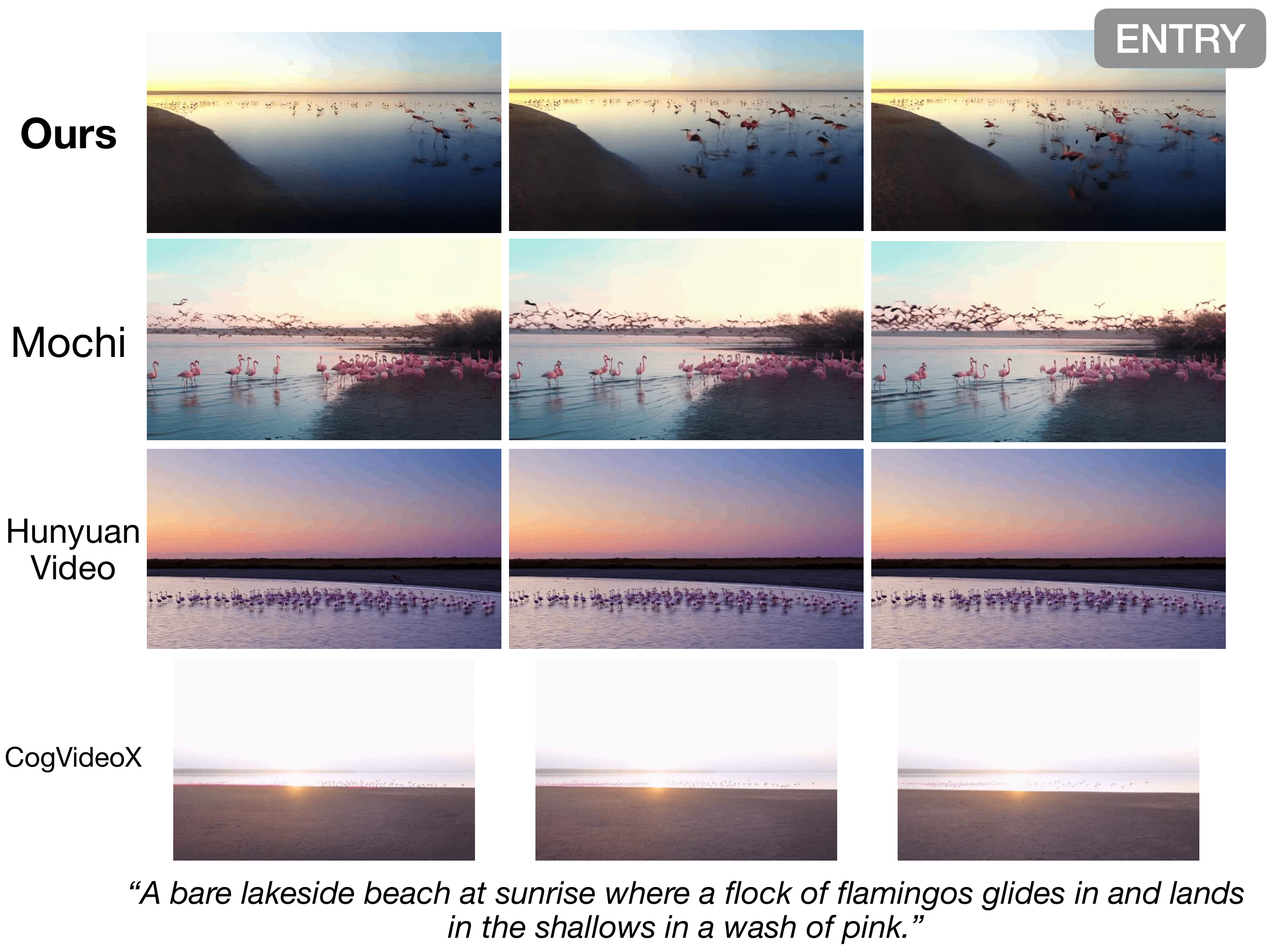}
    \caption{\textbf{ENTRY.}
    Prompt: ``A bare lakeside beach at sunrise where a flock of flamingos glides in and lands in the shallows in a wash of pink.''}
\end{figure*}

\begin{figure*}[t]
    \centering
    \includegraphics[width=0.95\linewidth]{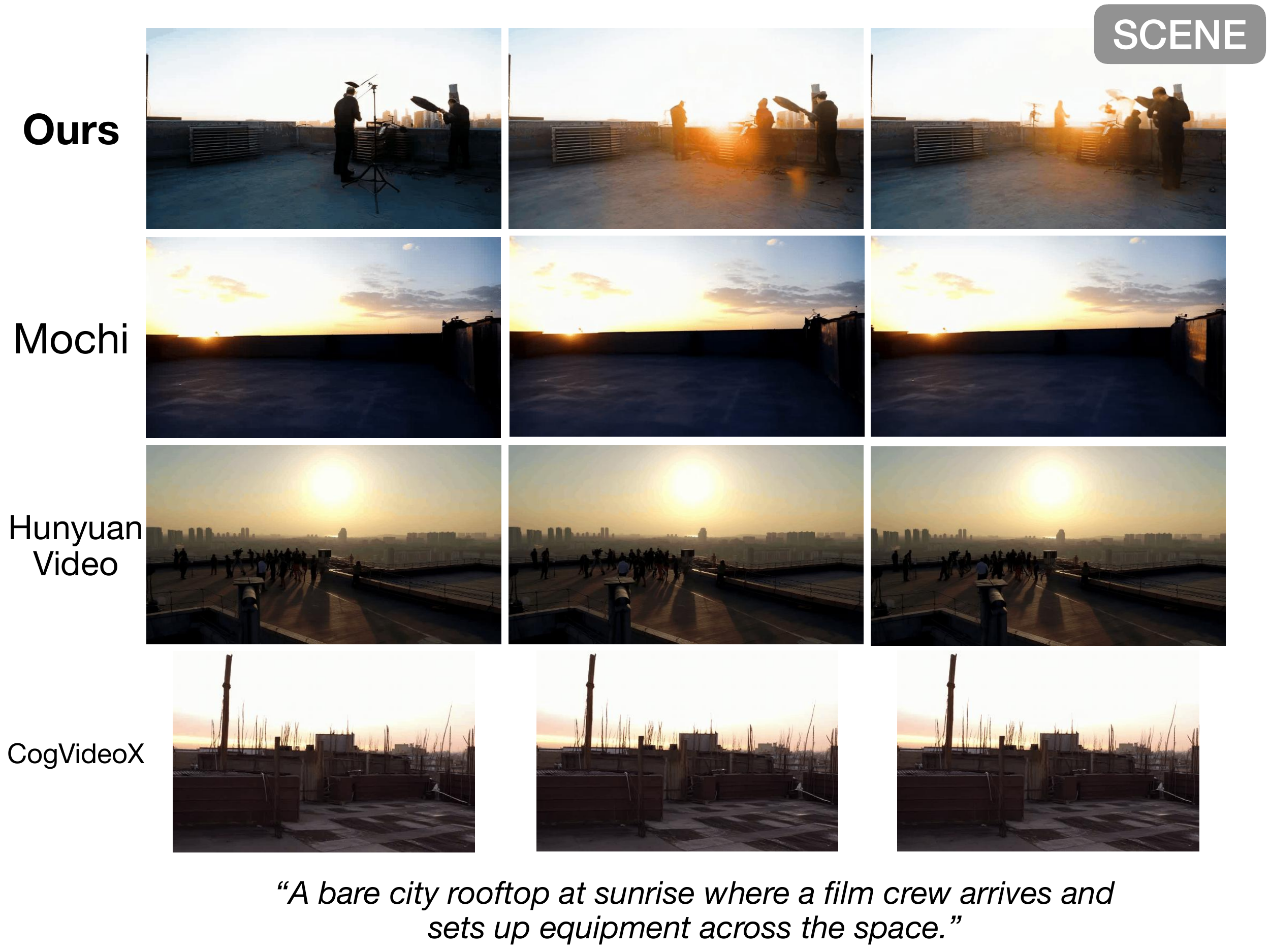}
    \caption{\textbf{SCENE.}
    Prompt: ``A bare city rooftop at sunrise where a film crew arrives and sets up equipment across the space.''}
\end{figure*}

\begin{figure*}[t]
    \centering
    \includegraphics[width=0.95\linewidth]{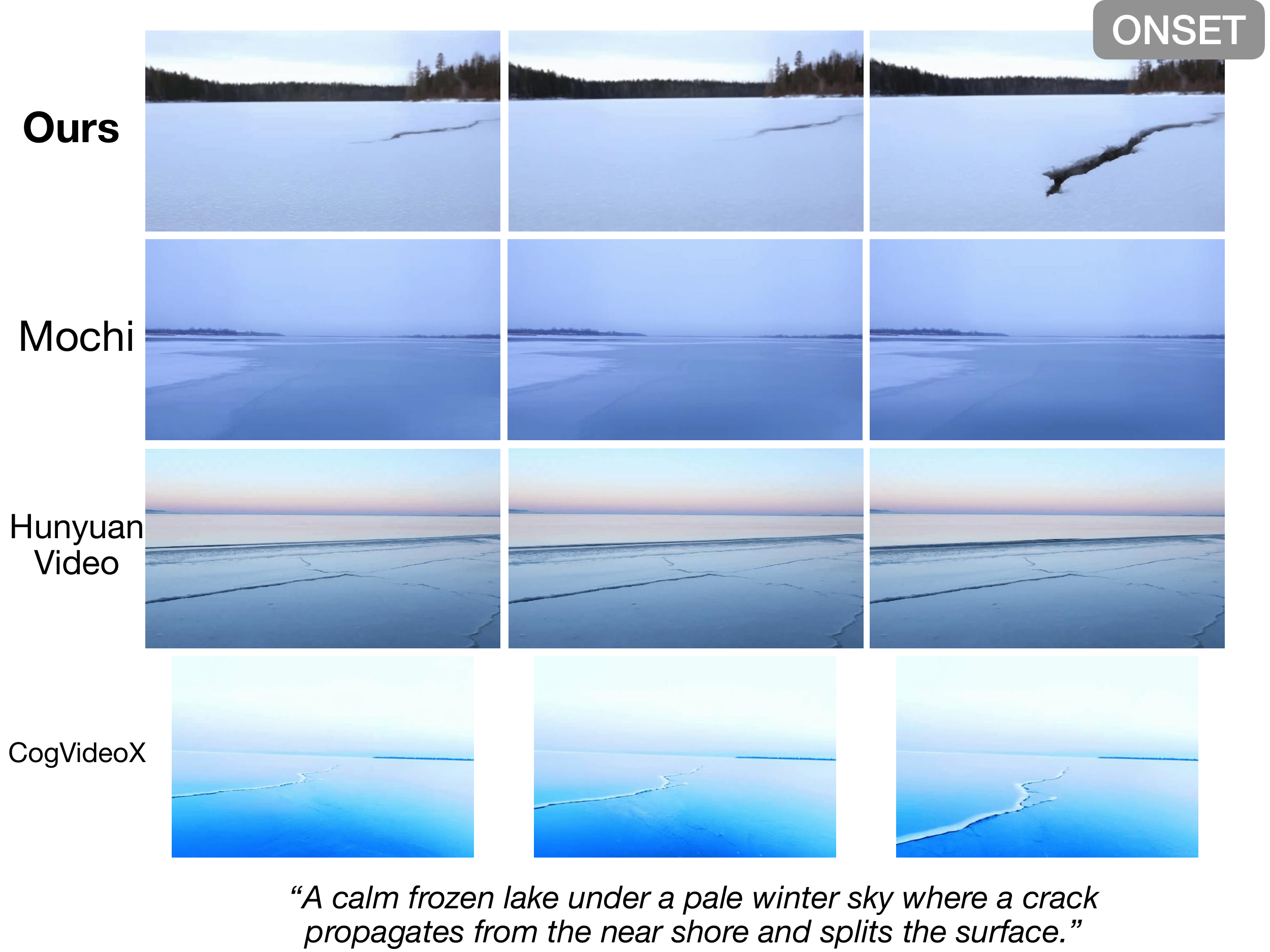}
    \caption{\textbf{ONSET.}
    Prompt: ``A calm frozen lake under a pale winter sky where a crack propagates from the near shore and splits the surface.''}
\end{figure*}

\begin{figure*}[t]
    \centering
    \includegraphics[width=0.95\linewidth]{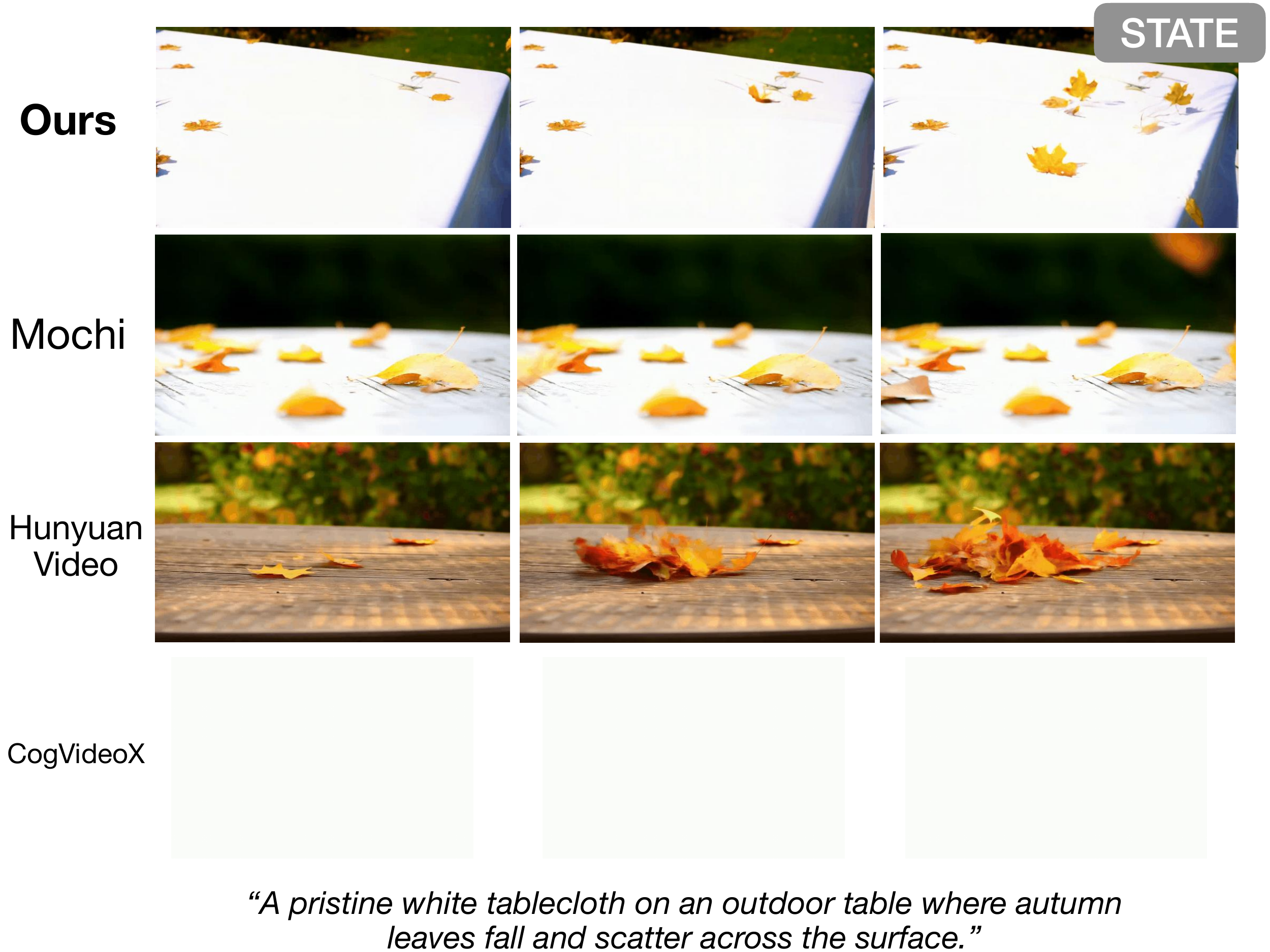}
    \caption{\textbf{STATE.}
    Prompt: ``A pristine white tablecloth on an outdoor table where autumn leaves fall and scatter across the surface.''}
\end{figure*}

\begin{figure*}[t]
    \centering
    \includegraphics[width=0.95\linewidth]{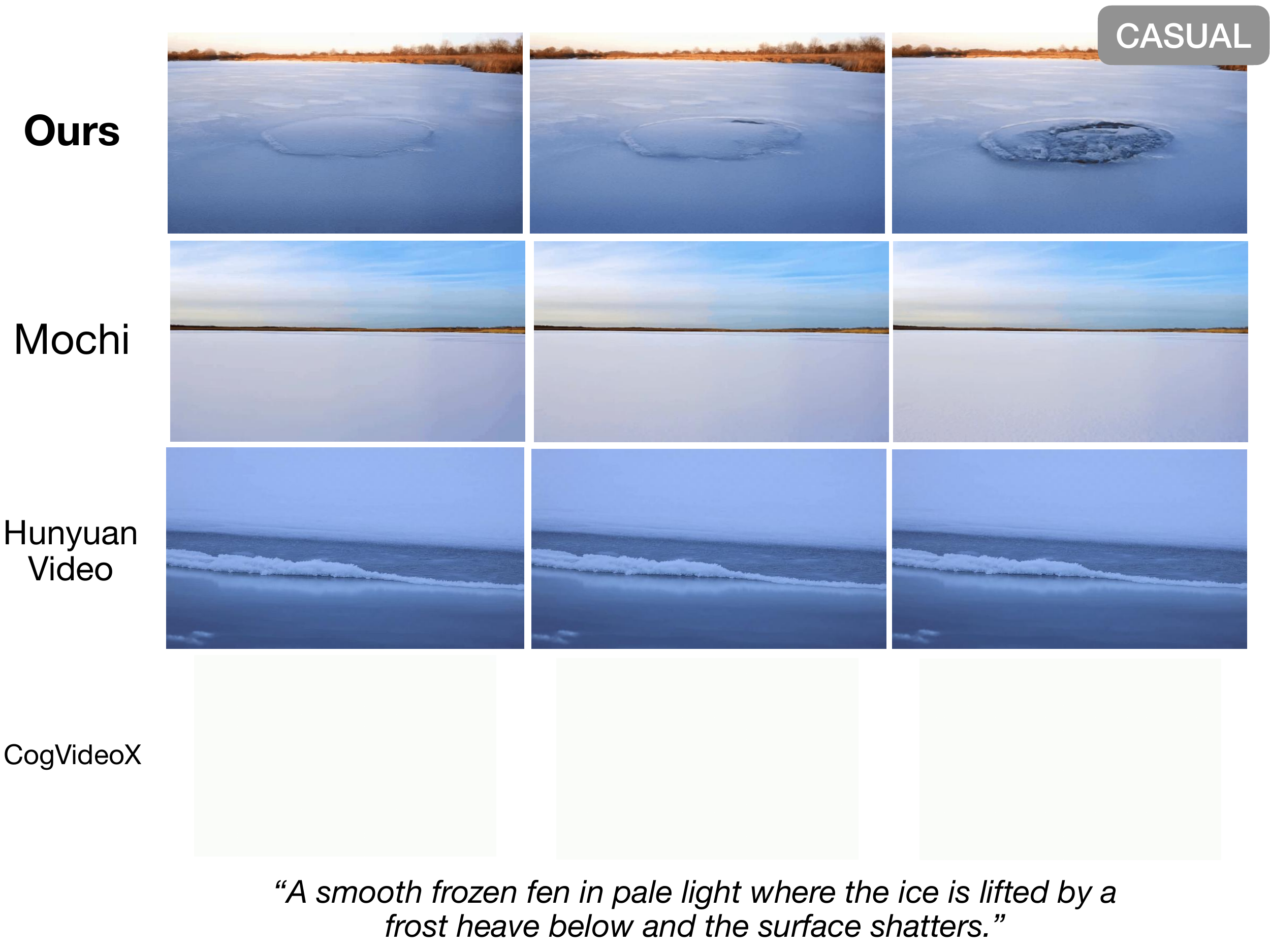}
    \caption{\textbf{CAUSAL.}
    Prompt: ``A smooth frozen fen in pale light where the ice is lifted by a frost heave below and the surface shatters.''}
\end{figure*}

\begin{figure*}[t]
    \centering
    \includegraphics[width=0.95\linewidth]{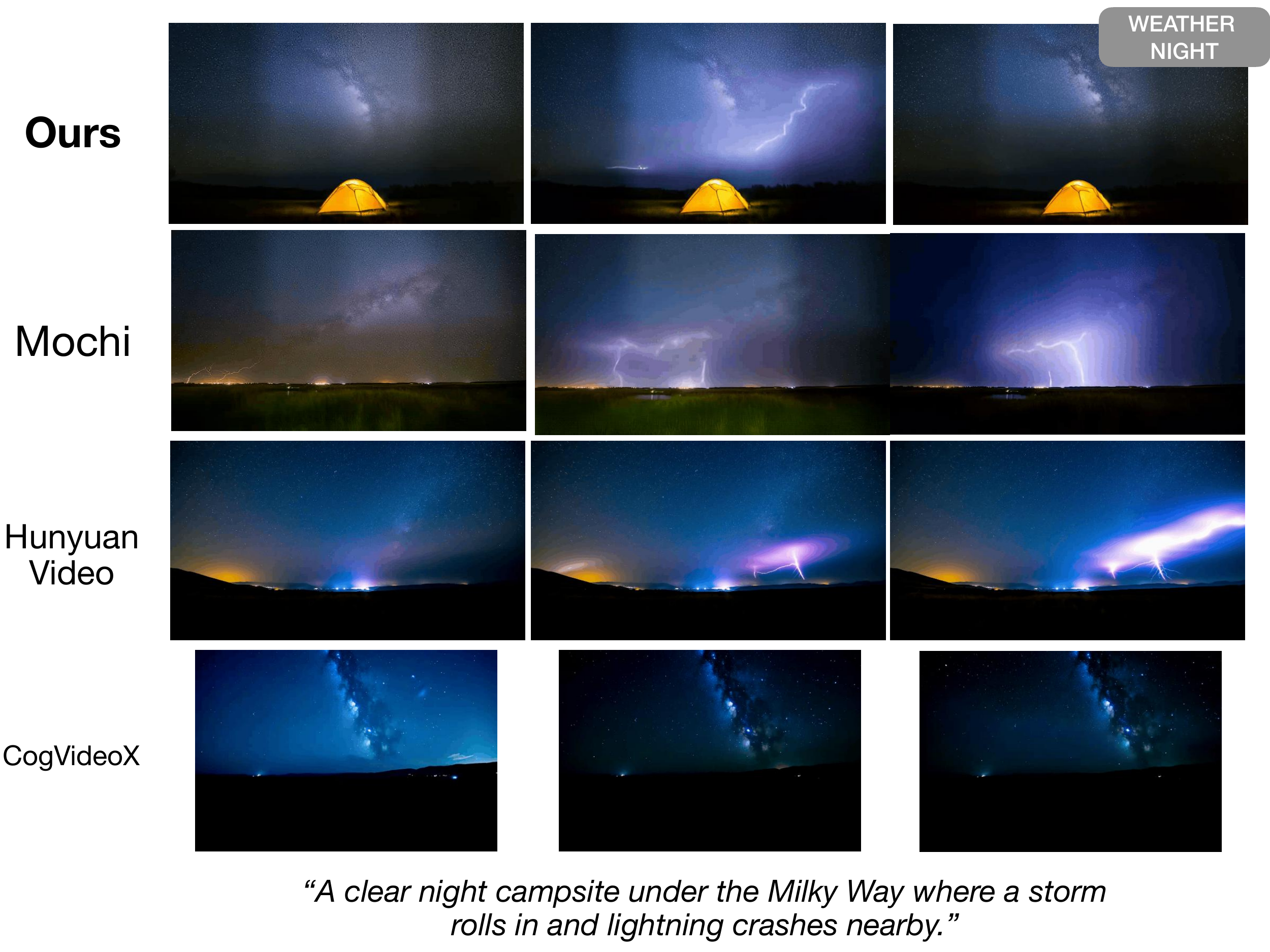}
    \caption{\textbf{WEATHER\_NIGHT.}
    Prompt: ``A clear night campsite under the Milky Way where a storm rolls in and lightning crashes nearby.''}
\end{figure*}

\begin{figure*}[t]
    \centering
    \includegraphics[width=0.95\linewidth]{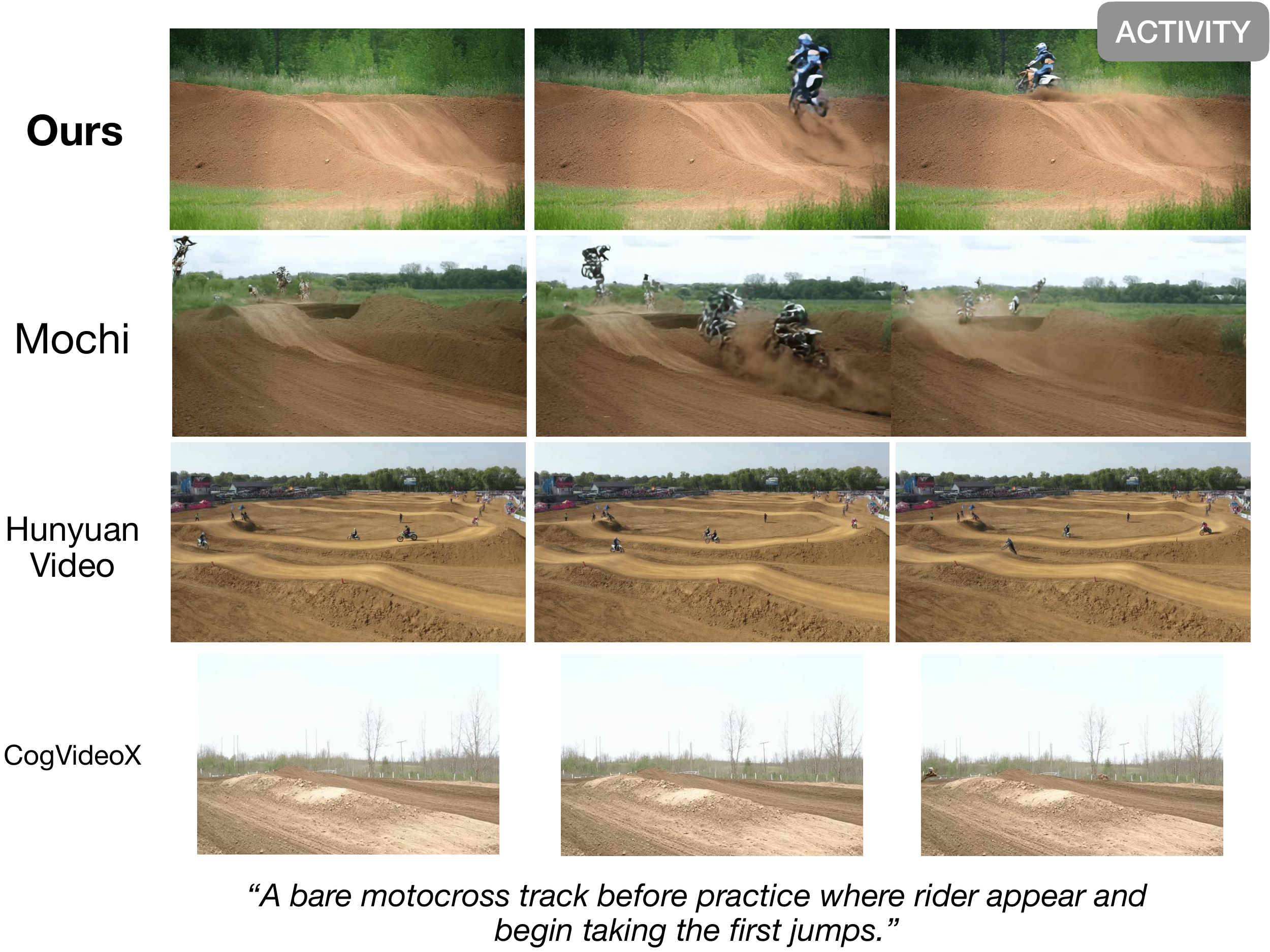}
    \caption{\textbf{ACTIVITY.}
    Prompt: ``A bare motocross track before practice where riders appear and begin taking the first jumps.''}
\end{figure*}

\begin{figure*}[t]
    \centering
    \includegraphics[width=0.95\linewidth]{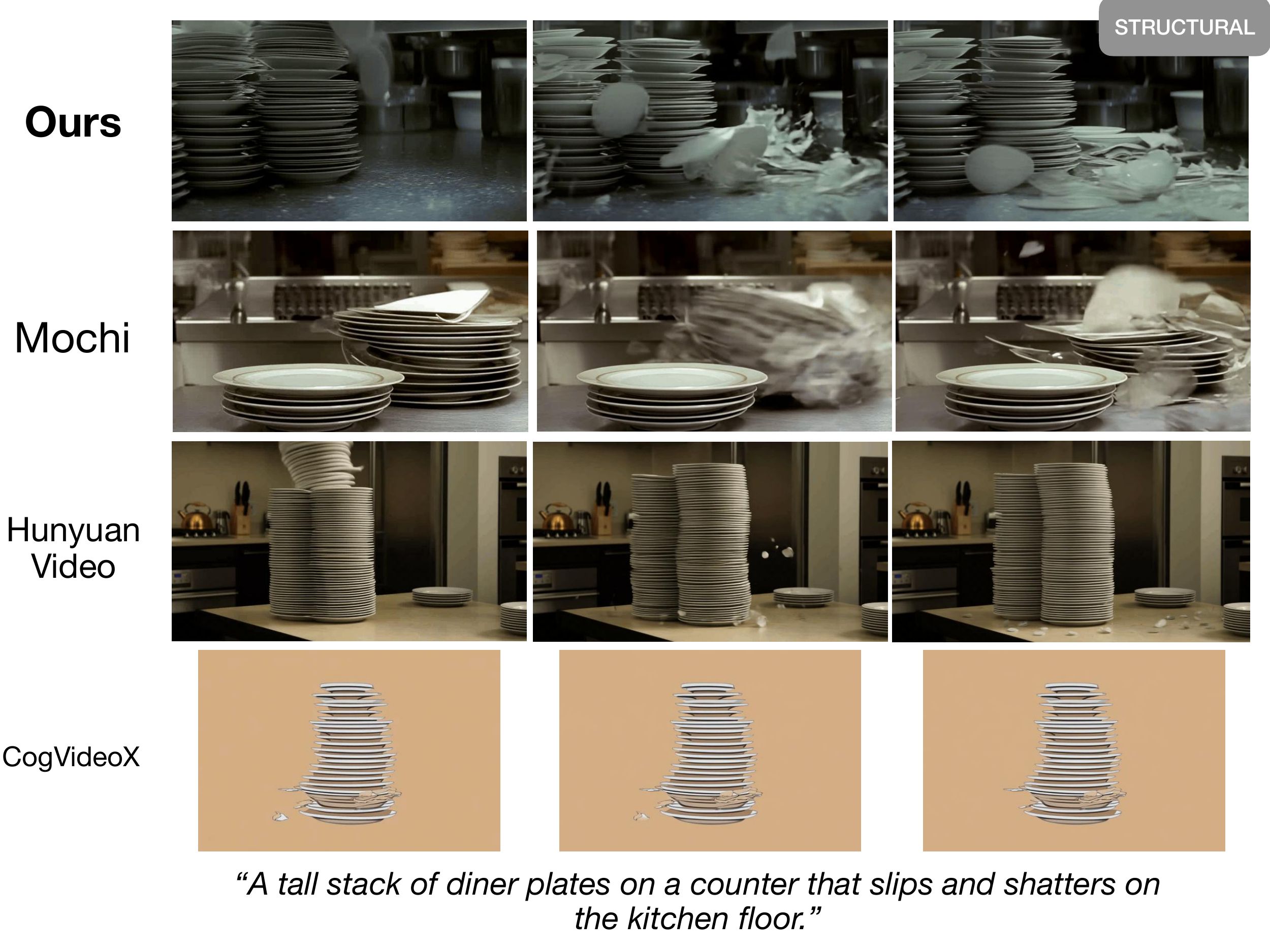}
    \caption{\textbf{STRUCTURAL.}
    Prompt: ``A tall stack of diner plates on a counter that slips and shatters on the kitchen floor.''}
\end{figure*}

\end{document}